\documentclass[]{fairmeta}

\author[*]{Tom\'{a}\v{s} Sou\v{c}ek}
\author[*]{Pierre Fernandez}
\author{Hady Elsahar}
\author{Sylvestre-Alvise Rebuffi}
\author{Valeriu Lacatusu}
\author{Tuan Tran}
\author{Tom Sander}
\author{Alexandre Mourachko}

\affiliation[]{Meta FAIR}
\contribution[*]{Equal contribution}

\abstract{

Invisible watermarking is essential for tracing the provenance of digital content. However, training state-of-the-art models remains notoriously difficult, with current approaches often struggling to balance robustness against true imperceptibility. This work introduces \ours, which sets a new state-of-the-art for image and video watermarking. We first identify three fundamental issues of existing methods: (i) the reliance on proxy perceptual losses such as MSE and LPIPS that fail to mimic human perception and result in visible watermark artifacts; (ii) the optimization instability caused by conflicting objectives, which necessitates exhaustive hyperparameter tuning; and (iii) reduced robustness and imperceptibility of watermarks when scaling models to high-resolution images and videos. To overcome these issues, we first propose an adversarial-only training paradigm that eliminates unreliable pixel-wise imperceptibility losses. Second, we introduce a three-stage training schedule that stabilizes convergence by decoupling robustness and imperceptibility. Third, we address the resolution gap via high-resolution adaptation, employing JND-based attenuation and training-time inference simulation to eliminate upscaling artifacts. We thoroughly evaluate the robustness and imperceptibility of \ours{} on different image types and across a wide range of transformations, and show clear improvements over the state-of-the-art. We finally demonstrate that the model efficiently adapts to video via temporal watermark pooling, positioning \ours{} as a practical and scalable solution for reliable provenance in real-world image and video settings.
}

\correspondence{\email{soucek@meta.com}, \email{pfz@meta.com}}

\metadata[Website]{\url{https://facebookresearch.github.io/meta-seal}}
\metadata[Code]{\url{https://github.com/facebookresearch/videoseal}}

\usepackage{ifthen}  %
\newboolean{set_me_to_remove_todos}
\setboolean{set_me_to_remove_todos}{false} %

\ifthenelse{\boolean{set_me_to_remove_todos}}{
    \newcommand{\pierre}[1]{}
    \newcommand{\hady}[1]{}
    \newcommand{\tomas}[1]{}
    \newcommand{\alex}[1]{}
    \newcommand{\todo}[1]{}
    \newcommand{\remove}[1]{}
}{
    \usepackage[normalem]{ulem}
    \newcommand{\remove}[1]{{\color{red} \sout{#1}}}
    \newcommand{\pierre}[1]{{\color{blue} [\textbf{Pierre}: #1]}}
    \newcommand{\hady}[1]{{\color{purple} [\textbf{Hady}: #1]}}
    \newcommand{\tomas}[1]{{\color{orange} [\textbf{Tomas}: #1]}}
    \newcommand{\alex}[1]{{\color{olive} [\textbf{Alex}: #1]}}
    
    \newcommand{\todo}[1]{{\color{red} [\textbf{TODO}: #1]}}
}

\newcommand{\paragraphcustom}[1]{\vspace{4pt}\noindent\textbf{#1}}

\def\1{\mathbbm{1}}

\newcommand{\eg}{e.g.,\@ }

\newcommand{\nbits}{ n_{\text{bits}} }

\newcommand{\msg}{m}
\newcommand{\w}{w}

\newcommand{\logpval}{\log_{10}p}

\definecolor{Gray}{gray}{0.95}

\newlength\savewidth
\definecolor{metablue}{HTML}{0064E0}

\newcommand{\ours}{\textsc{Pixel Seal}}
\usepackage{float} %

\usepackage{booktabs}
\usepackage{graphicx}
\usepackage{subcaption}
\usepackage{url}
\usepackage{amsmath, amsthm, amsfonts, amssymb, bbm}
\usepackage{dsfont} %
\usepackage{multirow}
\usepackage{array}
\usepackage{pifont}
\usepackage{stfloats}
\usepackage{makecell}
\usepackage{siunitx}

\title{\ours: Adversarial-only training for invisible image and video watermarking}

\begin{document}

\maketitle

\begin{figure}[t]
    \centering
    \includegraphics[width=0.7\linewidth]{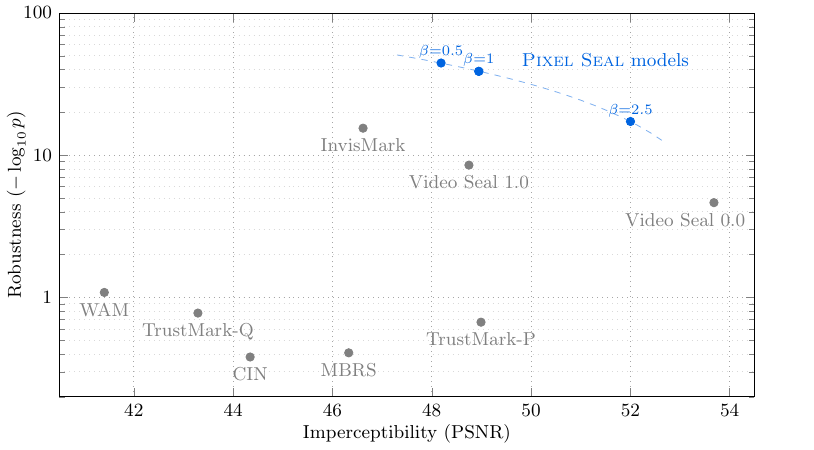}\vspace{-0.2cm}
    \caption{
    \textbf{Imperceptibility and robustness of multi-bit image watermarking methods.} The \ours{} family of models sets new state-of-the-art results for multi-bit image watermarking, both in watermark imperceptibility and robustness. The figure shows average values across 1000 test images generated by Meta AI. %
    The robustness is measured for a combined attack of brightness change (0.5), crop (50\% area), and JPEG compression (quality 40). 
    Imperceptibility of Video Seal~0.0 is heavily skewed due to its small but very visible artifacts. Each \ours{} model is trained with a different watermark boosting factor $\beta$ (see Section~\ref{sec:method:adv} for more details).
    }
    \vspace{-0.1cm}
    \label{fig:teaser}
\end{figure}

\section{Introduction}
\label{sec:intro}

The rapid advancement of generative models such as DALL·E~\citep{ramesh2022dalle2}, Stable Diffusion~\citep{rombach2022high}, Sora~\citep{brooks2024video}, Veo~\citep{googledeepmind2025veo3}, and MovieGen~\citep{polyak2024movie} has enabled the creation of high-fidelity synthetic content at scale. In response, invisible watermarking---the process of imperceptibly embedding a message into digital content---has emerged as a critical infrastructure for ensuring authenticity. This technique not only serves to distinguish synthetic media from real images, but also to establish broader provenance, such as verifying original uploaders and identifying source tools~\citep{castro2025video}. %
With that, there is a pressing need for techniques that are both robust and imperceptible.

Multi-bit image watermarking, as established by the seminal work of \citet{zhu2018hidden}, uses an embedder neural network to embed a binary message into an image as an imperceptible perturbation and an extractor neural network that retrieves the embedded binary message from the perturbed image. 
Typically, the embedder and extractor models are trained end-to-end by minimizing a compound loss function with two opposing objectives: message reconstruction loss to ensure the message can be recovered from a watermarked image, and perceptual losses such as MSE and LPIPS, or adversarial discriminator loss that ensures the embedded watermark remains imperceptible for humans. To achieve robustness of the embedded watermark against common user manipulations, such as application of Instagram filters, cropping, etc., during training, the watermarked image is augmented before the hidden message is retrieved using the extractor. By backpropagating through the augmentations, the model allows for the hidden message to be recovered even after image edits.

Despite the conceptual simplicity of this framework, training a model that is simultaneously robust, fast, and truly imperceptible remains notoriously difficult, as increasing the model's robustness can lead to more perceptible watermarks and increasing the model's imperceptibility often leads to less robust watermarks. We identify three fundamental bottlenecks in existing methods:
First, standard training pipelines rely on a complex mix of perceptual losses to achieve watermark imperceptibility. These losses are often pixel-wise metrics, such as mean squared error (MSE), or deep perceptual metrics, such as LPIPS, but they remain imperfect proxies for human perception. For example, mean squared error loss does not discriminate between smooth and highly textured areas of an image, whereas humans are more likely to notice artifacts in the smooth regions of an image than in the highly textured ones.
Second, jointly optimizing for robustness and imperceptibility creates a contradictory loss landscape---a global optimum for perceptual loss is a zero watermark (no message can be recovered), whereas a global optimum for robustness yields a highly visible watermark. Finding the optimal tradeoff between robustness and imperceptibility requires precise tuning of the learning dynamics. Without it, training frequently collapses---either the model fails to hide the information, or it hides it so well that the decoder cannot retrieve it.
Third, applying existing models to high-resolution media at inference time requires watermark upscaling. This is due to the models being trained on low-resolution crops, and it results in distracting artifacts and inconsistencies that are easily visible to the human eye.
In this work, we systematically address these challenges to advance the state of the art in image watermarking.

Our main contributions are as follows:
\begin{itemize}
    \item First, we introduce an adversarial-only training paradigm that removes standard perceptual losses such as MSE or LPIPS. By relying solely on a discriminator, we avoid the failure modes and manual tuning associated with traditional loss functions.
    
    \item Second, we propose a three-stage training schedule that decouples the competing objectives of robustness and imperceptibility. By first achieving robust (but visible) watermarks and gradually enforcing invisibility, we ensure stable convergence across random initializations.
    
    \item Third, we propose to simulate the inference pipeline during training and apply Just-Noticeable Difference (JND) attenuation at the original input resolution. This eliminates the artifacts commonly found in upscaled watermarks while improving the robustness of the watermarks.
    
    \item Finally, we introduce \ours{}, an image watermarking model that achieves state-of-the-art robustness and imperceptibility (see Figure~\ref{fig:teaser}). We demonstrate that the model can be efficiently adapted to video via temporal watermark pooling with no performance drop.
\end{itemize}

\section{Related work}\label{sec:related}

\noindent\textbf{Traditional methods.}
Early image watermarking methods operated either in the spatial domain~\citep{van1994digital, nikolaidis1998robust} or in frequency domains like DFT~\citep{urvoy2014perceptual}, DCT~\citep{bors1996image, piva1997dct}, and DWT~\citep{xia1998wavelet, barni2001improved}.
In the video domain, watermarking approaches leveraged the specificities of video codecs to decrease the codec's impact on robustness. 
The resulting methods were typically designed for a specific video codec, such as MPEG-2~\citep{biswas2005-RVLC, noorkami2007-RVLC} or H.264/AVC~\citep{chen2006analysis, zhang2007robust, mohaghegh2008h-motionvec}.

\paragraphcustom{Deep-learning based image methods} have progressively replaced the traditional ones, as they make robustness easier by incorporating transformations directly into the training process.
HiDDeN~\citep{zhu2018hidden}, a seminal work in this direction, has inspired many extensions,  through adversarial training~\citep{luo2020distortion}, attention mechanisms~\citep{zhang2020robust}, and robust optimization~\citep{wen2019romark}.
CIN~\citep{ma2022towards} introduced invertible networks for the task of watermarking, while MBRS~\citep{jia2021mbrs} focused on improving robustness to compression through the use of both real and an approximation of JPEG compression.
More recent image watermarking works include resolution-independent embedding~\citep{bui2023trustmark, xu2025invismark}, robustness to diffusion purification~\citep{pan2024jigmark,lu2025vine}, joint copyright protection and tamper localization~\citep{zhang2024editguard}, localized multi-message extraction~\citep{sander2024watermark}, or very long hidden message length~\citep{chunkyseal}.
In the industry, Google's SynthID~\citep{gowal2025synthid} is deployed to watermark AI-generated content, whereas Meta~\citep{castro2025video} utilizes watermarking for a broader range of applications. Yet, details on these methods are scarce.

A parallel research direction focuses on watermarking AI-generated content during the generation process~\citep{yu2020attention, yu2021responsible}, with notable works including Stable Signature~\citep{fernandez2023stable}, Gaussian Shading~\citep{yang2024gaussian}, Tree-Ring~\citep{wen2023tree}, or their follow-ups~\citep{kim2023wouaf, hong2024exact, ci2024ringid} for diffusion models and WMAR~\citep{jovanovic2025wmar} and BitMark~\citep{kerner2025bitmark} for autoregressive models.
In contrast to these works, we focus on watermarking in the \emph{post-hoc} scenario, i.e., after the generation process.

\paragraphcustom{Deep-learning based video methods} are less numerous.
Early works, such as VStegNet~\citep{mishra2019vstegnet} and RivaGan~\citep{zhang2019robust}, adapted HiDDeN to the video domain but faced efficiency challenges with 4D tensors.
Recent works, like DVMark~\citep{luo2023dvmark}, VHNet~\citep{shen2023vhnet}, RC-VWN~\citep{chen2024robust}, or StegaVideo~\citep{hu2024stegavideo}, have focused on improving robustness and imperceptibility through the use of video architectures, differentiable compression simulation, and embedding/extraction in the frequency domain~\citep{zhang2024hide, chang2024dnn}, or tamper localization in videos~\citep{zhang2024v2a}. In contrast to these methods that rely on complex 3D architectures, ItoV~\citep{ye2023itov} and Video Seal~\citep{fernandez2024video} adapt image-based architectures for video watermarking. Similarly, we also adapt \ours{} for video watermarking, but we propose inference time-only adaptation without the need for finetuning.

\section{Preliminaries}\label{sec:img-watermarking}

In this section, we first describe a common post-hoc image watermarking framework, which will serve as a foundation for our \ours{} model described in the next section.

\subsection{Neural-based image watermarking}\label{sec:inference}

A common approach for post-hoc multi-bit image watermarking is to train an embedder neural network that adds a binary message of length $\nbits$ to an image and an extractor neural network that recovers it from the watermarked image. 
We detail their inference next, and describe the joint training process in Section~\ref{sec:training}.

\paragraphcustom{Embedding and extraction of the watermark.}
The watermark embedder takes an image $x \in [0,1]^{H \times W \times 3}$ and a binary message $\msg \in \{0, 1\}^{\nbits}$ as input. 
The output of the embedder is the watermarked image $x_w$ or the watermark $\w \in [-1, 1]^{H \times W \times 3}$ which is then summed with the input image to produce $x_w$:
\begin{equation}\label{eq:watermark_image}
x_w = x + \alpha \cdot \w, \quad \w = \text{Emb}(x, m),
\end{equation}
where $\alpha$ is a scaling factor that controls the strength of the watermark.
The watermark extractor takes a possibly altered version of the watermarked image $x_w$ and outputs a ``soft'' message $\tilde{\msg} \in [0,1]^{\nbits}$ which is thresholded to recover the binary message $\hat{\msg} \in \{0, 1\}^{\nbits}$.

\paragraphcustom{Watermark detection with statistical guarantees.}
Multi-bit watermarking can be used for simple yes/no watermark detection in an image by consistently embedding the same binary message $m$ and then verifying its presence in the image. A key advantage of this protocol is that it provides a statistical bound on false positives, which can be computed as:
\begin{equation}\label{eq:pvalue}
    \mathbb{P}_{m' \sim \mathcal{B}(0.5)^{\nbits}}
        \big[
            d_H(m, m') \leq d_H(m, \hat{m}) 
        \big] = \sum_{k \leq d_H(m , \hat{m}) }^{\nbits} \binom{\nbits}{k} 1/2^{\nbits}
\end{equation}
where $\mathcal{B}(0.5)^{\nbits}$ denotes multivariate Bernoulli distribution and $d_H(\cdot,\cdot)$ is Hamming distance between binary messages. 
Equation~\eqref{eq:pvalue} yields a theoretical error bound assuming, for any non-watermarked image, all binary messages are predicted with equal probability. This theoretical bound illustrates the benefit of multi-bit watermarking, which, unlike other approaches, does not require rigorous calibration on non-watermarked data to establish a reliable false positive rate.

\subsection{Training image watermarking models}\label{sec:training}

Commonly, during training, an image $x$ is processed by the watermark embedder to produce the watermarked image $x_w$. This image is then augmented with various geometric, valuemetric, and compression transformations to simulate editing by users. 
Lastly, the augmented image $\tilde{x}_w$ is processed by the extractor to retrieve the original message hidden in the watermark. The models are trained end-to-end with a combination of various losses, as detailed next.

\paragraphcustom{Losses.}
The post-hoc image and video watermarking training commonly use a combination of perceptual, adversarial, and message losses.
The perceptual loss $\mathcal{L}_{\text{perc}}$ ensures the watermarked image $x_w$ produced by the watermark embedder is close to the original input image $x$. The exact implementation of the loss varies significantly in the literature, ranging from simple mean squared error (MSE) to more complex ones, such as LPIPS~\citep{zhang2018perceptual}, focal frequency~\citep{jiang2021focal}, or Watson perceptual models~\citep{czolbe2020loss}. 
To improve the imperceptibility of the embedded watermark, many works rely on the adversarial loss $\mathcal{L}_{\text{adv}}$ often formulated as $\mathcal{L}_{\text{adv}}=-D(x_w)$.
It maximizes the likelihood that a watermarked image $x_w$ is classified as non-watermarked by the discriminator network $D$. The discriminator itself is jointly trained using a dual-hinge loss~\citep{lim2017geometric} to distinguish between original and watermarked images. 
Lastly, to hide a binary message $m$ in the watermarked image, the binary cross-entropy loss $\mathcal{L}_{\text{msg}}$ is applied to the extractor outputs $\tilde{m}$:
\begin{equation}\label{eq:bce}
\mathcal{L}_{\text{msg}}(m,\tilde{m}) = -\frac{1}{\nbits} \sum_{k=1}^{\nbits} m_k \log(\tilde{m}_k) + (1-m_k) \log(1-\tilde{m}_k).
\end{equation}
The loss is backpropagated through both the extractor and the embedder to guide them to correctly embed and read the hidden message in the watermark.
The final loss function used for the training is the weighted combination of the three losses:
\begin{equation}\label{eq:losses}
\mathcal{L} = \lambda_{\text{perc}} \mathcal{L}_{\text{perc}} + \lambda_{\text{adv}} \mathcal{L}_{\text{adv}} + \lambda_{\text{msg}} \mathcal{L}_{\text{msg}}.
\end{equation}

\paragraphcustom{Augmentations.}
To ensure the embedded message in the watermarked image can be read even if a user modifies the image, during training, the watermarked images are randomly transformed before being passed to the extractor.
Common transformations include geometric changes (e.g., cropping, resizing, and rotation), valuemetric adjustments (e.g., brightness and contrast), and compression (e.g., JPEG). 
Some transformations, such as the JPEG compression, are not differentiable. 
In such a case, similarly to \cite{zhang2021asl}, this work uses a straight-through estimator that approximates the gradient with the identity function:
\begin{equation}\label{eq:straightthrough}
\tilde{x}_w = x_w + \text{nograd}(T(x_w) - x_w),
\end{equation}
where $T$ is the non-differentiable transformation. 
Other works approximate non-differentiable transformations with a trained neural network~\citep{luo2023dvmark, shen2023vhnet}.

\section{Invisible and robust image watermarking}\label{sec:method}

Image watermarking presents several key challenges.
First, the watermark must be invisible to humans; any visible artifacts, especially in flat regions such as the sky, are highly distracting to the viewer. 
Second, the watermark must be detectable even if the watermarked image is altered by various user edits, such as cropping or Instagram filters. 
Training such a robust model is highly sensitive to exact hyperparameter values, and multiple random initializations can yield significantly different outcomes.
Lastly, watermarking must be fast and imperceptible even for very high-resolution images. A common inference-time interpolation technique for high-resolution watermarking introduces distracting artifacts that are easily detectable by the human eye.

To address these challenges, we introduce the following contributions.
First, to produce invisible watermarks, we introduce adversarial-only training without any perceptual losses (Section~\ref{sec:method:adv}). Second, to make the model robust and the training stable and repeatable, we introduce a three-stage training schedule in which we first achieve visible but very robust watermarks that are made invisible later during training (Section~\ref{sec:method:threestage}).
Third, to allow for practical high-resolution watermarking, we propose training-time inference simulation to eliminate upscaling artifacts and increase model's robustness (Section~\ref{sec:method:highrestrain}).
Lastly, we introduce temporal watermark pooling---a technique for efficiently adapting image watermarking models to video (Section~\ref{sec:method:temporalpool}).

\begin{figure*}[t!]
    \centering
    \includegraphics[width=\linewidth, clip, trim={0 6.19in 3.15in 0}]{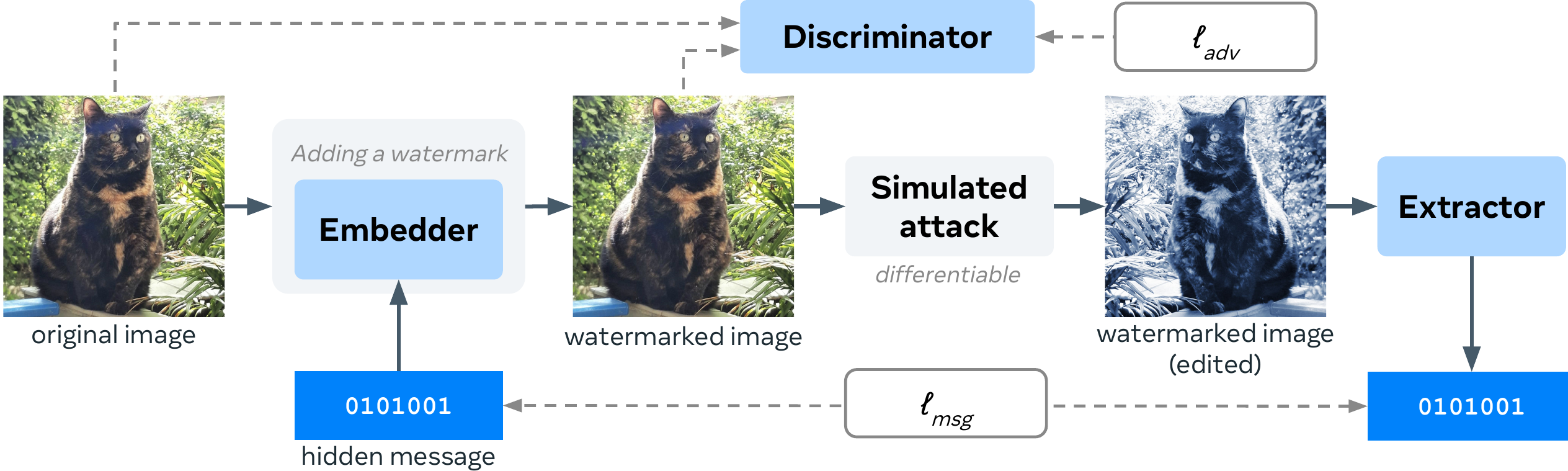}
    \caption{
    \textbf{Our training setup.} 
    \ours{} is trained using only adversarial and message loss, which results in highly imperceptible watermarks. The message loss is backpropagated through the simulated attack (image augmentation) to ensure the embedded watermarks are robust to common user edits.
    }\label{fig:training}
\end{figure*}

\subsection{Adversarial-only loss function}\label{sec:method:adv}

Ensuring that a watermark remains invisible in the content is the primary challenge in post-hoc watermarking research. For example, \cite{bui2023trustmark} utilize four different losses specifically designed to make watermarks invisible. However, many such loss functions have modes in which the loss value is small, but the watermark is very visible to humans. 
Therefore, making the watermark invisible requires careful hyperparameter tuning without any guarantees of success.
And even with carefully tuned hyperparameters, watermarks often remain visible in flat areas as waves \citep{bui2023trustmark} or blobs \citep{fernandez2024video}.

In the attempt to solve the imperceptibility issue, in contrast to previous work, as shown in Figure~\ref{fig:training}, we remove all perceptual losses and train the watermarking model using only the cross-entropy message loss $\mathcal{L}_{\text{msg}}$ and an adversarial loss $\mathcal{L}_{\text{adv}}(x, x_w)$:

\begin{equation}
    \frac{1}{2}\nabla_{\theta_D}\Bigl[\text{ReLU}\bigl(1-D(x)\bigr) + \text{ReLU}\bigl(1+D(x_w)\bigr)\Bigr]-\lambda_{adv}\nabla_{\theta_{\text{Emb}}}D(x_w)
    \label{eq:adv_loss}
\end{equation}

where $x$ and $x_w$ are the original and watermarked images, $D$ is a patch-based discriminator network, $\theta_D$ its parameters, and $\theta_{\text{Emb}}$ are the parameters of the watermark embedder. 
The architecture of the discriminator affects how effectively it can remove visible artifacts. The more effective the discriminator is in removing visible artifacts, the less robust the final watermark is to various content modifications. In this work, we chose the patch-based discriminator of \cite{rombach2022high} as it provided a good balance. %

To control the imperceptibility of the watermark, we introduce the \emph{watermark boosting} technique. The watermark boosting artificially amplifies the artifacts that are present in a watermarked image, making them easier for the discriminator to be detected and removed. In detail, the boosted watermarked image $\hat{x}_w$, computed as $\hat{x}_w=x + \beta(x_w-x)$, replaces the regular watermarked image $x_w$ in Equation~\eqref{eq:adv_loss}. Setting $\beta>1$ ensures greater imperceptibility of the watermark, while $\beta<1$ allows for very robust watermarks.

While the adversarial loss ensures the watermarked image does not contain any noticeable unnatural artifacts, the watermarked image can still significantly deviate from the original reference image. 
This is a lesser issue for watermarking of AI-generated content, where users are not provided with the original reference image, but it can be a major challenge for other provenance applications~\citep{castro2025video}, such as professional photography devices, where the watermark must be minimal (ideally comparable to camera noise), so as not to alter the user’s intent.
To restrict the allowed deviation the watermark $\w$ can introduce to the original content, we reduce the watermark magnitude by a global scaling factor $\alpha$ and a local Just Noticeable Difference \citep{zhang2008just} attenuation map as done by~\cite{sander2024watermark}. 
The attenuation map is computed using a non-learnable function $\text{JND}:[0,1]^{H\times W\times 3}\to [0,1]^{H\times W}$ which assigns large values for pixels where changes to pixel intensities will likely remain unnoticed by humans (\eg edges) and small values otherwise. The watermarked image $x_w$ is therefore computed as follows:
\begin{equation}
x_w = x + \alpha \cdot \left(w \odot \text{JND}(x)\right), \quad w = \text{Emb}(x, m).
\end{equation}

\begin{figure*}[t!]
    \centering
    \includegraphics[width=\linewidth, clip, trim={0 8.05in 4in 0}]{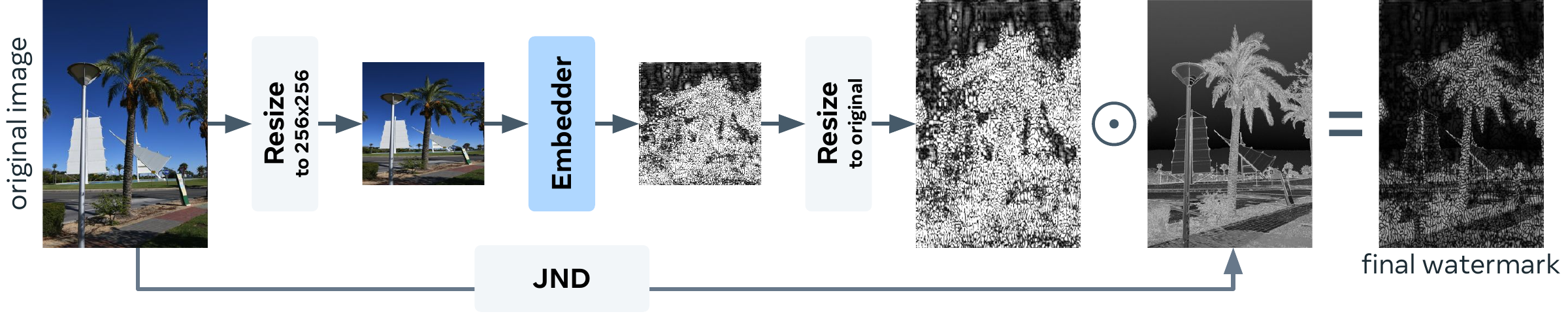}\vspace{-0.15cm}%
    \caption{
    \textbf{Training and inference of \ours{} on high-resolution images.} The input image is first resized to the model resolution (256$\times$256) and the raw watermark is computed using the \ours{} embedder. Then, this watermark is resized to the original input resolution and pixel-wise multiplied by the Just-Noticeable Difference (JND) map to obtain the final high-resolution watermark.
    }\label{fig:highres}
    
\end{figure*}

\subsection{Three-stage training schedule}\label{sec:method:threestage}
Training the watermark embedder and extractor can be very brittle and sensitive to hyperparameters, as global optima of perceptual and adversarial losses correspond to an empty watermark $x_w=x$ ($w=\bm{0}$). Related works stabilize the training process by reducing the weight of perceptual losses $\lambda_{\text{perc}}$ at the beginning of the training. \cite{xu2025invismark} even delay the application of augmentations $\tau$ to the later stages of training to improve the convergence of the model. 
Rather than carefully tuning hyperparameters, we observe a three-stage training can resolve the instabilities and allow the model to repeatably converge to a similar solution.

In detail, in the first stage, the model is trained only with the message reconstruction loss $\mathcal{L}_{\text{msg}}$ and the watermark scaling factor is set to a large value $\alpha=\alpha_0$, as done by~\citet{sander2024watermark}.
At this stage, the predicted watermark is highly visible, yet it facilitates a stable learning process.
We train the model until the bit accuracy is saturated, after which we move to the next stage.
In the second stage, we add the adversarial loss $\mathcal{L}_{\text{adv}}$ and we gradually decrease the watermark scaling factor $\alpha$ from the initial value $\alpha_0$ to the final value $\alpha_1$ using a cosine schedule.
We use the following schedule:
\begin{equation}
\alpha(t) = \alpha_1 + (\alpha_0 - \alpha_1) \cdot \cos\left(\frac{\pi}{2}\varphi\right), \quad \varphi=\mathrm{clip}\left(\frac{t-N_{start}}{N_{end}-N_{start}}, 0, 1\right).
\end{equation}
The factor $\varphi$ controls the interpolation schedule based on the current epoch $t$, the second stage start epoch $N_{start}$, and the second stage end epoch $N_{end}$. The second-stage training ensures that the resulting watermark is imperceptible while still being robust against various attacks.
In the final stage, we finetune the model with the final watermark scaling factor value $\alpha=\alpha_1$, potentially using different types of data, attacks, etc., to tailor the watermarking model for any specific application needs.

\subsection{High-resolution adaptation}\label{sec:method:highrestrain}
Computing a watermark directly on a high-resolution image with millions of pixels is computationally prohibitively expensive. Therefore, a common approach is to train a watermark embedder and extractor using a fixed spatial resolution, e.g., 256$\times$256, and use watermark interpolation technique~\citep{bui2023trustmark,sander2024watermark} for inference. 
The technique works by downsampling the original image $x$ to the fixed model resolution using bilinear interpolation and later upsampling the low-resolution watermark $w$ back to the original resolution as shown in Equation~\eqref{eq:downsample_upsample}.
\begin{equation}\label{eq:downsample_upsample}
    x_w = x + \alpha \cdot \text{resize}_\uparrow(w), \quad w = \text{Emb} \left( 
        \text{resize}_\downarrow(x), m
    \right)
\end{equation}

However, this approach has three major downsides. First, the watermark upsampling can often result in artifacts being spilled into regions where they are more visible (e.g., areas around edges). Second, some artifacts that are imperceptible in low resolution can be easily spotted in high resolution. And third, the watermark extractor is trained to extract hidden messages from images $\tilde{x}_w = \text{resize}_\downarrow(x) + \alpha \cdot w$, where both $\tilde{x}_w$ and $w$ are in the fixed model resolution, while during inference the extractor detects the hidden message in image $\tilde{x}'_w = \text{resize}_\downarrow(x + \alpha \cdot \text{resize}_\uparrow(w))$. Because in the general case $\tilde{x}_w$ and $\tilde{x}'_w$ are not equal, the watermark extractor is tasked with detecting watermarks in out-of-distribution data.

We address these challenges as follows: First, to address the spill of watermark artifacts into flat regions where they are more visible, we compute the JND attenuation map in the original image resolution. Although this increases the computational requirements during inference, the increase is substantially lower than the computation of the watermark in higher resolution. Second, we address the perceptibility of certain artifacts in higher resolution by computing the adversarial loss at the original image resolution. %
Although this increases training compute requirements, it has no effect on inference. 
Third, to eliminate the distribution shift of the watermark extractor input during inference, we simulate the same inference process during training. Again, this has no effect on inference. In detail, the training pipeline is outlined in Equation~\eqref{eq:pipeline}, with the key components illustrated in Figure~\ref{fig:highres}.
\begin{equation}\label{eq:pipeline}
    \tilde{m} = \text{Ext}(\tilde{x}_w), \quad
    \tilde{x}_w = \text{resize}_\downarrow(\tau(x_w)), \quad
    x_w = x + \alpha \cdot \left(\text{resize}_\uparrow(w) \odot \text{JND}(x)\right),
    \quad w = \text{Emb} \left(\text{resize}_\downarrow(x), m\right)
\end{equation}
The original image $x$ and the watermarked image $x_w$ are in the full resolution, while the watermark $w$ and the input to the watermark extractor $\tilde{x}_w$, augmented by a random differentiable augmentation $\tau$, are in the fixed model resolution.

\begin{figure*}[t!]
    \centering
    \includegraphics[width=\linewidth, clip, trim={0 8.2in 1.8in 0}]{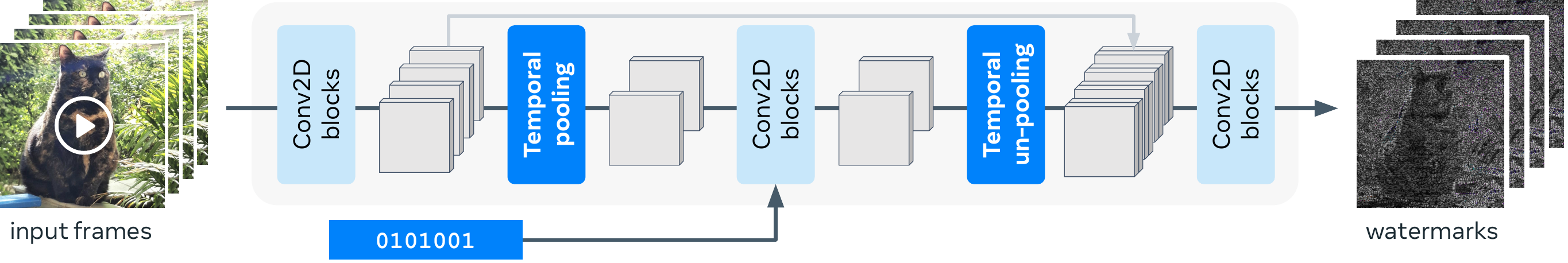}
    \caption{
    \textbf{The embedder with temporal watermark pooling enabled.} During inference, a temporal average pooling and un-pooling layer is inserted into the embedder. This modification results in a significant speedup for video watermarking, with no impact on imperceptibility or robustness.
    }\label{fig:temppooling}
\end{figure*}

\subsection{Video watermarking through image model adaptation}\label{sec:method:temporalpool}

Video watermarking presents two additional challenges compared to image watermarking. The first challenge is the complexity of video training itself---video decoding and pre-processing are resource-intensive, and the high correlation within adjacent video frames results in insufficient diversity within a batch. The second challenge is the computationally expensive process of watermarking each video frame during inference, which renders any application of a video watermarking model impractical, even with the resizing trick introduced in Section~\ref{sec:method:highrestrain}.

The simplest solution to the second challenge, i.e., increasing inference speed, is to watermark every $k$-th frame. However, this introduces distracting flickering if the watermark is not fully imperceptible, %
makes the watermark vulnerable to video compression, and requires exhaustive scanning during extraction since most frames are watermark-free. Therefore, \cite{fernandez2024video} propose to propagate the same watermark computed for every $k$-th frame to the remaining $k-1$ frames. While fast, this approach introduces ghosting artifacts and requires video finetuning with the exact parameter $k$ to make the method robust. Instead, we introduce \textit{temporal watermark pooling}---an inference-time method applicable to our image watermarking model that alleviates the need to train on video while allowing for much faster inference without any loss in visual quality or robustness compared to the baseline of watermarking every frame.

The temporal watermark pooling, shown in Figure~\ref{fig:temppooling}, is based on the observation that adjacent video frames are highly correlated. That is, the frames will likely have very similar high-level semantic features. Therefore, we propose replacing these per-frame high-level semantic features with their average across neighboring frames. In detail, during inference only, we insert a temporal average pooling layer with a kernel size and stride of $ k$ into our embedder after the $d$-th downsampling U-Net block. Similarly, we insert a temporal un-pooling (repeat) layer at the same position in the upsampling blocks of the U-Net. With $d$ small, our temporal watermark pooling achieves a significant speedup while having no impact on imperceptibility or robustness due to the low-level per-frame changes computed using the last U-Net layers.

\begin{figure*}[b!]
    \centering
    \footnotesize
    \newcommand{\imwidth}{0.19\textwidth}
    \setlength{\tabcolsep}{0pt}
    \resizebox{1.0\linewidth}{!}{
    \begin{tabular}{c@{\hskip 2pt}c@{\hskip 2pt}c@{\hskip 2pt}c@{\hskip 2pt}c}
\textit{Original} & CIN & InvisMark & MBRS & TrustMark-P \\
\includegraphics[width=\imwidth]{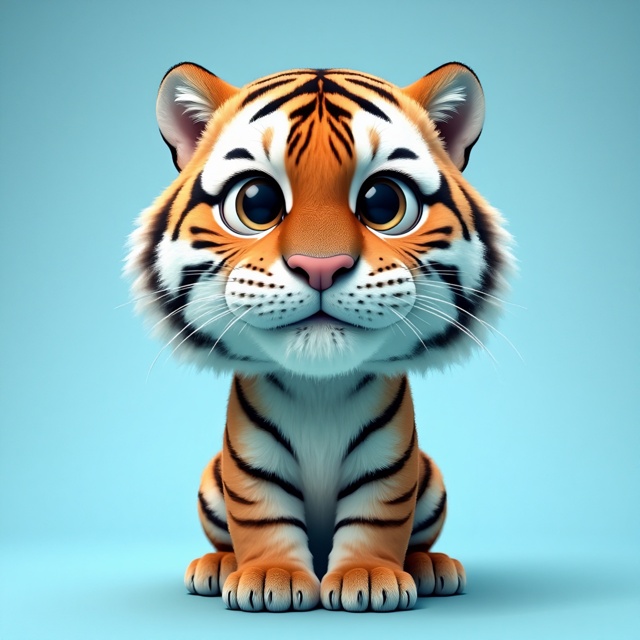} &
\includegraphics[width=\imwidth]{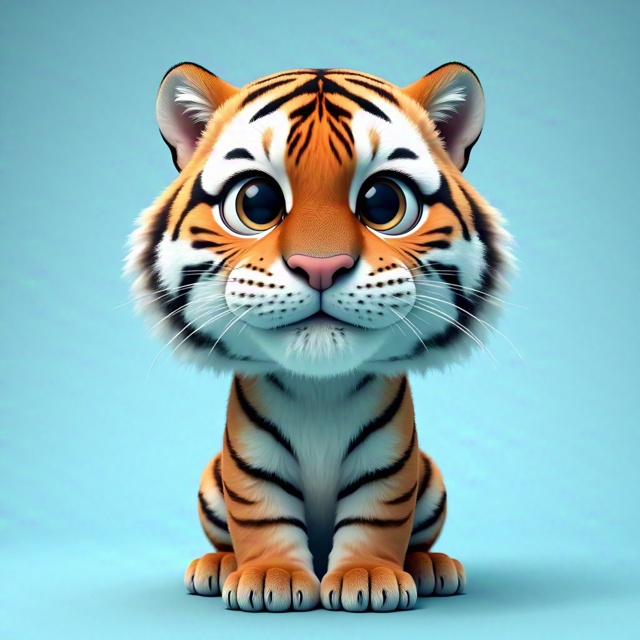} &
\includegraphics[width=\imwidth]{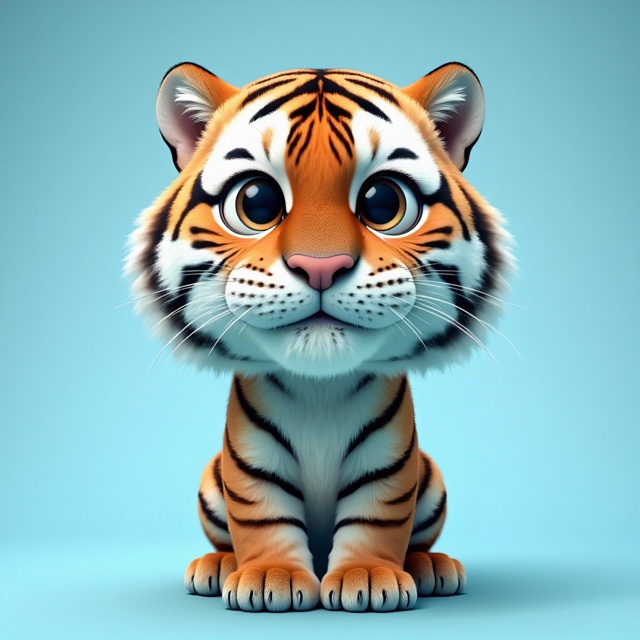} &
\includegraphics[width=\imwidth]{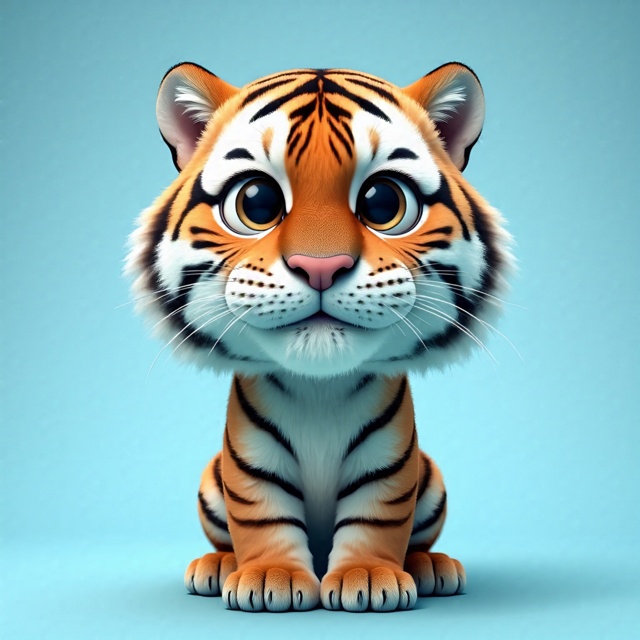} &
\includegraphics[width=\imwidth]{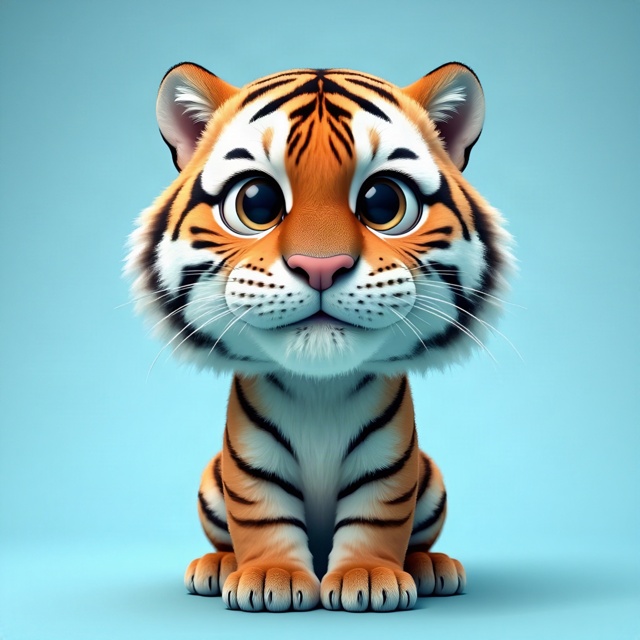} \\
& \includegraphics[width=\imwidth]{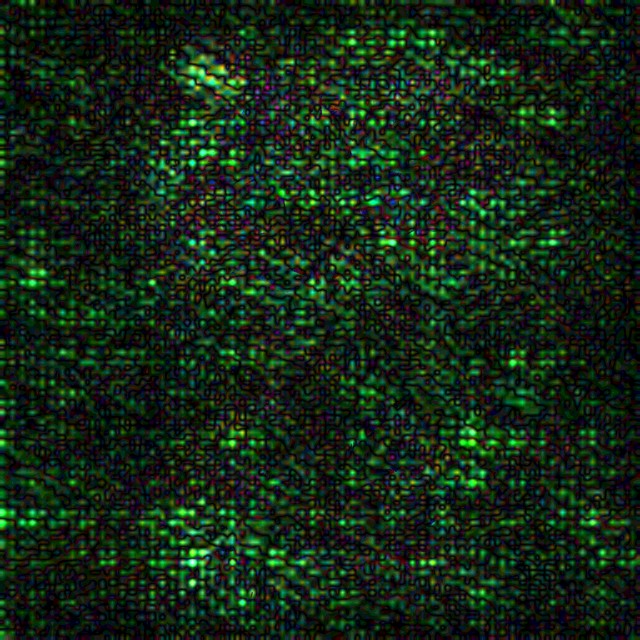} &
\includegraphics[width=\imwidth]{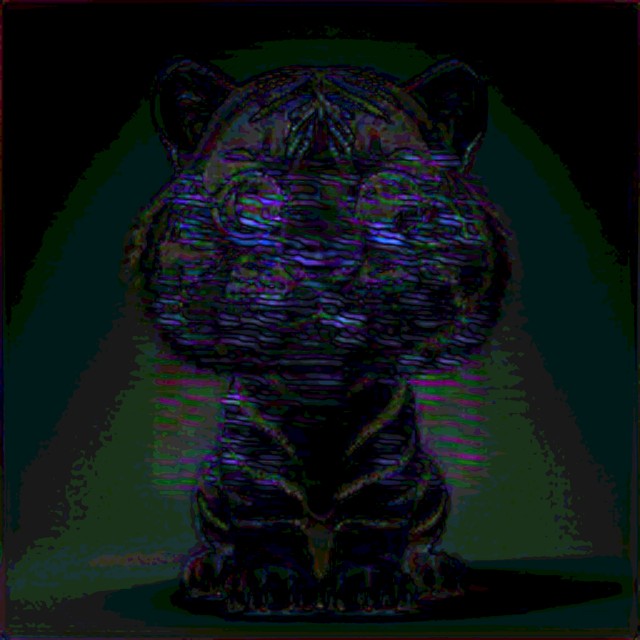} &
\includegraphics[width=\imwidth]{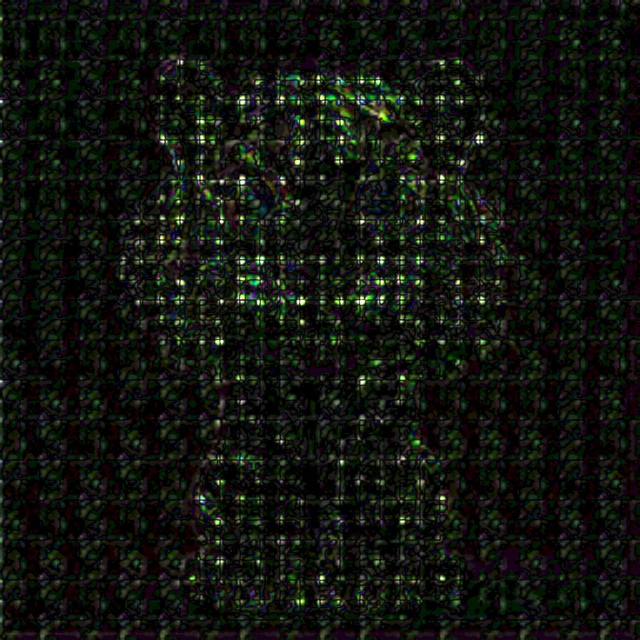} &
\includegraphics[width=\imwidth]{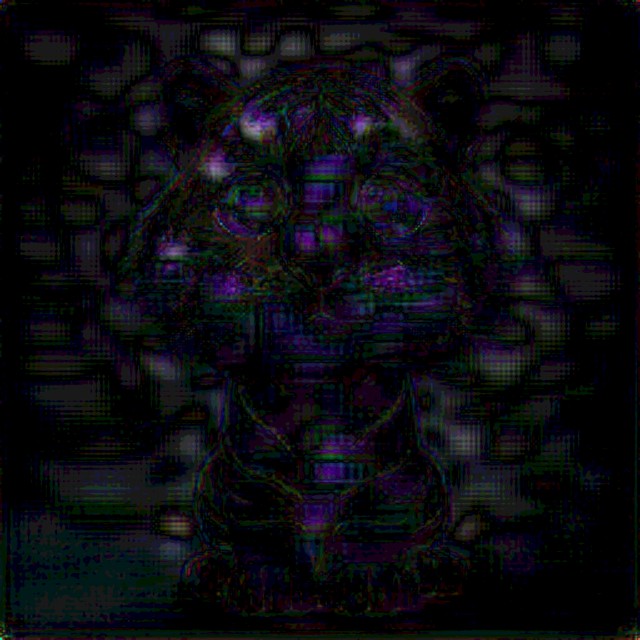}\\
\end{tabular}
}
\vspace{3pt}
\resizebox{1.0\linewidth}{!}{
\begin{tabular}{c@{\hskip 2pt}c@{\hskip 2pt}c@{\hskip 2pt}c@{\hskip 2pt}c}
TrustMark-Q & Video Seal 0.0 & Video Seal 1.0 & WAM & \ours{} (ours) \\
\includegraphics[width=\imwidth]{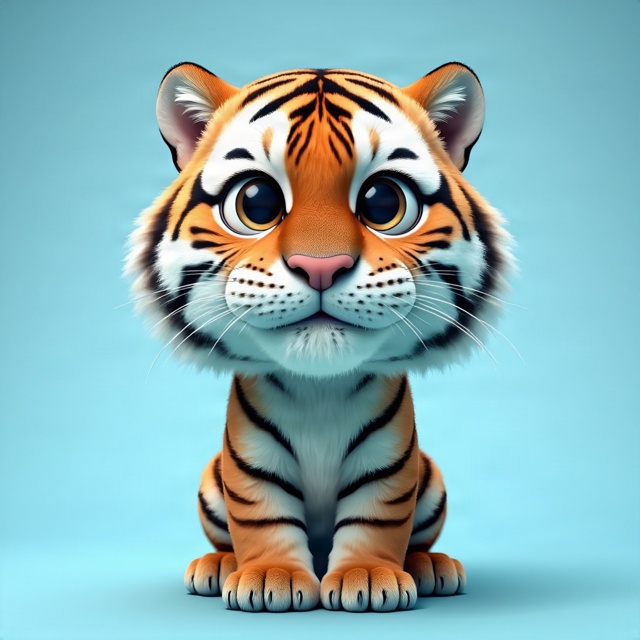} &
\includegraphics[width=\imwidth]{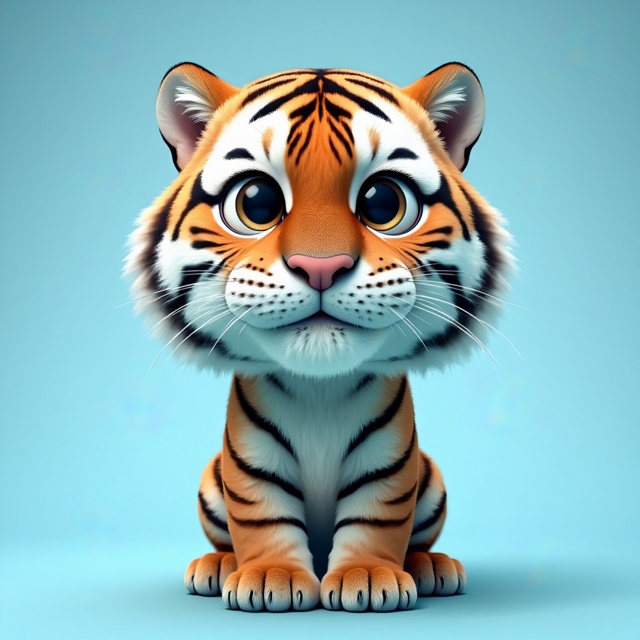} &
\includegraphics[width=\imwidth]{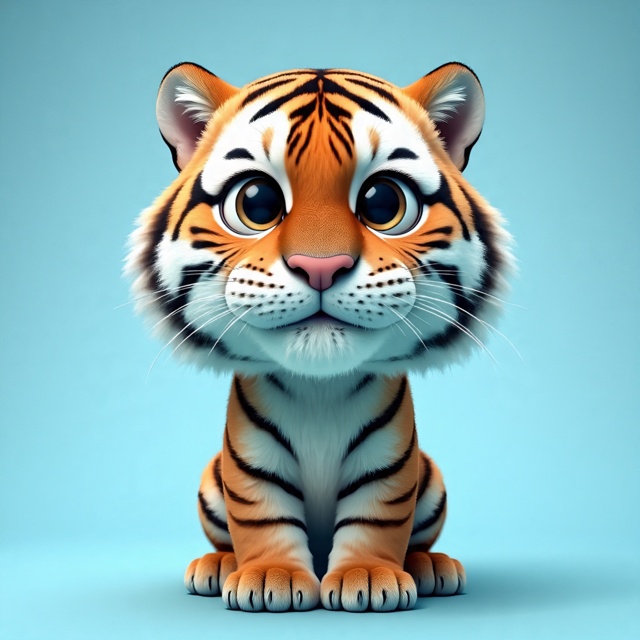} &
\includegraphics[width=\imwidth]{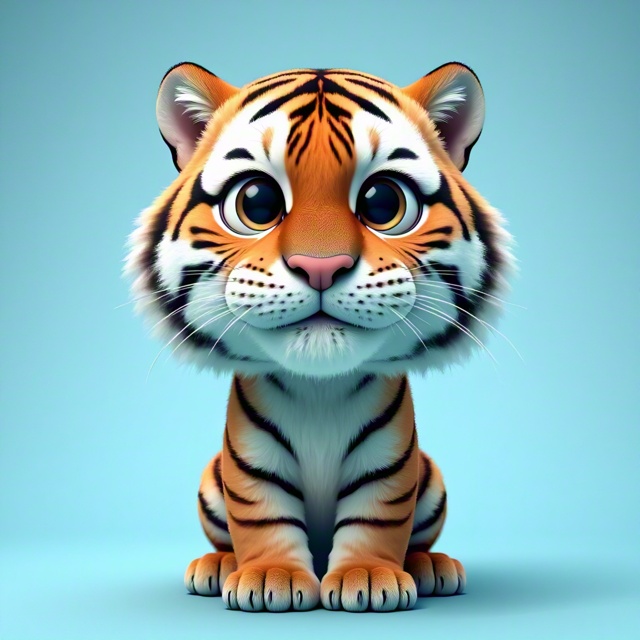} &
\includegraphics[width=\imwidth]{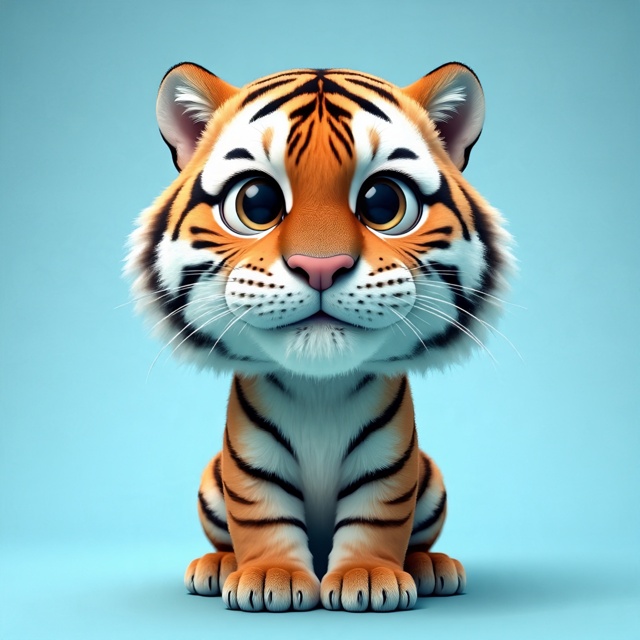} \\
\includegraphics[width=\imwidth]{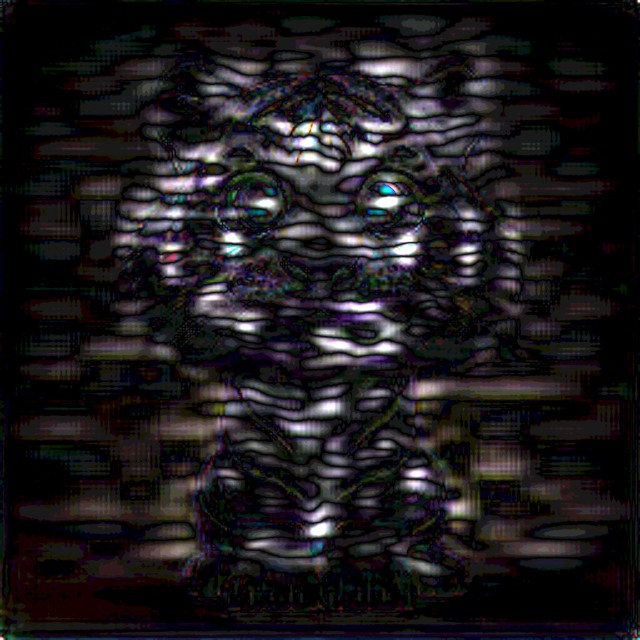} &
\includegraphics[width=\imwidth]{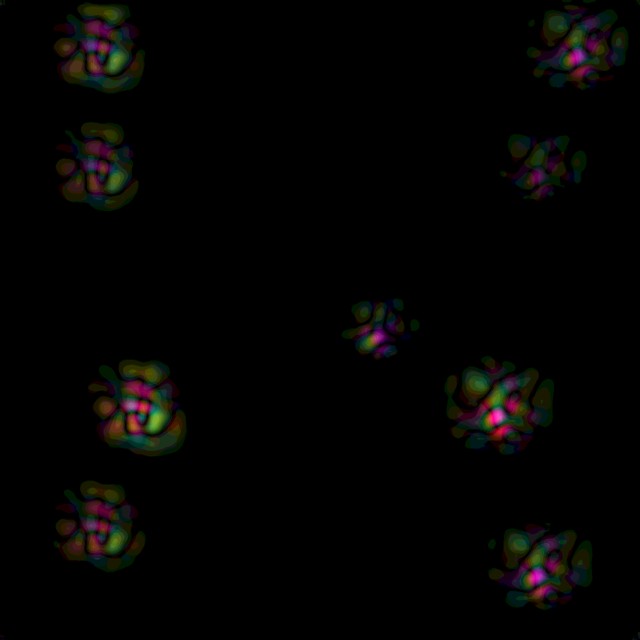} &
\includegraphics[width=\imwidth]{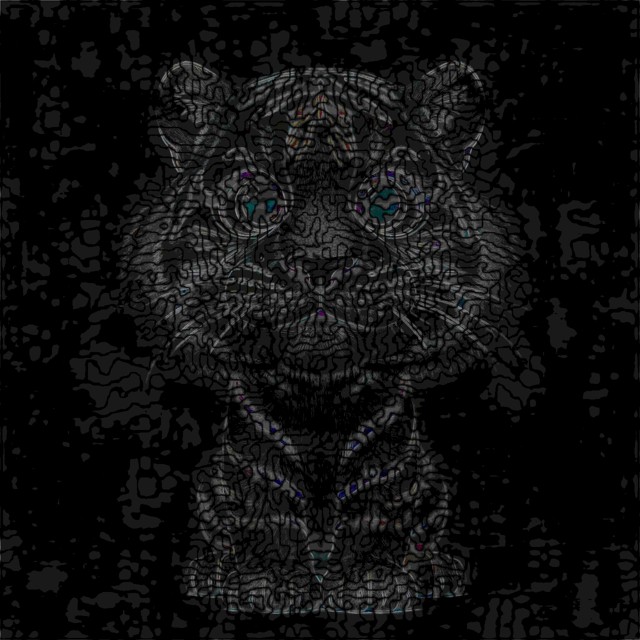} &
\includegraphics[width=\imwidth]{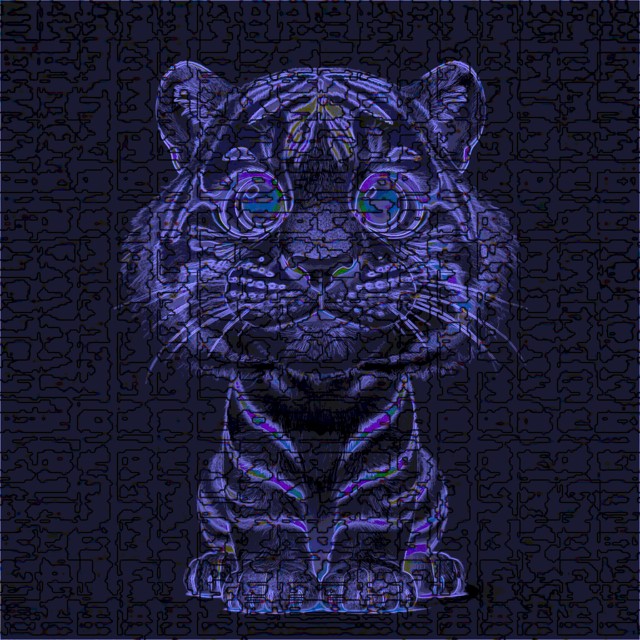} &
\includegraphics[width=\imwidth]{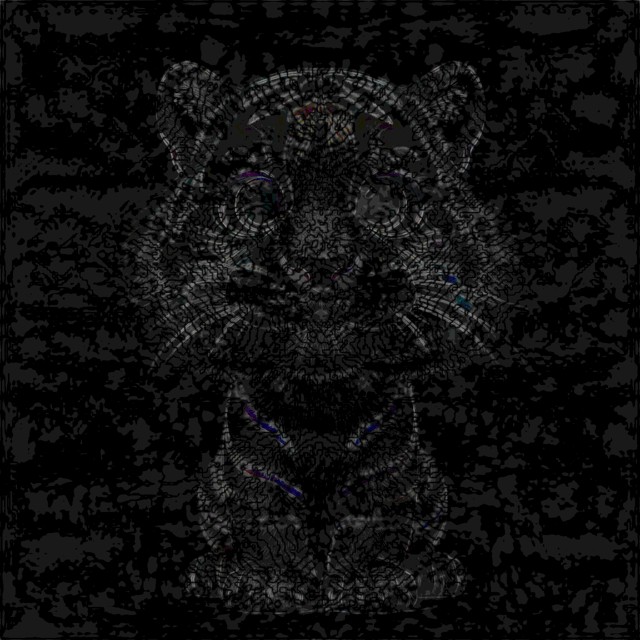} \\
    \end{tabular}
    }
    \caption{\textbf{Comparison with related work on an AI-generated image.} We show both the watermarked image (top) and the predicted watermark brightened for clarity (bottom).
    Many related methods leave visible artifacts in areas with a single color. In contrast, \ours\ does not leave visible artifacts in such areas while being more robust to various transformations.
    More examples are available in the appendix.
}\label{fig:example_comp} 
\vspace{-0.4cm}
\end{figure*}

\section{Experiments}\label{sec:results}

In this section, we present the experimental setup and results of our image watermarking approach. 
We describe the implementation details and evaluate our method against state-of-the-art image watermarking models. 
Additionally, we conduct ablation studies to thoroughly verify our design choices. %

\subsection{Implementation and evaluation details}

\noindent\textbf{Implementation details.}
We train our model, called \ours{}, with a U-Net-based architecture~\citep{saharia2022photorealistic} with 43.8M parameters for watermark embedding and ConvNext-v2 Tiny~\citep{woo2023convnext} with 33.4M parameters for watermark extraction.
The model training is performed on images from the SA-1b dataset~\citep{kirillov2023segment} for 600k steps with a batch size of 256. 
We use AdamW optimizer with a learning rate of 5$\times 10^{-4}$, decayed via a cosine schedule, and a linear warm-up period of 20k steps.
We apply our high-resolution training recipe described in Section~\ref{sec:method:highrestrain}, but for practical reasons, we limit the maximum image size to $S_{max}$$\times$$S_{max}$.
To further ensure robustness against different image aspect ratios, we randomly resize the image into a width/height sampled from the interval $[S_{min}, S_{max}]$. We set $S_{min}=256$ and $S_{max}=768$.
During the fist stage training we set the watermark scaling factor to $\alpha_0=1.0$ and decrease it during the second stage to the final value of $\alpha_1=0.2$. The released \ours{} model is trained with the discriminator of~\cite{rombach2022high} with the default watermark boosting factor $\beta=1$.
The objectives are weighted with $\lambda_{\text{msg}} = 1.0$ and $\lambda_{\text{adv}} = 0.1$. We do not use any perceptual loss.
Our models encode $\nbits=256$ bits.

\paragraphcustom{Baselines.}
We compare \ours{} with other publicly available multi-bit watermarking methods. We consider CIN~\citep{ma2022towards} with 30-bit message, InvisMark~\citep{xu2025invismark} with 94-bit message\footnote{InvisMark encodes 100-bit message; however, due to its random message sampling process during training, 6 bits of the message have their values fixed.}, MBRS~\citep{jia2021mbrs} with 256-bit message, TrustMark~\citep{bui2023trustmark} with 100-bit message, Video Seal 0.0~\citep{fernandez2024video} with 96-bit message, Video Seal 1.0 with 256-bit message, and WAM~\citep{sander2024watermark} with 32-bit message.
For video watermarking, we also compare with RivaGAN~\citep{zhang2019robust} with 32 32-bit message. Similarly to \ours{}, the baselines also operate at a fixed resolution, i.e., we use Equation~\eqref{eq:downsample_upsample} when evaluating them on high-resolution content.

\paragraphcustom{Evaluation details.}
We evaluate the methods using 1000 images of resolution 1280$\times$1280 generated by Meta AI using the prompts from \cite{gowal2025synthid} and 100 real high-resolution photos from the SA-1b validation set \citep{kirillov2023segment}.
For video evaluation, we further use a dataset of 121 videos\footnote{The videos are available at \href{https://www.youtube.com/playlist?list=PL86eLlsPNfyi27GSizYjinpYxp7gEl5K8}{www.youtube.com/playlist?list=PL86eLlsPNfyi27GSizYjinpYxp7gEl5K8}.} generated by Movie~Gen~\citep{polyak2024movie} and an SA-V validation set of 96 high-quality videos.
The evaluation is performed on the first 3 seconds of each video.
We measure both watermark imperceptibility and robustness. For imperceptibility, we use PSNR, SSIM, LPIPS~\citep{zhang2018perceptual}, CVVDP~\citep{mantiuk2024colorvideovdp}, and our just-noticeable difference (JND) metric. 
The JND metric is computed using a regressor trained on internal user study data to measure the perceptual quality difference between the watermarked image and the reference. 
A difference of 1 JND means that 75\% of observers would find the difference noticeable\footnote{Please note that the JND metric described here differs from the JND heatmap used to attenuate the watermark on flat areas.}.
For evaluating watermark robustness, we apply various transformations (e.g., brightness change, cropping, compression) and report both the bit accuracy and $\logpval$.
\looseness=-1 The latter represents the logarithm of the probability of observing the measured bit accuracy by chance (Equation~\eqref{eq:pvalue}). 
This allows for a fair comparison between methods with different payload sizes (we refer to \citet[App. A]{fernandez2024video} for more information on $\logpval$).
The full list of evaluated transformations, along with visual examples, is available in the appendix.

\begin{table}[b!]
    \caption{
      \textbf{Robustness evaluation.} We evaluate the robustness of all methods under various attacks on image (SA-1b and Meta AI images) and video datasets (MovieGen and SA-V videos) using the original content resolution.
      In addition to the bit accuracy, we report negative $\logpval$ to account for the different number of bits encoded by each method. \ours{} constantly outperforms all related methods, especially in the case of difficult geometric and combined attacks. The image and video attacks differ; the full list of attacks is available in the appendix.
    }
    \label{tab:robustness-sa-1b-sa-v}
    \resizebox{\linewidth}{!}{\sisetup{
  table-align-uncertainty=true,
  separate-uncertainty=true,
}%
\renewrobustcmd{\bfseries}{\fontseries{b}\selectfont}%
\renewrobustcmd{\boldmath}{}%
\begin{tabular}{rl *{10}{S[table-format=3.1,detect-weight,mode=text]}}
\toprule
& & \multicolumn{2}{c}{\shortstack{Identity}} & \multicolumn{2}{c}{\shortstack{Valuemetric}} & \multicolumn{2}{c}{\shortstack{Compression}} & \multicolumn{2}{c}{\shortstack{Geometric}} & \multicolumn{2}{c}{\shortstack{Combined}} \\
\cmidrule(lr){3-4} \cmidrule(lr){5-6} \cmidrule(lr){7-8} \cmidrule(lr){9-10} \cmidrule(lr){11-12}
&  & \multicolumn{2}{c}{\scalebox{0.7}{{Bit~acc. ($\uparrow$)/$-\logpval$ ($\uparrow$)}}} & \multicolumn{2}{c}{\scalebox{0.7}{{Bit~acc. ($\uparrow$)/$-\logpval$ ($\uparrow$)}}} & \multicolumn{2}{c}{\scalebox{0.7}{{Bit~acc. ($\uparrow$)/$-\logpval$ ($\uparrow$)}}} & \multicolumn{2}{c}{\scalebox{0.7}{{Bit~acc. ($\uparrow$)/$-\logpval$ ($\uparrow$)}}} & \multicolumn{2}{c}{\scalebox{0.7}{{Bit~acc. ($\uparrow$)/$-\logpval$ ($\uparrow$)}}}\\
\midrule
\multirow{9}{*}{\rotatebox[origin=c]{90}{Meta AI (\textit{images})}}
& CIN                  & 1.00 &   9.0 & 0.85 &   7.5 & 1.00 &   9.0 & 0.52 &   0.7 & 0.50 &   0.4 \\
& InvisMark            & 1.00 &  28.3 & 0.89 &  22.0 & 0.98 &  26.1 & 0.77 &  14.8 & 0.94 &  22.1 \\
& MBRS                 & 1.00 & \bfseries  75.9 & 0.95 &  62.0 & 1.00 & \bfseries  75.2 & 0.52 &   3.6 & 0.50 &   0.4 \\
& TrustMark-P          & 1.00 &  29.6 & 0.89 &  21.2 & 0.98 &  27.1 
& 0.65 &   8.2 & 0.52 &   0.7 \\
& TrustMark-Q          & 1.00 &  30.0 & 0.97 &  27.1 & 1.00 &  29.8 & 0.67 &   9.4 & 0.54 &   0.9 \\
& Video Seal 0.0       & 0.97 &  25.0 & 0.84 &  16.7 & 0.93 &  20.9 & 0.80 &  13.3 & 0.74 &   7.4 \\
& Video Seal 1.0       & 0.98 &  69.4 & 0.95 &  60.7 & 0.95 &  60.6 & 0.83 &  42.9 & 0.69 &  16.0 \\
& WAM                  & 1.00 &   9.6 & 0.89 &   7.5 & 1.00 &   9.5 & 0.77 &   4.7 & 0.62 &   1.6 \\
& \ours{} (ours)       & 1.00 &  75.8 & 0.97 & \bfseries  68.1 & 0.98 &  68.9 & 0.95 & \bfseries  63.9 & 0.91 & \bfseries  50.3 \\
\midrule
\multirow{9}{*}{\rotatebox[origin=c]{90}{SA-1b (\textit{photos})}} 
& CIN                  & 1.00 &   9.0 & 0.85 &   7.5 & 1.00 &   9.0 & 0.52 &   0.7 & 0.50 &   0.4 \\
& InvisMark            & 1.00 &  28.3 & 0.89 &  21.4 & 1.00 &  27.8 & 0.77 &  14.1 & 0.98 &  25.7 \\
& MBRS                 & 0.99 &  71.8 & 0.93 &  55.9 
& 0.99 &  71.1 & 0.52 &   3.4 & 0.50 &   0.4 \\
& TrustMark-P          & 0.99 &  28.8 & 0.84 &  17.3 & 0.98 &  26.7 & 0.64 &   7.7 & 0.52 &   0.6 \\
& TrustMark-Q          & 1.00 &  29.9 & 0.96 &  25.9 & 1.00 &  29.7 & 0.65 &   8.5 & 0.53 &   0.8 \\
& Video Seal 0.0       & 0.98 &  25.2 & 0.85 &  17.2 & 0.96 &  23.3 & 0.79 &  12.6 & 0.74 &   7.6 \\
& Video Seal 1.0       & 0.99 &  72.8 & 0.96 &  64.6 & 0.99 &  71.0 & 0.82 &  41.8 & 0.70 &  17.1 \\
& WAM                  & 1.00 &   9.6 & 0.90 &   7.7 & 1.00 &   9.6 & 0.78 &   4.8 & 0.72 &   3.0 \\
& \ours{} (ours)       & 1.00 & \bfseries  75.4 & 0.98 & \bfseries  68.5 & 0.99 & \bfseries  73.6 & 0.93 & \bfseries  59.0 & 0.94 & \bfseries  55.6 \\
\midrule

\midrule
\multirow{4}{*}{\rotatebox[origin=c]{90}{MovieGen}}
& RivaGAN              & 0.93 &   7.2 & 0.83 &   5.1 & 0.81 &   4.5 & 0.58 &   1.0 & 0.56 &   0.7 \\
& Video Seal 0.0       & 0.98 &  26.3 & 0.88 &  20.0 & 0.86 &  16.7 & 0.82 &  14.4 & 0.70 &   6.9 \\
& Video Seal 1.0       & 1.00 &  76.1 & 1.00 &  74.9 & 0.89 & \bfseries  51.5 & 0.88 &  50.0 & 0.63 &  12.1 \\
& \ours{} (ours)       & 1.00 & \bfseries  76.9 & 1.00 & \bfseries  76.5 & 0.86 &  48.6 & 0.97 & \bfseries  68.0 & 0.70 & \bfseries  25.6 \\
\midrule
\multirow{4}{*}{\rotatebox[origin=c]{90}{SA-V}}
& RivaGAN              & 0.97 &   8.2 & 0.87 &   6.1 & 0.82 &   4.7 & 0.60 &   1.3 & 0.56 &   0.7 \\
& Video Seal 0.0       & 0.99 &  27.3 & 0.89 &  21.1 & 0.81 &  14.6 & 0.83 &  15.7 & 0.64 &   4.5 \\
& Video Seal 1.0       & 1.00 &  76.6 & 1.00 &  76.0 & 0.87 & \bfseries  49.4 & 0.87 &  50.4 & 0.59 &   8.0 \\
& \ours{} (ours)       & 1.00 & \bfseries  77.1 & 1.00 & \bfseries  76.8 & 0.86 &  48.1 & 0.98 & \bfseries  70.8 & 0.68 & \bfseries  24.2 \\
\bottomrule
\end{tabular}%
}
\end{table}

\subsection{Comparison with the state-of-the-art}\label{sec:image-baselines}

In this section, we evaluate the robustness and imperceptibility of all image watermarking methods. 
Additionally, we also compare the robustness of \ours{} with other methods in video watermarking. For evaluation on videos, \ours{} uses temporal watermark pooling with step size $k=4$ and depth $n=2$.

\paragraphcustom{Robustness.} For image watermarking, the results in Table~\ref{tab:robustness-sa-1b-sa-v} for both Meta AI-generated images and real SA-1b photos show that all methods perform well against simple valuemetric attacks, such as a small brightness change, and image compression algorithms, such as JPEG. However, for geometric attacks, such as cropping, we see \ours{} clearly outperforming all other methods. The performance difference is further amplified in combined attacks consisting of cropping, JPEG compression, and brightness change, where only \ours{} and InvisMark maintain bit accuracy above 90\% while \ours{} embeds a significantly larger message than InvisMark (256 vs. 94 bits). In video watermarking, compression algorithms such as H.264 and HEVC present a significantly greater challenge, resulting in lower bit recovery rates across all methods, especially in the combined attack where JPEG image compression is replaced by H.264 video compression. Nonetheless, \ours{} still outperforms all related methods.

\paragraphcustom{Imperceptibility.}
We evaluate watermark imperceptibility both quantitatively and qualitatively. 
The quantitative evaluation of the images generated by Meta AI is presented in Table~\ref{tab:eval-quality}. We can see Video Seal 0.0 substantially outperforms all other methods in the PSNR and SSIM metrics. However, this is due to its very localized artifacts, which yield, on average, a high metric value but are very visible. This can be seen in Figure~\ref{fig:example_comp} and is further confirmed by other quality metrics that perform more complex aggregation than just averaging. Overall, across all quality metrics, only \ours{} and TrustMark-P rank among the top three methods for 4 out of 5 metrics. However, TrustMark-P is significantly less robust than \ours{}.
Strong imperceptibility of \ours{} watermarks is further confirmed by the qualitative results in Figure~\ref{fig:example_comp}. \ours{} produces watermarks that are localized to the edges of objects in the image and are of small magnitude. In contrast, for example, WAM produces a very localized watermark, but its magnitude is much larger, or InvisMark produces watermarks of a small magnitude but not well localized.
Additional qualitative results are available in the appendix.

\begin{table}
    \caption{
      \textbf{Quantitative evaluation of the imperceptibility.} 
      We report the visual quality metrics on the 1000 images generated by Meta AI. Only \ours{} and TrustMark-P rank among the top three methods for 4 out of 5 metrics. However, TrustMark-P is significantly less robust than \ours{}.
    }\vspace{-0.2cm}
    \label{tab:eval-quality}
    \centering
    \sisetup{
  table-align-uncertainty=true,
  separate-uncertainty=true,
}
\renewrobustcmd{\bfseries}{\fontseries{b}\selectfont}
\renewrobustcmd{\boldmath}{}

\begin{tabular}{l S[table-format=2.1,detect-weight,mode=text]S[table-format=1.4,detect-weight,mode=text]S[table-format=1.2,detect-weight,mode=text]S[table-format=1.4,detect-weight,mode=text]S[table-format=1.2,detect-weight,mode=text]}

\toprule
& \multicolumn{1}{c}{PSNR ($\uparrow$)} & \multicolumn{1}{c}{SSIM ($\uparrow$)} & \multicolumn{1}{c}{CVVDP ($\uparrow$)} & \multicolumn{1}{c}{LPIPS ($\downarrow$)} & \multicolumn{1}{c}{JND ($\downarrow$)} \\
\midrule
CIN                   &  44.3 & 0.9894 &  9.51 & 0.0274 &  1.93 \\
InvisMark             &  46.6 & 0.9823 &  \underline{9.93} & 0.0020 &  \underline{0.26} \\
MBRS                  &  46.3 & \underline{0.9928} &  9.54 & 0.0049 &  1.59 \\
TrustMark-P           &  \underline{49.0} & 0.9908 &  \bfseries 9.94 & \underline{0.0018} &  \bfseries 0.19 \\
TrustMark-Q           &  43.3 & 0.9886 &  9.73 & 0.0022 &  1.66 \\
Video Seal 0.0        &  \bfseries 53.7 & \bfseries 0.9989 &  9.85 & 0.0039 &  1.07 \\
Video Seal 1.0        &  48.7 & \underline{0.9958} &  9.84 & \bfseries 0.0012 &  0.45 \\
WAM                   &  41.4 & 0.9760 &  9.72 & 0.0344 &  0.95 \\
\ours{} (ours)        &  \underline{48.9} & 0.9905 &  \underline{9.87} & \underline{0.0013} &  \underline{0.34} \\
\bottomrule
\end{tabular}

\end{table}

\subsection{Ablations}
We ablate all key components introduced in Section~\ref{sec:method}: adversarial-only loss function with watermark boosting, three-stage training schedule, high-resolution training, and temporal watermark pooling. The main ablation results are shown in Table~\ref{tab:ablations} and are detailed next.

\paragraphcustom{Adversarial-only loss function.} Unlike related work, our \ours{} model is trained using only adversarial loss (Equation~\eqref{eq:adv_loss}) and message loss (Equation~\eqref{eq:bce}). To justify our choice of loss function, we also train our model using a mean-squared error perceptual loss, a common approach in related work. In Table~\ref{tab:ablations} (section \textbf{(a)}), we show the results for training runs with different perceptual loss weights $\lambda_{perc}$. Training with a small $\lambda_{perc}=0.1$ has only a small effect on the final results in terms of both watermark robustness and imperceptibility. When the perceptual loss weight is increased to $\lambda_{perc}=1.0$, we observe that the model is no longer able to learn any watermark, yielding a random bit accuracy of 50\%. We stop this experiment after 100 epochs of no change in bit accuracy and therefore do not report quality metrics.

To show the effectiveness of removing visible artifacts using the discriminator, we also train our model without the adversarial loss ($\lambda_{adv}=0$). From the visual quality metrics in Table~\ref{tab:ablations} (section \textbf{(a)}), it is clear that the discriminator guides \ours{} to produce much less visible visual artifacts. Similarly, the JND map limits the locations of watermark artifacts in the image to areas of high pixel variance (edges), yielding a much less visible watermark.

\paragraphcustom{Watermark boosting.} To control the imperceptibility of the watermark, we vary the watermark boosting parameter $\beta$. With $\beta > 1$, the watermark is artificially amplified for the discriminator, making it easier to detect and remove. In Table~\ref{tab:ablations} (section \textbf{(b)}), we verify that selecting $\beta=2.5$ significantly reduces visibility of artifacts while slightly decreasing the model's robustness. Setting $\beta=0.5$ yields the opposite results, i.e., more visible artifacts with improved robustness.

\paragraphcustom{Three-stage training schedule.} To stabilize the training process, \ours{} is trained with a three-stage schedule that allows the model to first learn how to produce robust watermarks. Only later is the model guided to make these robust watermarks less visible. 
In Table~\ref{tab:ablations} (section \textbf{(c)}), we show the results when the model is trained with the final watermark scaling factor $\alpha=0.2$ from the beginning, as well as when the model is trained with the adversarial loss from the beginning. In both cases, the model is unable to learn any watermark, resulting in a random bit accuracy of 50\%. We stop this experiment after 100 epochs of no change in bit accuracy and therefore do not report quality metrics.

\paragraphcustom{High-resolution training.} During training, we compute the JND map, apply the discriminator, and perform all attacks in high resolution instead of using the low 256$\times$256 resolution of the watermarking models. As shown in Table~\ref{tab:ablations} (section \textbf{(d)}), the high-resolution training outperforms the baseline approach in robustness for all types of attacks. Additional improvement is achieved in geometric and combined attacks due to the variable image aspect ratio. We also observe a reduction in visible watermark artifacts due to the application of the discriminator in high resolution.

\paragraphcustom{Temporal watermark pooling.} For efficient watermarking of videos using \ours{}, we insert a temporal average pooling layer into the embedder, reducing the size of the internal video feature representations. We measure the robustness, imperceptibility, and relative speedup of pooling with a kernel size of $k=4$ at various depths $d$. 
In Table~\ref{tab:ablations-temppool}, we see that inserting the temporal average pooling layer after the first U-Net block ($d=1$) produces a speedup of 2.2$\times$ compared to baseline, while having minimal impact on performance. 
The inference speed was measured on the Quadro GP100 GPU using 64 video frames and a vanilla PyTorch implementation of the model.

\begin{table}[t!]
    \caption{
      \textbf{Ablations of the key \ours{} improvements.} The results show our key contributions (adversarial-only loss function, three-stage training schedule, and high-resolution adaptation) significantly contribute to the state-of-the-art performance of \ours{}. A large drop in the metrics is indicated in red. The results are reported on the SA-1b validation set of 100 photos.
    }
    \label{tab:ablations}
    \centering
    \resizebox{\linewidth}{!}{\sisetup{
  table-align-uncertainty=true,
  separate-uncertainty=true,
}%
\renewrobustcmd{\bfseries}{\fontseries{b}\selectfont}%
\renewrobustcmd{\boldmath}{}%
\begin{tabular}{c@{~~}l S[table-format=1.2,detect-weight,mode=text]S[table-format=1.2,detect-weight,mode=text]S[table-format=1.2,detect-weight,mode=text]S[table-format=1.2,detect-weight,mode=text]S[table-format=1.2,detect-weight,mode=text]S[table-format=2.1,detect-weight,mode=text]S[table-format=1.2,detect-weight,mode=text]}
\toprule
& & \multicolumn{1}{c}{Identity} & \multicolumn{1}{c}{Valuemetric} & \multicolumn{1}{c}{Compression} & \multicolumn{1}{c}{Geometric} & \multicolumn{1}{c}{Combined} & \multicolumn{1}{c}{\multirow{2}{*}{PSNR ($\uparrow$)}} & \multicolumn{1}{c}{\multirow{2}{*}{JND ($\downarrow$)}} \\
\cmidrule(lr){3-3} \cmidrule(lr){4-4} \cmidrule(lr){5-5} \cmidrule(lr){6-6} \cmidrule(lr){7-7} 
& & \multicolumn{1}{c}{\scalebox{0.7}{{Bit~acc. ($\uparrow$)}}} & \multicolumn{1}{c}{\scalebox{0.7}{{Bit~acc. ($\uparrow$)}}} & \multicolumn{1}{c}{\scalebox{0.7}{{Bit~acc. ($\uparrow$)}}} & \multicolumn{1}{c}{\scalebox{0.7}{{Bit~acc. ($\uparrow$)}}} & \multicolumn{1}{c}{\scalebox{0.7}{{Bit~acc. ($\uparrow$)}}}\\
\midrule
\multirow{5}{*}{\rotatebox[origin=c]{90}{\textbf{(a)}}}
& \ours{}                                                               & 1.00 & 0.98 & 0.99 & 0.93 & 0.94 &  47.0 &  0.44 \\
& w/ perceptual loss ($\mathcal{L}_{perc}=L_2^2$, $\lambda_{perc}=0.1$) & 0.99 & 0.97 & 0.99 & 0.92 & 0.91 &  47.1 &  0.41 \\
& w/ perceptual loss ($\mathcal{L}_{perc}=L_2^2$, $\lambda_{perc}=1.0$) & \textcolor{red}{0.50} & \textcolor{red}{0.50} & \textcolor{red}{0.50} & \textcolor{red}{0.50} & \textcolor{red}{0.50} &  \multicolumn{1}{c}{--} &  \multicolumn{1}{c}{--} \\
& w/o discriminator ($\lambda_{adv}=0$)                                 & 1.00 & 0.99 & 1.00 & 0.95 & 0.97 &  \textcolor{red}{44.9} &  \textcolor{red}{0.96} \\
& w/o JND                                                               & 1.00 & 0.98 & 1.00 & 0.93 & 0.97 &  \textcolor{red}{43.1} &  \textcolor{red}{1.34} \\
\midrule
\multirow{3}{*}{\rotatebox[origin=c]{90}{\textbf{(b)}}}
& $\beta=0.5$             & 1.00 & 0.98 & 1.00 & 0.95 & 0.96 &  46.2 &  0.56 \\
& \ours{} ($\beta=1$)     & 1.00 & 0.98 & 0.99 & 0.93 & 0.94 &  47.0 &  0.44 \\
& $\beta=2.5$             & 0.99 & 0.96 & 0.98 & 0.90 & 0.86 &  50.1 &  0.30 \\
\midrule
\multirow{3}{*}{\rotatebox[origin=c]{90}{\textbf{(c)}}}
& \ours{}                                         & 1.00 & 0.98 & 0.99 & 0.93 & 0.94 &  47.0 &  0.44 \\
& w/o watermark scaling ($\alpha_0=\alpha_1=0.2$) & \textcolor{red}{0.50} & \textcolor{red}{0.50} & \textcolor{red}{0.50} & \textcolor{red}{0.50} & \textcolor{red}{0.50} &  \multicolumn{1}{c}{--} &  \multicolumn{1}{c}{--} \\
& w/o discriminator delay                         & \textcolor{red}{0.50} & \textcolor{red}{0.50} & \textcolor{red}{0.50} & \textcolor{red}{0.50} & \textcolor{red}{0.50} &  \multicolumn{1}{c}{--} &  \multicolumn{1}{c}{--} \\
\midrule
\multirow{2}{*}{\rotatebox[origin=c]{90}{\textbf{(d)}}}
& \ours{}                           & 1.00 & 0.98 & 0.99 & 0.93 & 0.94 &  47.0 &  0.44 \\
& fixed 256$\times$256 resolution   & 0.97 & 0.94 & 0.96 & \textcolor{red}{0.81} & \textcolor{red}{0.68} &  46.8 &  0.45 \\
\bottomrule
\end{tabular}%
}
\end{table}

\begin{table}[t!]
    \caption{
      \textbf{Ablation of the temporal watermark pooling parameters.} Inserting the temporal average pooling layer after the first U-Net block ($d=1$) produces a speedup of 2.2x compared to baseline \ours{}. The results are reported on the SA-V validation set of 96 videos.
    }
    \label{tab:ablations-temppool}
    \centering
    \resizebox{\linewidth}{!}{\sisetup{
  table-align-uncertainty=true,
  separate-uncertainty=true,
}%
\renewrobustcmd{\bfseries}{\fontseries{b}\selectfont}%
\renewrobustcmd{\boldmath}{}%
\begin{tabular}{l S[table-format=1.2,detect-weight,mode=text]S[table-format=1.2,detect-weight,mode=text]S[table-format=1.2,detect-weight,mode=text]S[table-format=1.2,detect-weight,mode=text]S[table-format=1.2,detect-weight,mode=text]S[table-format=2.1,detect-weight,mode=text]c}
\toprule
& \multicolumn{1}{c}{Identity} & \multicolumn{1}{c}{Valuemetric} & \multicolumn{1}{c}{Compression} & \multicolumn{1}{c}{Geometric} & \multicolumn{1}{c}{Combined} & \multicolumn{1}{c}{\multirow{2}{*}{PSNR ($\uparrow$)}} & \multicolumn{1}{c}{\multirow{2}{*}{Speedup ($\uparrow$)}} \\
\cmidrule(lr){2-2} \cmidrule(lr){3-3} \cmidrule(lr){4-4} \cmidrule(lr){5-5} \cmidrule(lr){6-6}
& \multicolumn{1}{c}{\scalebox{0.7}{{Bit~acc. ($\uparrow$)}}} & \multicolumn{1}{c}{\scalebox{0.7}{{Bit~acc. ($\uparrow$)}}} & \multicolumn{1}{c}{\scalebox{0.7}{{Bit~acc. ($\uparrow$)}}} & \multicolumn{1}{c}{\scalebox{0.7}{{Bit~acc. ($\uparrow$)}}} & \multicolumn{1}{c}{\scalebox{0.7}{{Bit~acc. ($\uparrow$)}}}\\
\midrule
\ours{} ($k=1$)           & 1.00 & 1.00 & 0.86 & 0.98 & 0.68 &  47.6 & -- \\
w/ temporal pooling ($k=4$, $d=3$) & 1.00 & 1.00 & 0.85 & 0.98 & 0.69 &  47.7 & 1.0x \\
w/ temporal pooling ($k=4$, $d=2$) & 1.00 & 1.00 & 0.85 & 0.98 & 0.69 &  47.6 & 1.7x \\
w/ temporal pooling ($k=4$, $d=1$) & 1.00 & 1.00 & 0.84 & 0.98 & 0.66 &  47.7 & 2.2x \\
\bottomrule
\end{tabular}%
}
\end{table}

\section{Conclusion}
In this work, we present \ours, a state-of-the-art image watermarking model both in terms of robustness and imperceptibility. The imperceptibility of the generated watermarks is achieved through a novel adversarial-only approach that utilizes watermark boosting without relying on perceptual losses. Further, due to our three-stage training with high-resolution adaptation, the training process is stable even under heavy augmentations, resulting in extremely robust watermarks.
Finally, we introduce an inference-time technique for adapting image watermarking models to video, achieving a substantial speedup without compromising the model's performance.
We release \ours{} model weights to foster further research in this field.

\clearpage
\bibliographystyle{tmlr}
\bibliography{main}

\clearpage
\appendix

\begin{figure*}
    \centering
    \footnotesize
    \newcommand{\imwidth}{0.19\textwidth}
    \setlength{\tabcolsep}{0pt}
    \resizebox{1.0\linewidth}{!}{
    \begin{tabular}{c@{\hskip 2pt}c@{\hskip 2pt}c@{\hskip 2pt}c@{\hskip 2pt}c}
\textit{Original} & CIN & InvisMark & MBRS & TrustMark-P \\
\includegraphics[width=\imwidth]{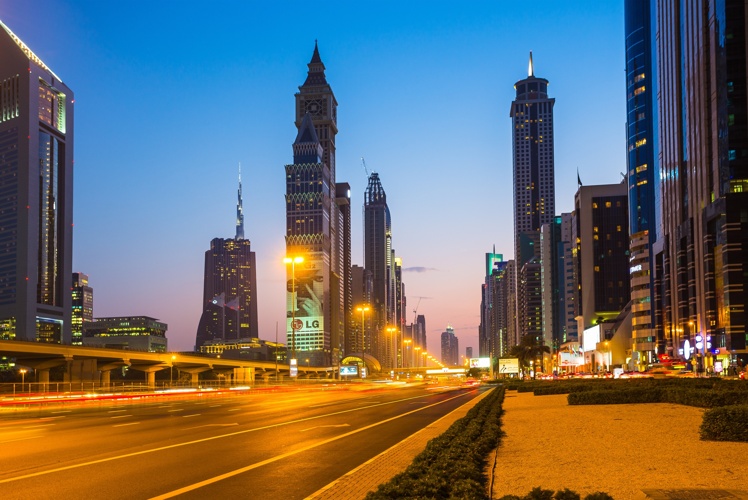} &
\includegraphics[width=\imwidth]{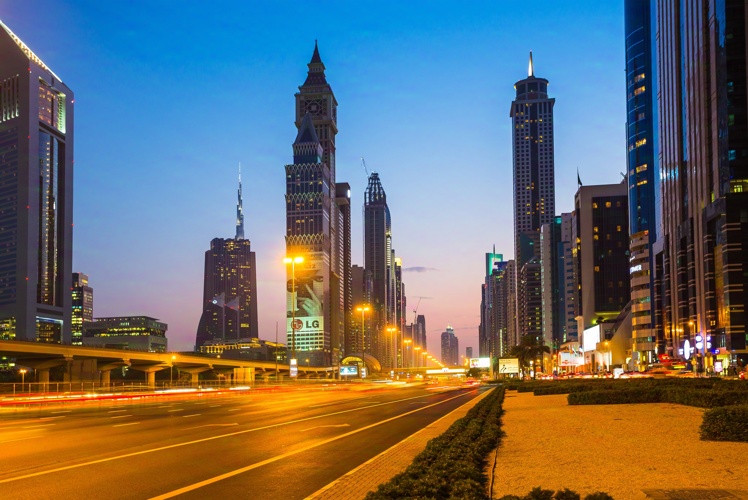} &
\includegraphics[width=\imwidth]{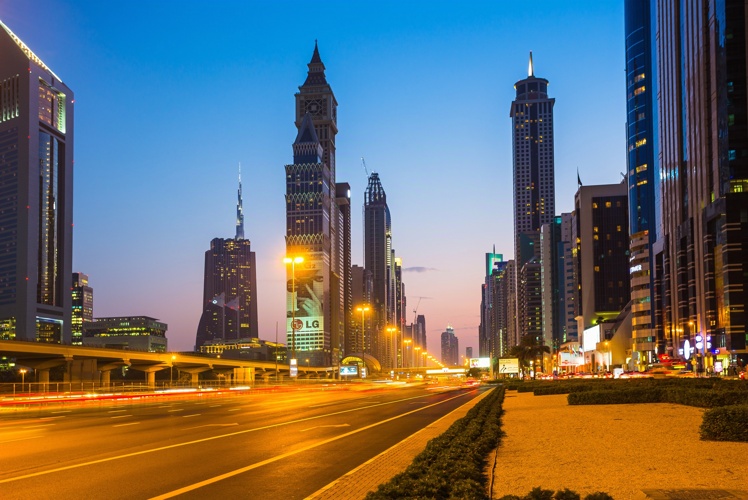} &
\includegraphics[width=\imwidth]{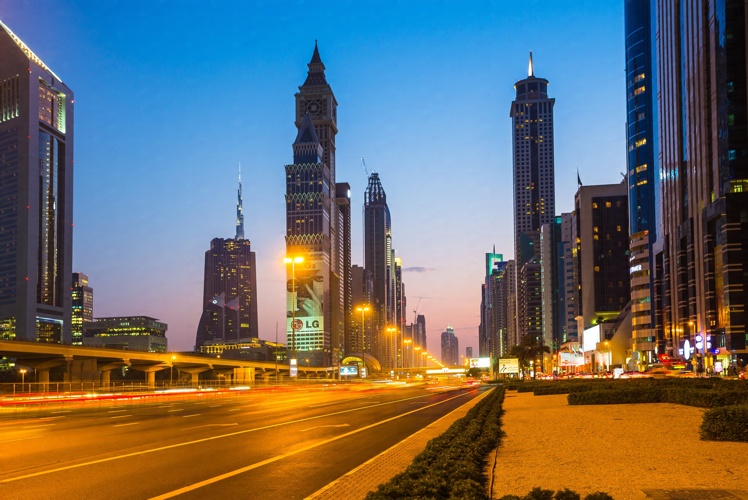} &
\includegraphics[width=\imwidth]{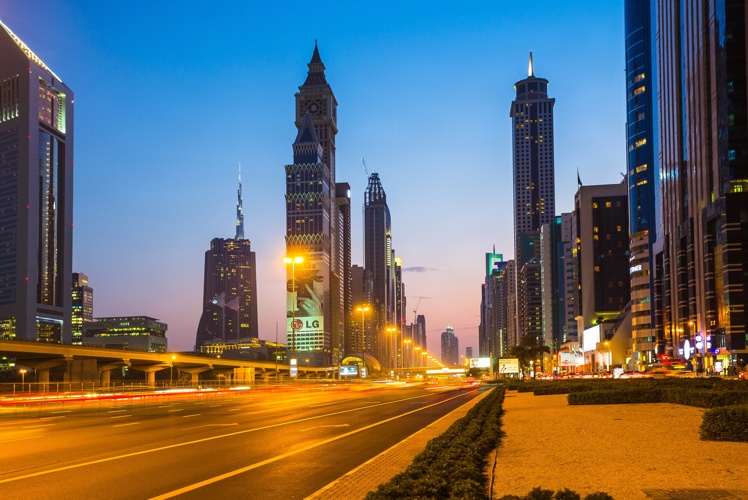} \\
& \includegraphics[width=\imwidth]{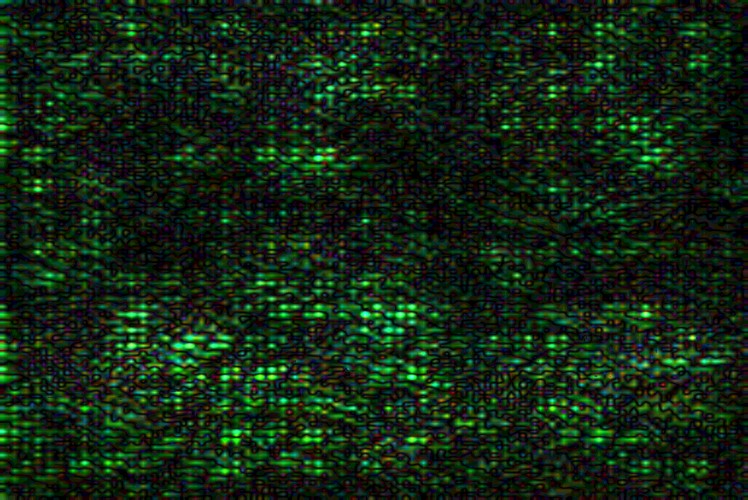} &
\includegraphics[width=\imwidth]{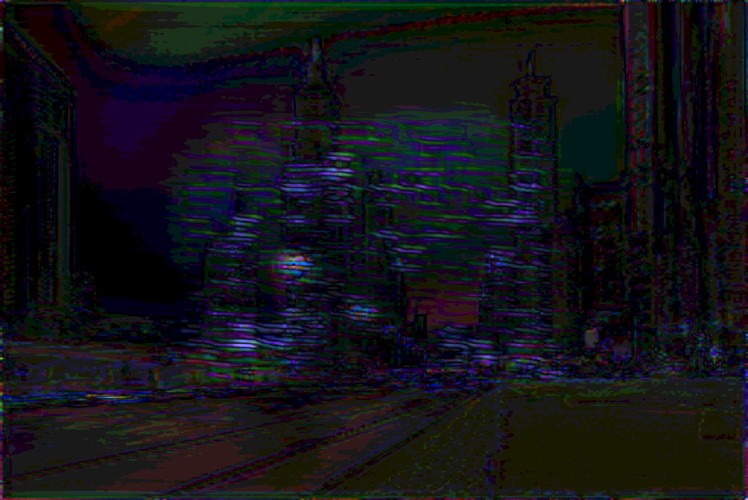} &
\includegraphics[width=\imwidth]{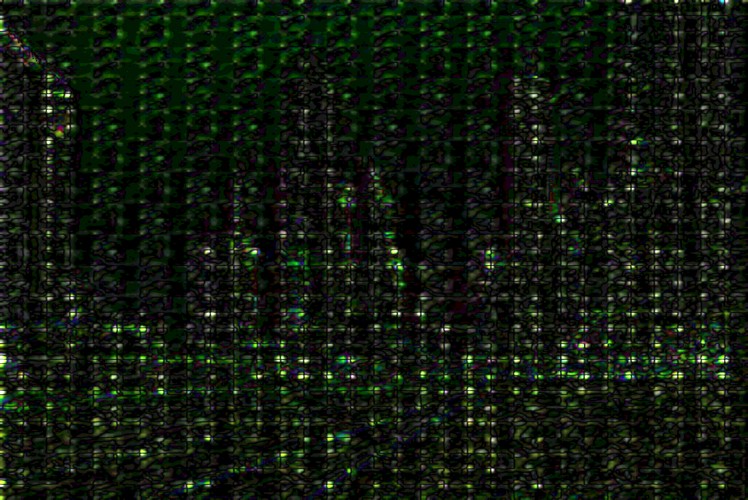} &
\includegraphics[width=\imwidth]{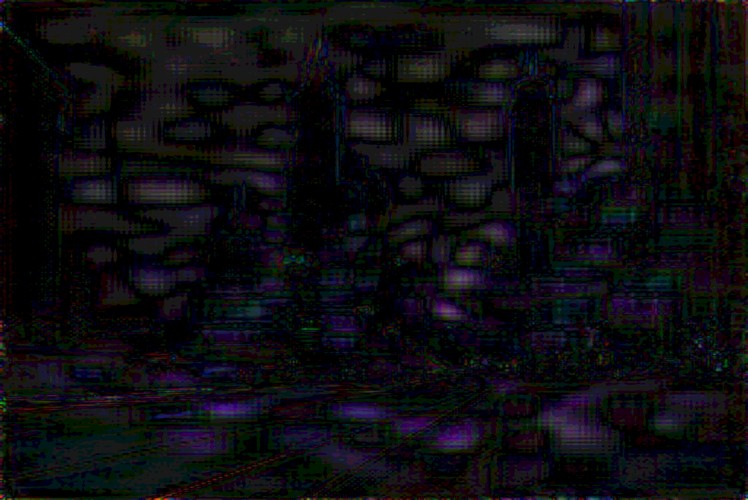}\\
\end{tabular}
}
\vspace{3pt}
\resizebox{1.0\linewidth}{!}{
\begin{tabular}{c@{\hskip 2pt}c@{\hskip 2pt}c@{\hskip 2pt}c@{\hskip 2pt}c}
TrustMark-Q & Video Seal 0.0 & Video Seal 1.0 & WAM & \ours{} (ours) \\
\includegraphics[width=\imwidth]{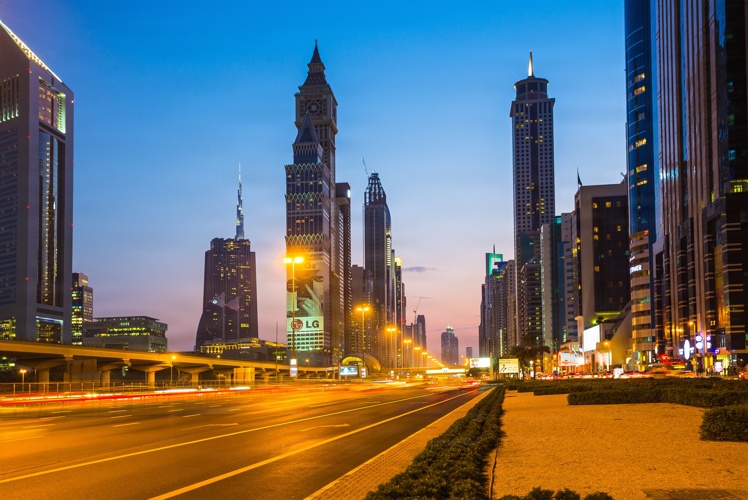} &
\includegraphics[width=\imwidth]{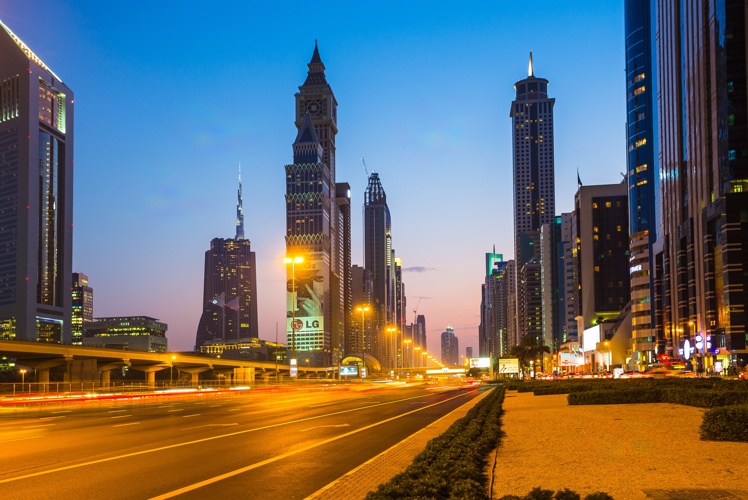} &
\includegraphics[width=\imwidth]{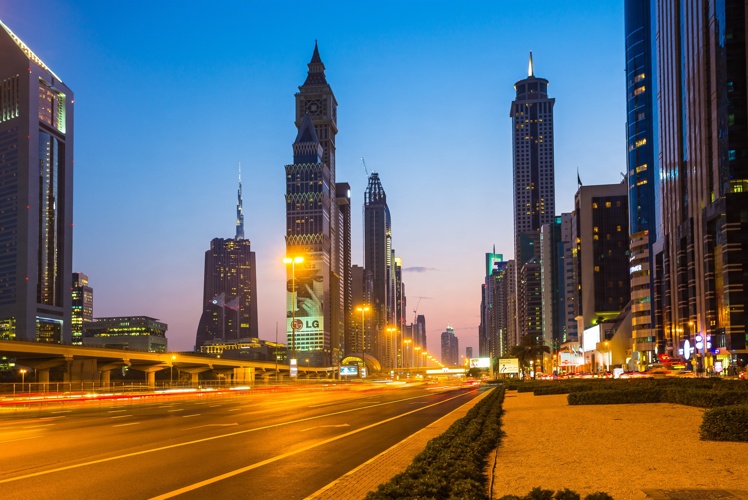} &
\includegraphics[width=\imwidth]{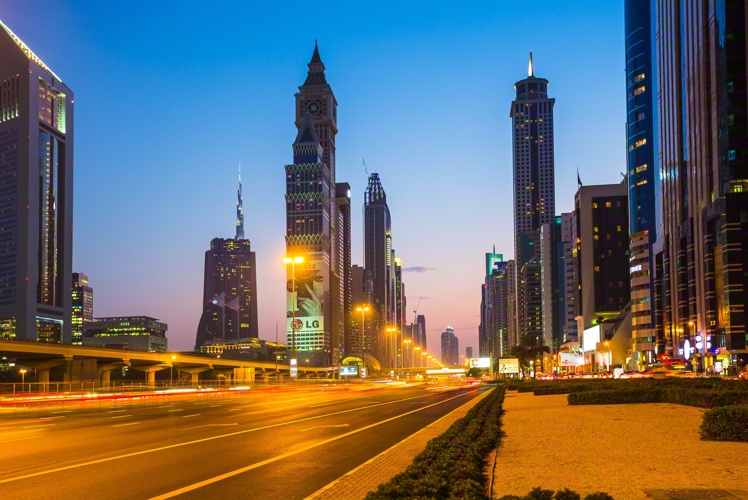} &
\includegraphics[width=\imwidth]{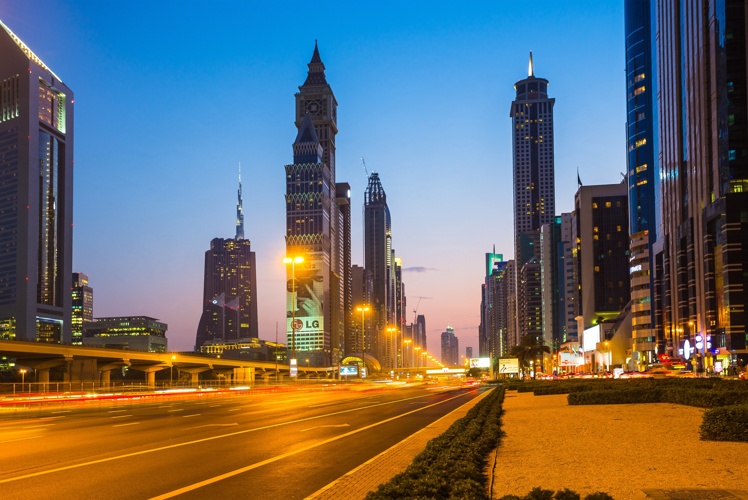} \\
\includegraphics[width=\imwidth]{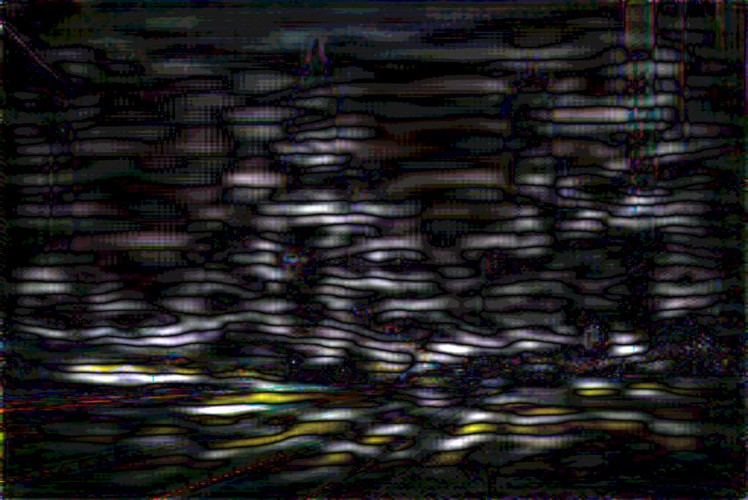} &
\includegraphics[width=\imwidth]{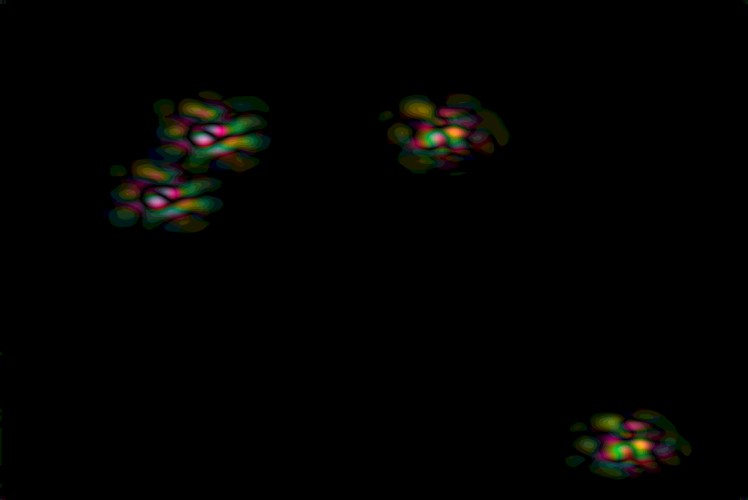} &
\includegraphics[width=\imwidth]{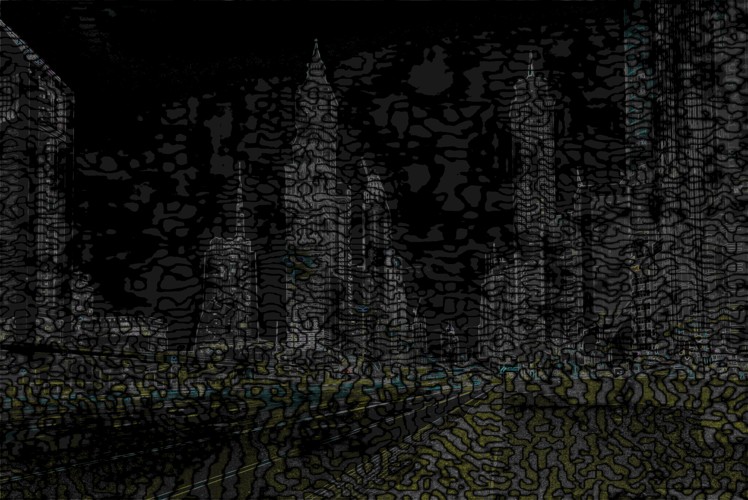} &
\includegraphics[width=\imwidth]{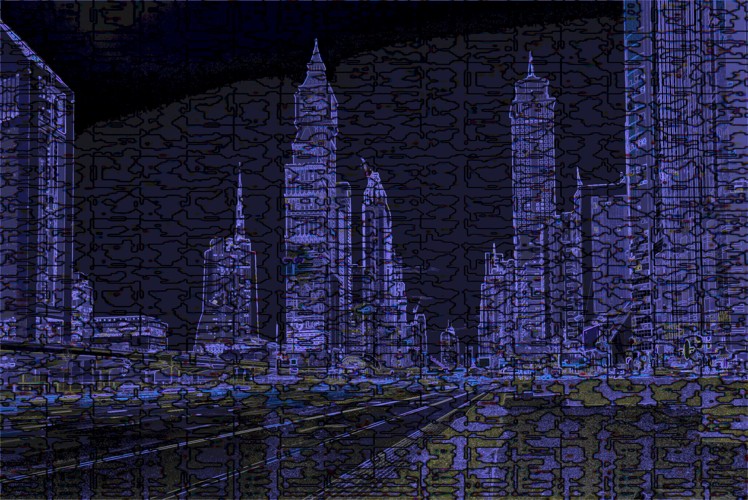} &
\includegraphics[width=\imwidth]{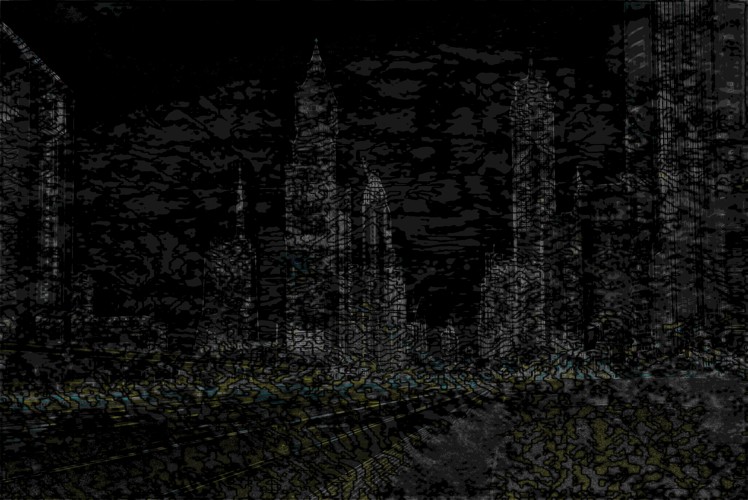} \\
    \end{tabular}
    }

\vspace{0.7cm}

    \resizebox{1.0\linewidth}{!}{
    \begin{tabular}{c@{\hskip 2pt}c@{\hskip 2pt}c@{\hskip 2pt}c@{\hskip 2pt}c}
\textit{Original} & CIN & InvisMark & MBRS & TrustMark-P \\
\includegraphics[width=\imwidth]{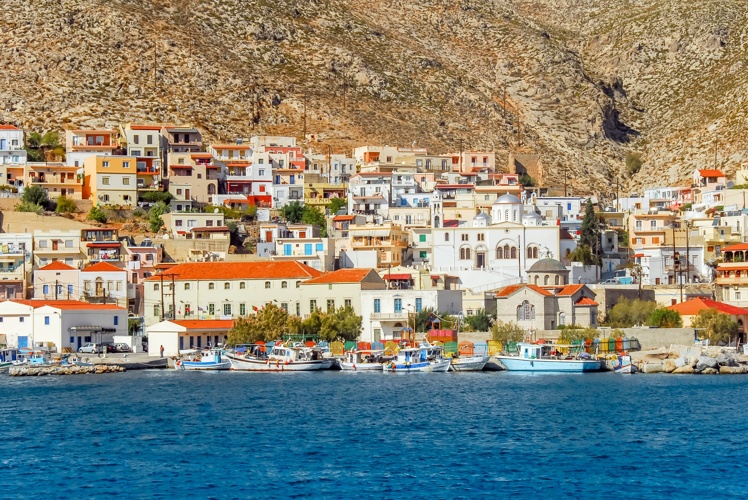} &
\includegraphics[width=\imwidth]{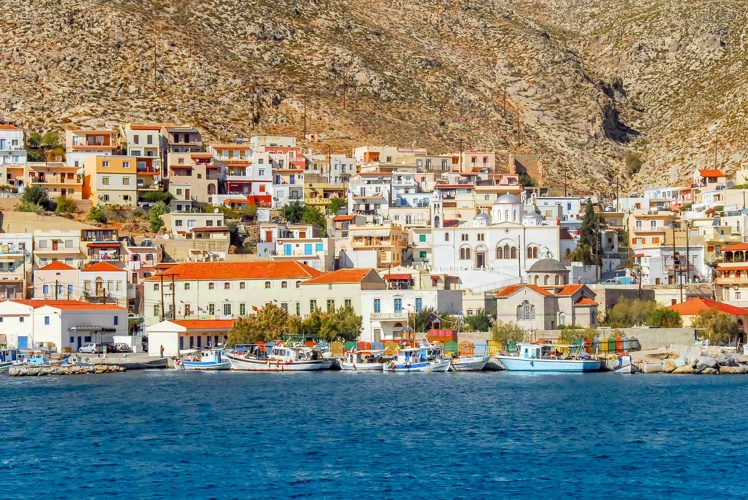} &
\includegraphics[width=\imwidth]{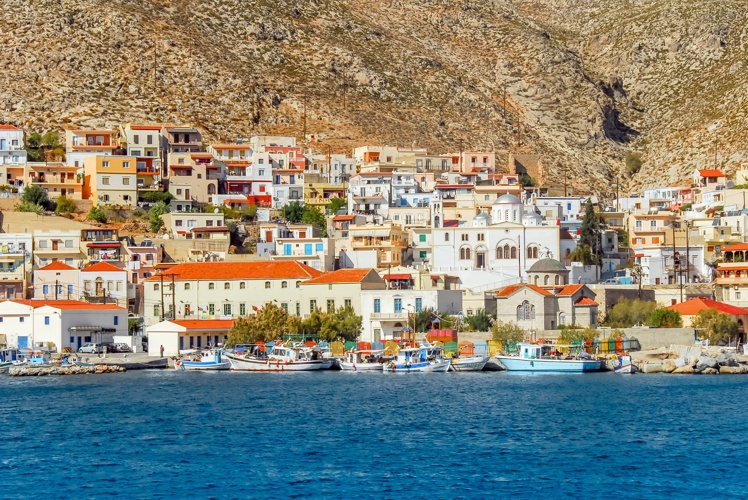} &
\includegraphics[width=\imwidth]{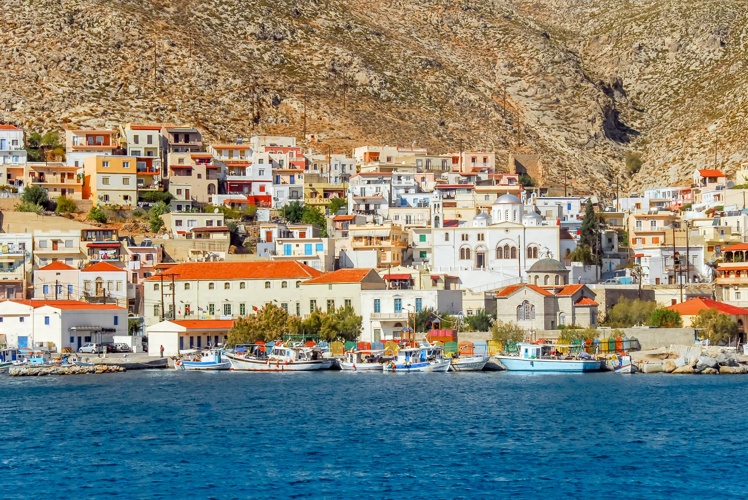} &
\includegraphics[width=\imwidth]{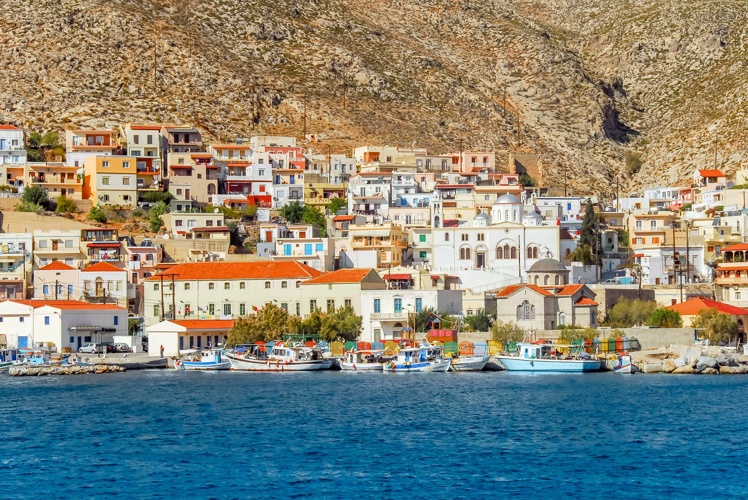} \\
& \includegraphics[width=\imwidth]{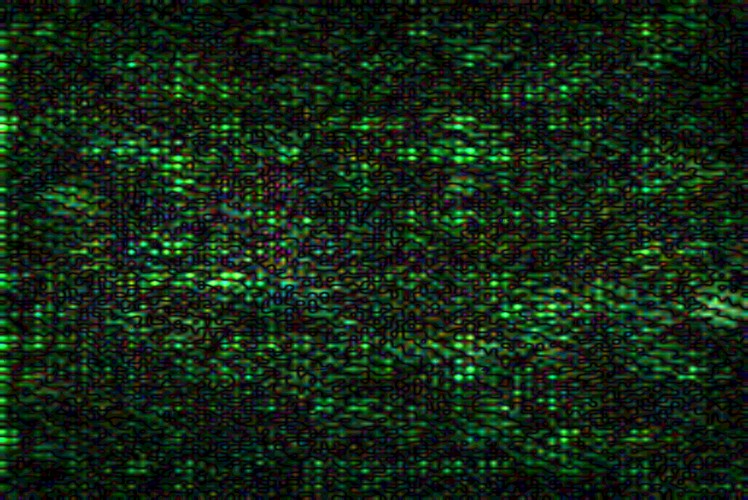} &
\includegraphics[width=\imwidth]{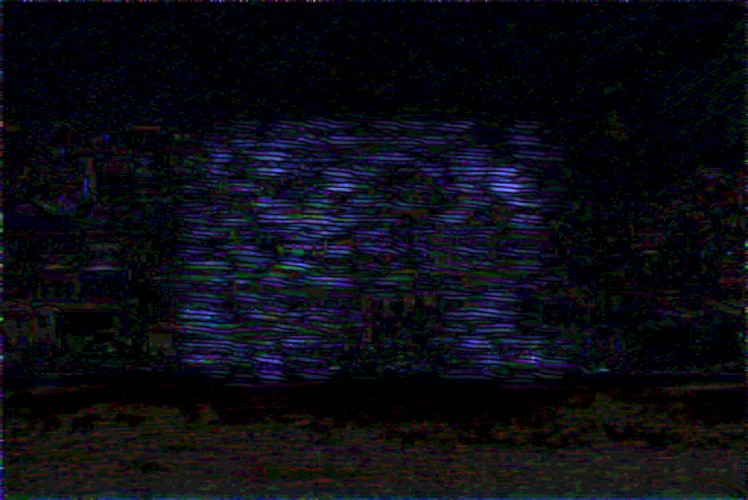} &
\includegraphics[width=\imwidth]{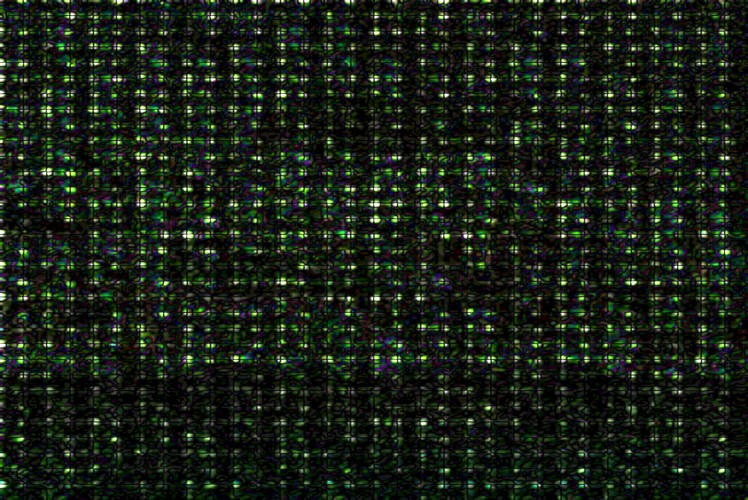} &
\includegraphics[width=\imwidth]{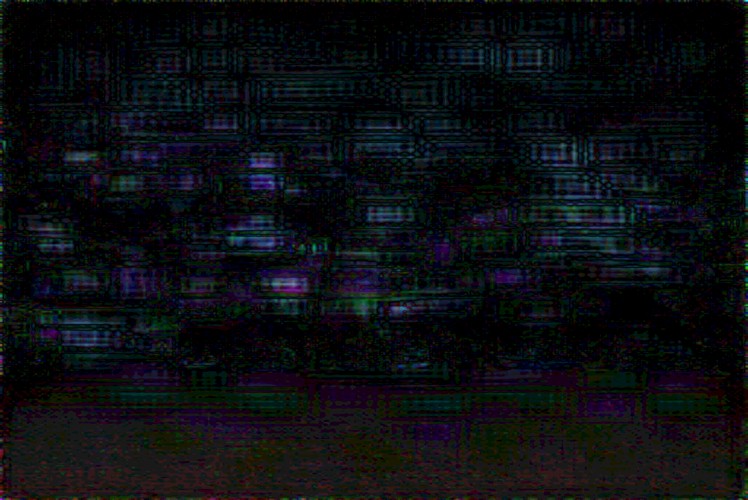}\\
\end{tabular}
}
\vspace{3pt}
\resizebox{1.0\linewidth}{!}{
\begin{tabular}{c@{\hskip 2pt}c@{\hskip 2pt}c@{\hskip 2pt}c@{\hskip 2pt}c}
TrustMark-Q & Video Seal 0.0 & Video Seal 1.0 & WAM & \ours{} (ours) \\
\includegraphics[width=\imwidth]{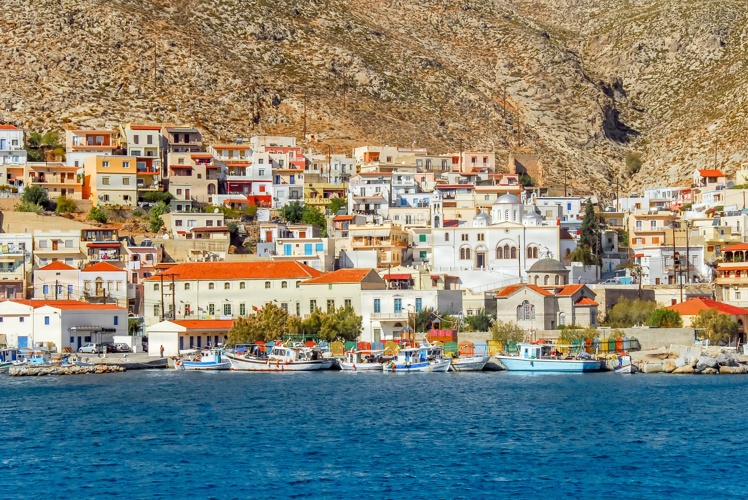} &
\includegraphics[width=\imwidth]{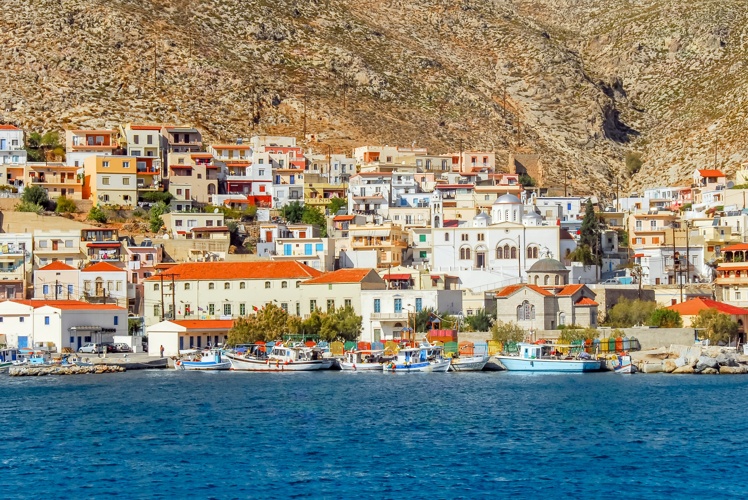} &
\includegraphics[width=\imwidth]{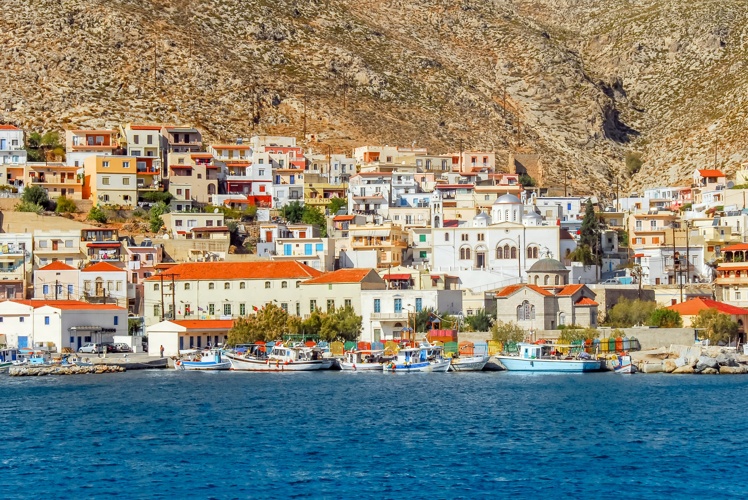} &
\includegraphics[width=\imwidth]{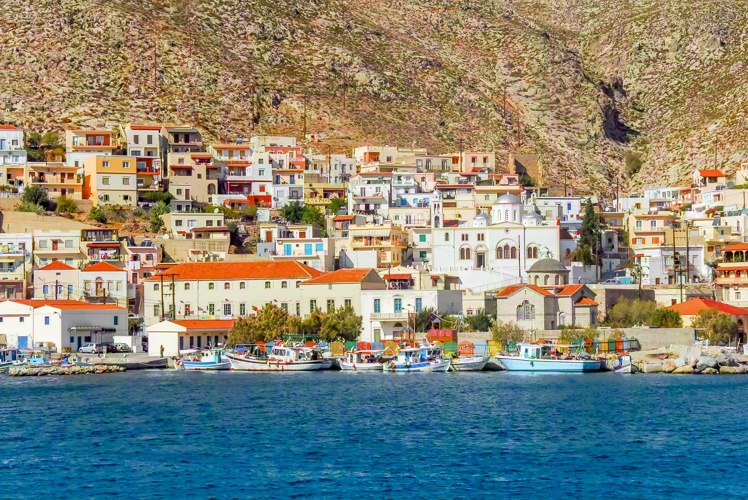} &
\includegraphics[width=\imwidth]{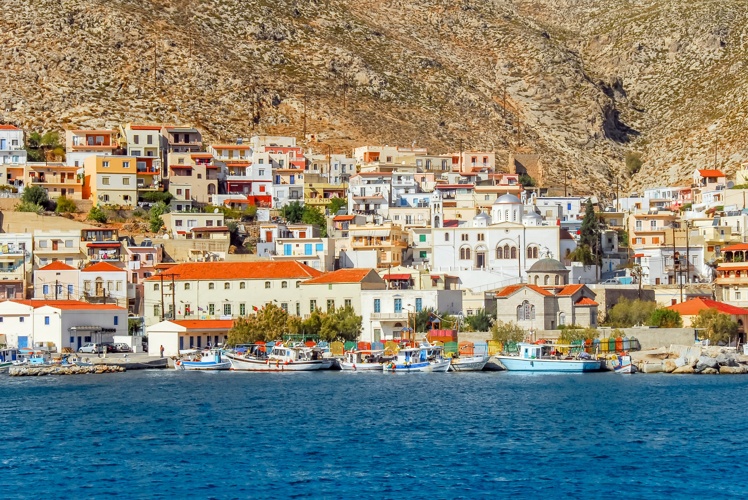} \\
\includegraphics[width=\imwidth]{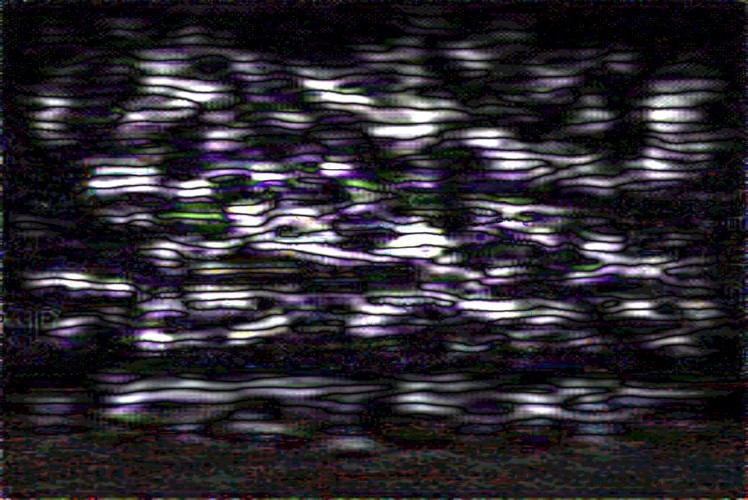} &
\includegraphics[width=\imwidth]{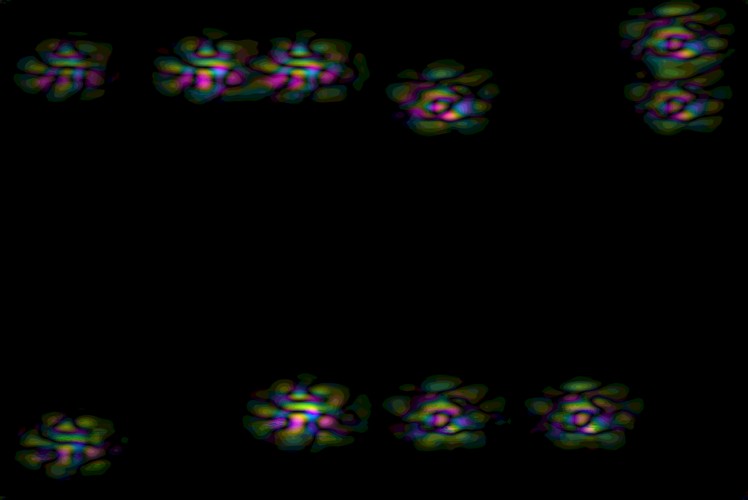} &
\includegraphics[width=\imwidth]{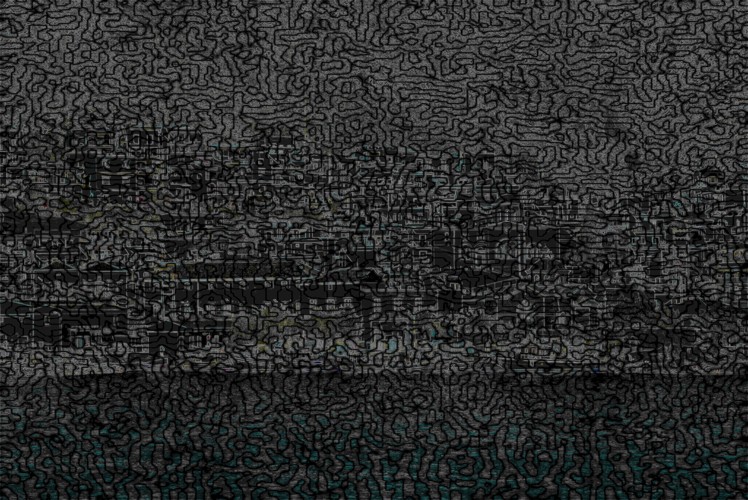} &
\includegraphics[width=\imwidth]{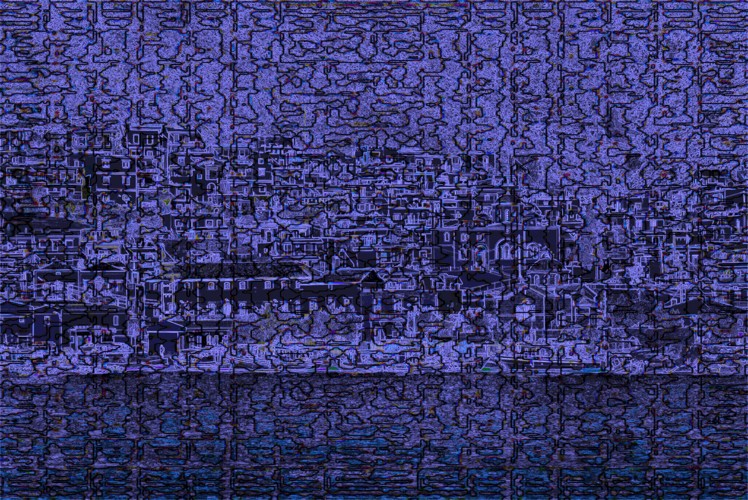} &
\includegraphics[width=\imwidth]{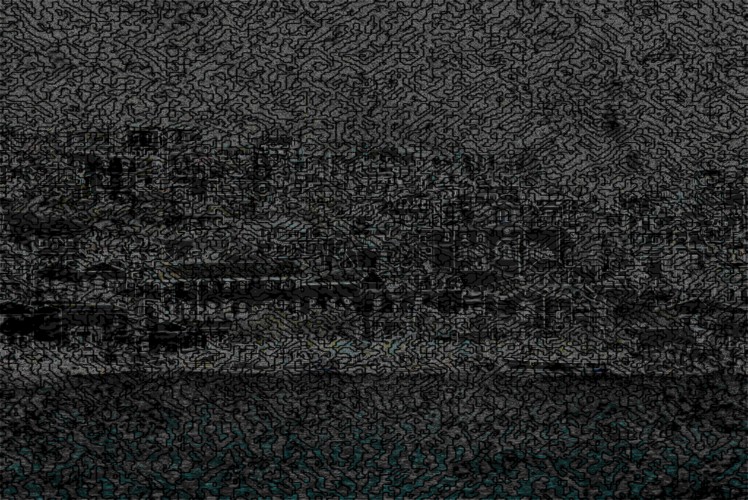} \\
    \end{tabular}
    }

    \caption{\textbf{Comparison with related work on photos from the SA-1b dataset.} We show both the watermarked image (top) and the predicted watermark brightened for clarity (bottom).
    Many related methods leave visible artifacts in areas with a single color. In contrast, \ours\ does not leave visible artifacts in such areas while being more robust to various transformations.
}\label{appendix:fig:example_comp2} 
\end{figure*}

\begin{figure*}
    \centering
    \footnotesize
    \newcommand{\imwidth}{0.19\textwidth}
    \setlength{\tabcolsep}{0pt}
    \resizebox{1.0\linewidth}{!}{
    \begin{tabular}{c@{\hskip 2pt}c@{\hskip 2pt}c@{\hskip 2pt}c@{\hskip 2pt}c}
\textit{Original} & CIN & InvisMark & MBRS & TrustMark-P \\
\includegraphics[width=\imwidth]{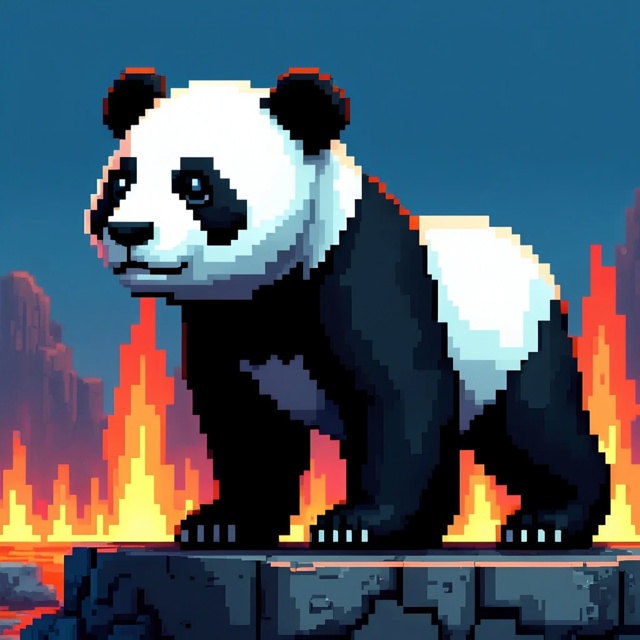} &
\includegraphics[width=\imwidth]{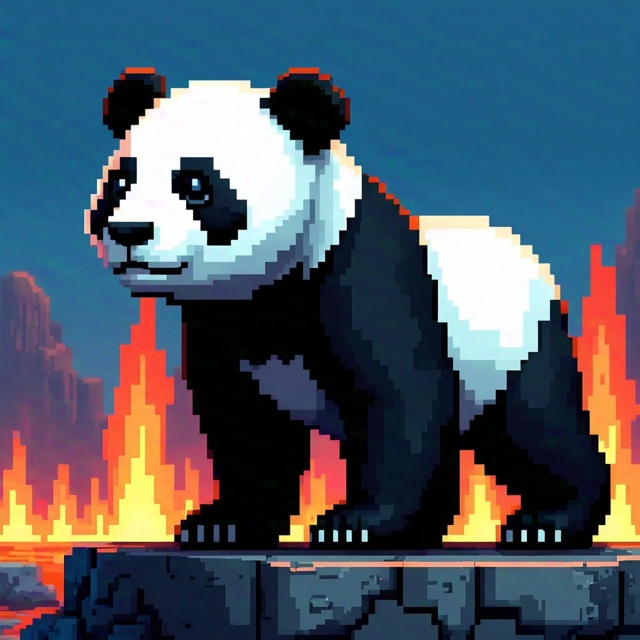} &
\includegraphics[width=\imwidth]{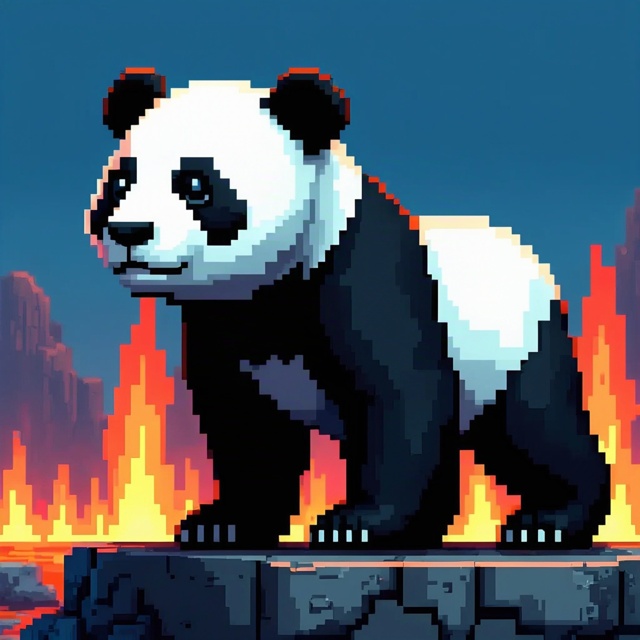} &
\includegraphics[width=\imwidth]{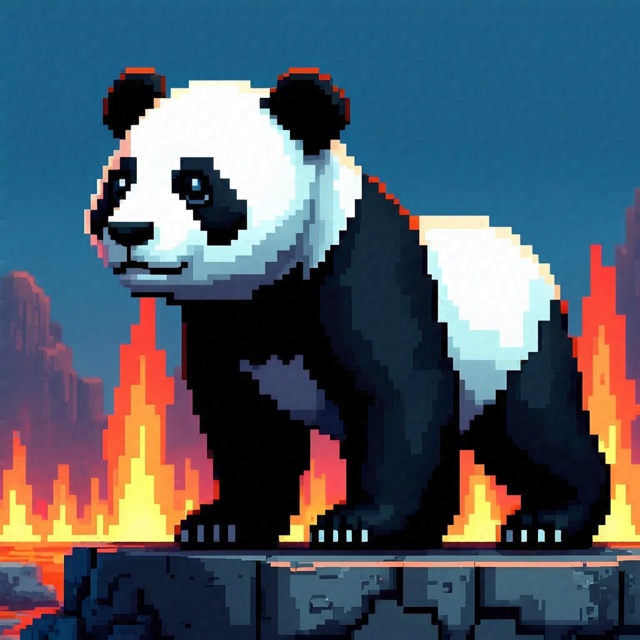} &
\includegraphics[width=\imwidth]{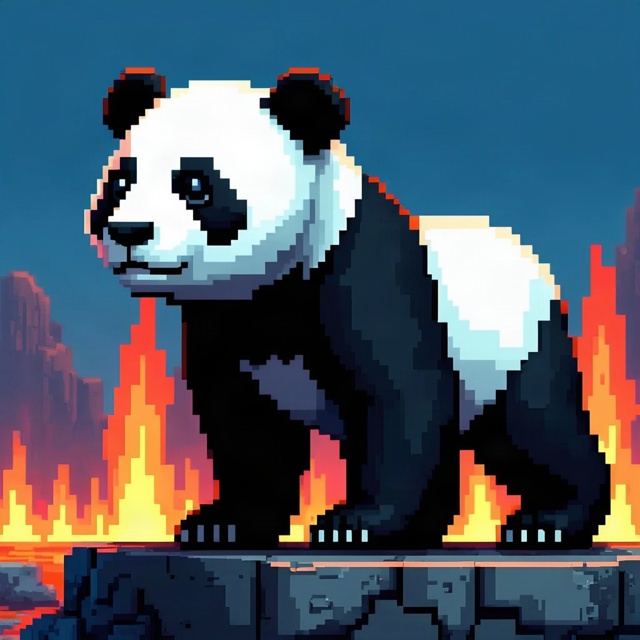} \\
& \includegraphics[width=\imwidth]{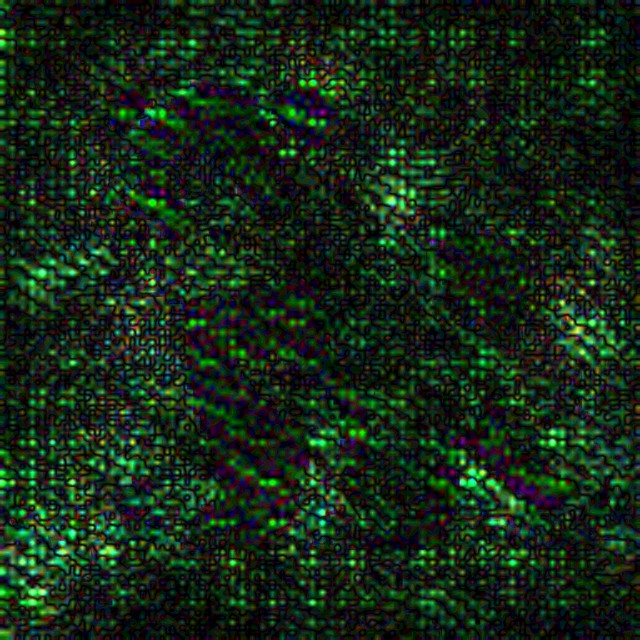} &
\includegraphics[width=\imwidth]{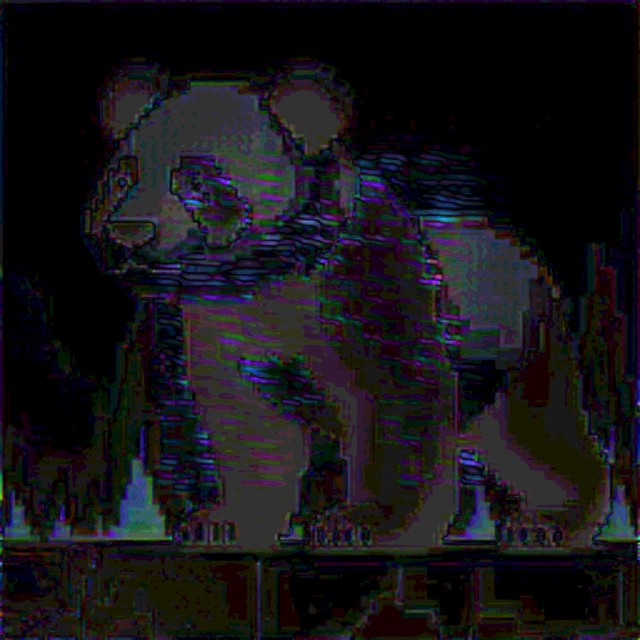} &
\includegraphics[width=\imwidth]{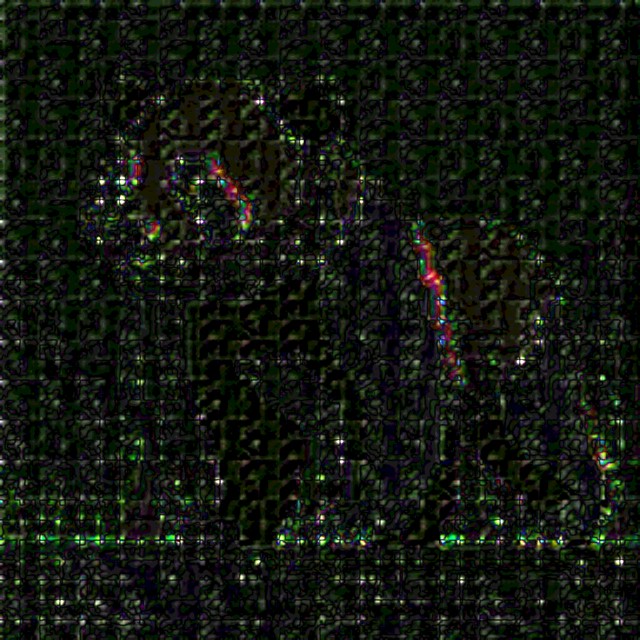} &
\includegraphics[width=\imwidth]{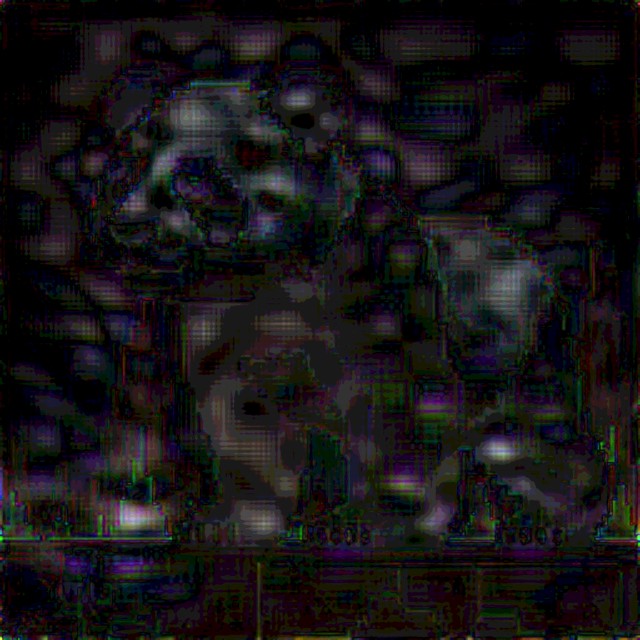}\\
\end{tabular}
}
\vspace{3pt}
\resizebox{1.0\linewidth}{!}{
\begin{tabular}{c@{\hskip 2pt}c@{\hskip 2pt}c@{\hskip 2pt}c@{\hskip 2pt}c}
TrustMark-Q & Video Seal 0.0 & Video Seal 1.0 & WAM & \ours{} (ours) \\
\includegraphics[width=\imwidth]{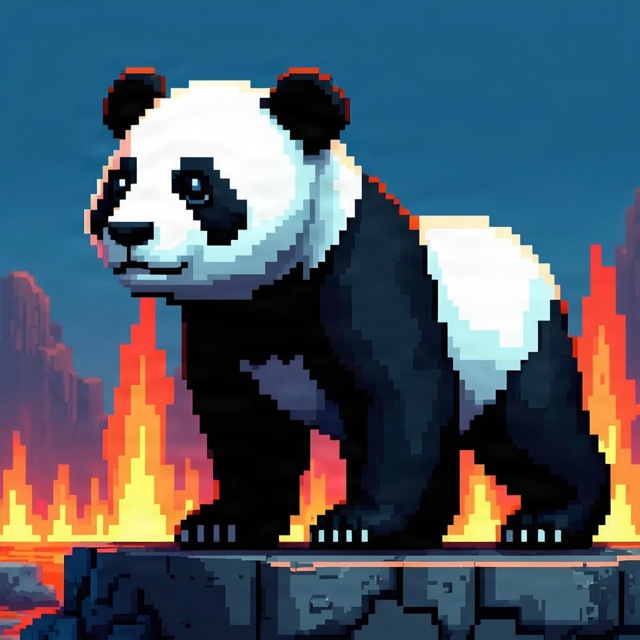} &
\includegraphics[width=\imwidth]{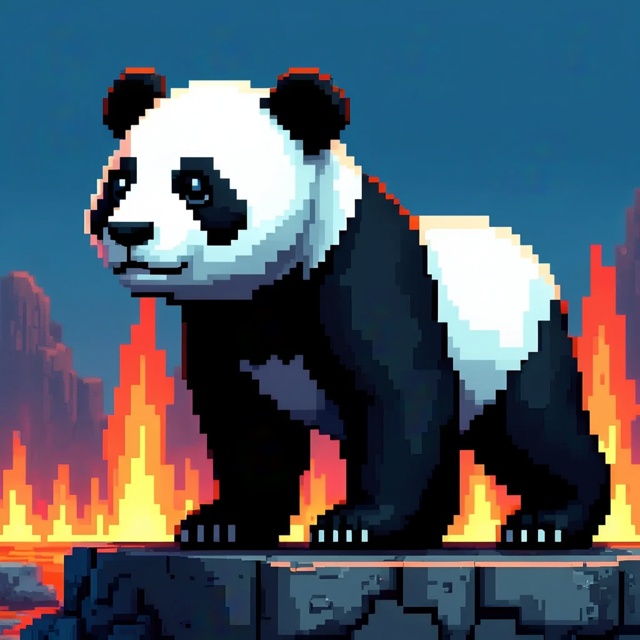} &
\includegraphics[width=\imwidth]{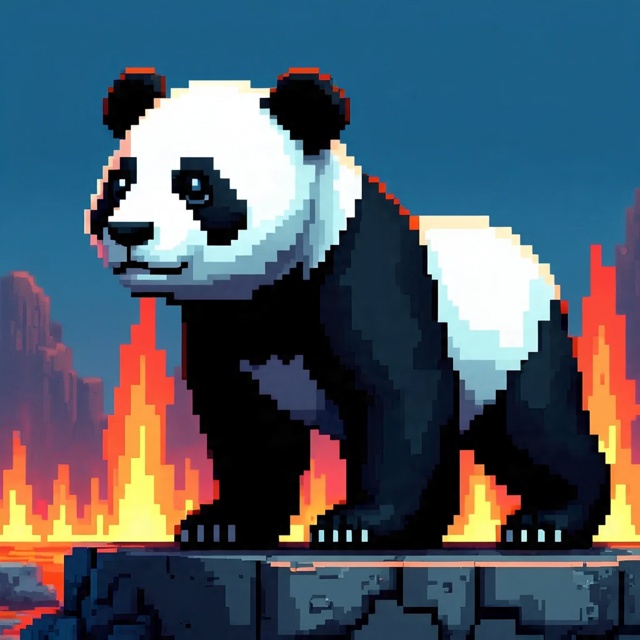} &
\includegraphics[width=\imwidth]{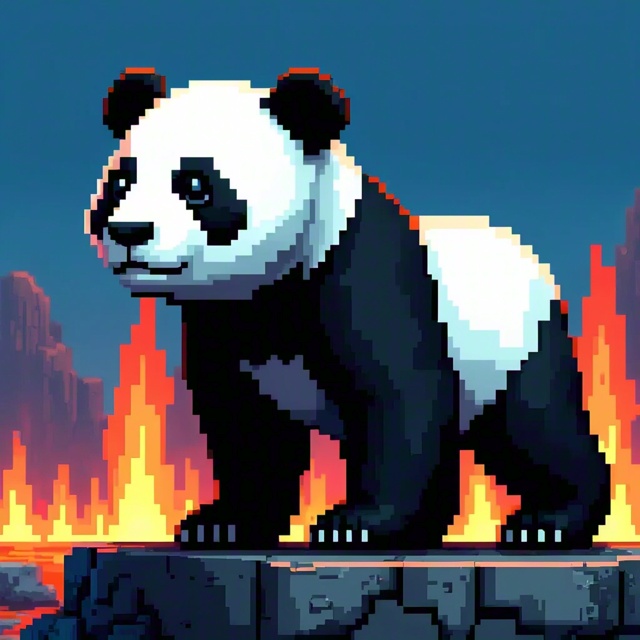} &
\includegraphics[width=\imwidth]{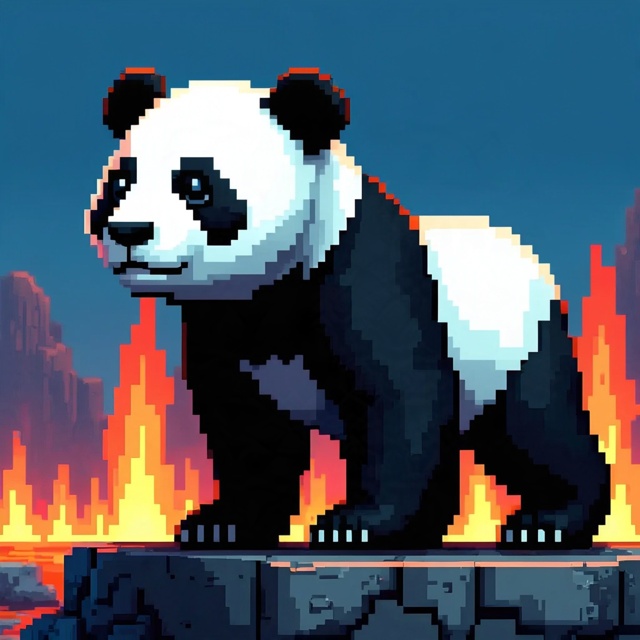} \\
\includegraphics[width=\imwidth]{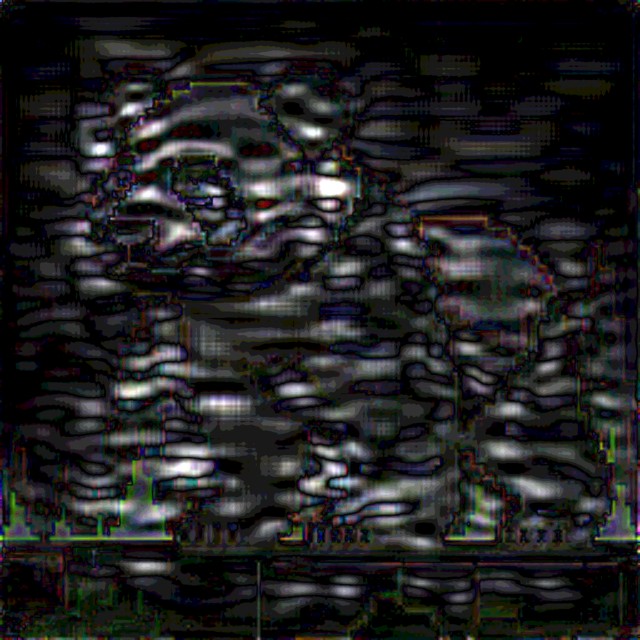} &
\includegraphics[width=\imwidth]{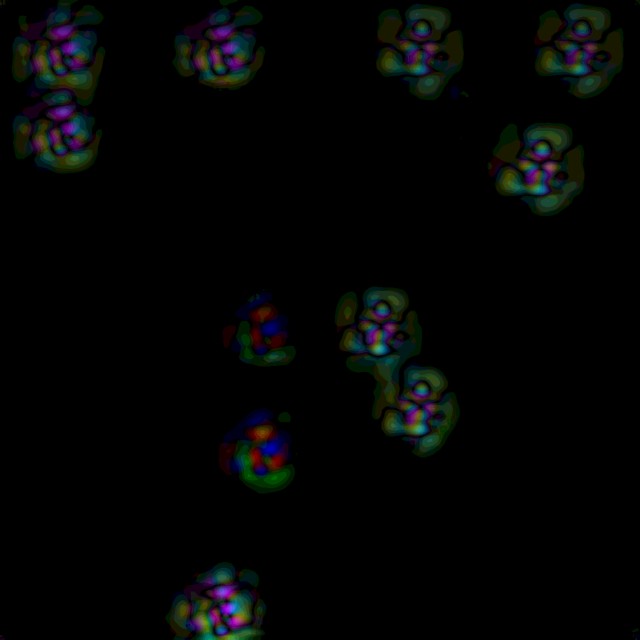} &
\includegraphics[width=\imwidth]{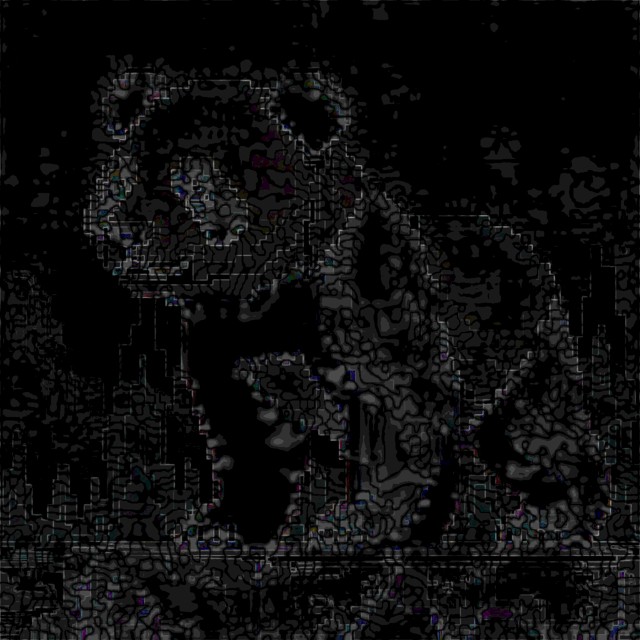} &
\includegraphics[width=\imwidth]{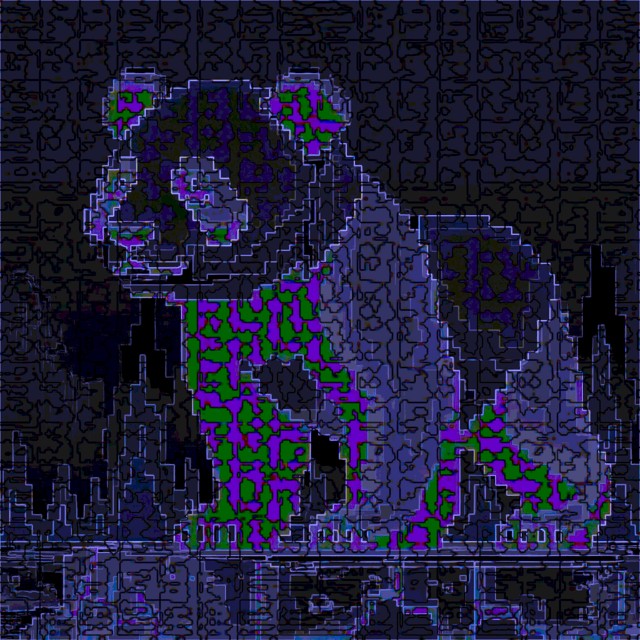} &
\includegraphics[width=\imwidth]{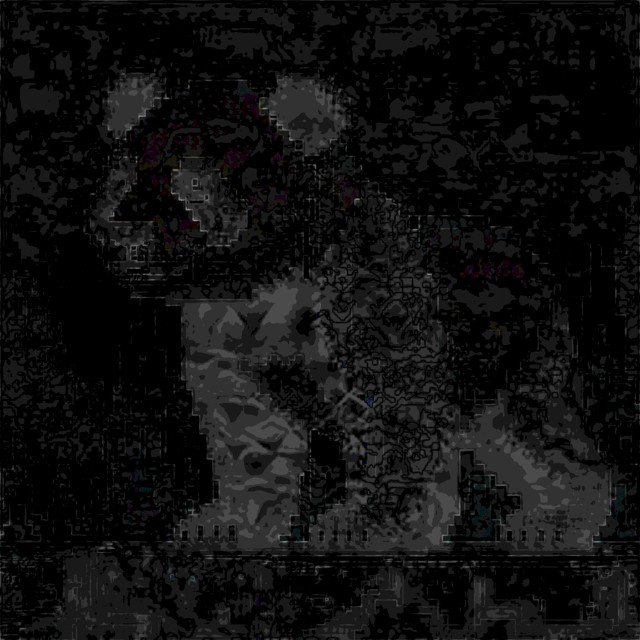} \\
    \end{tabular}
    }
    \caption{\textbf{Comparison with related work on an AI-generated image.} We show both the watermarked image (top) and the predicted watermark brightened for clarity (bottom).
    Many related methods leave visible artifacts in areas with a single color. In contrast, \ours\ does not leave visible artifacts in such areas while being more robust to various transformations.
}\label{appendix:fig:example_comp} 
\end{figure*}

\begin{figure*}
    \centering
    \resizebox{\linewidth}{!}{
        \begin{tabular}{r@{~~~~~~~~~~~~~~~}l}
\multicolumn{2}{c}{
\begin{tabular}{*{34}{c@{~~}}}
\multicolumn{34}{c}{\textit{Valuemetric attacks (images)}}\\
\toprule
\rotatebox[origin=c]{90}{Brightness 0.1}
&\rotatebox[origin=c]{90}{Brightness 0.25}
&\rotatebox[origin=c]{90}{Brightness 0.5}
&\rotatebox[origin=c]{90}{Brightness 0.75}
&\rotatebox[origin=c]{90}{Brightness 1.0}
&\rotatebox[origin=c]{90}{Brightness 1.25}
&\rotatebox[origin=c]{90}{Brightness 1.5}
&\rotatebox[origin=c]{90}{Brightness 1.75}
&\rotatebox[origin=c]{90}{Brightness 2.0}
&\rotatebox[origin=c]{90}{Contrast 0.1}
&\rotatebox[origin=c]{90}{Contrast 0.25}
&\rotatebox[origin=c]{90}{Contrast 0.5}
&\rotatebox[origin=c]{90}{Contrast 0.75}
&\rotatebox[origin=c]{90}{Contrast 1.0}
&\rotatebox[origin=c]{90}{Contrast 1.25}
&\rotatebox[origin=c]{90}{Contrast 1.5}
&\rotatebox[origin=c]{90}{Contrast 1.75}
&\rotatebox[origin=c]{90}{Contrast 2.0}
&\rotatebox[origin=c]{90}{Hue -0.4}
&\rotatebox[origin=c]{90}{Hue -0.3}
&\rotatebox[origin=c]{90}{Hue -0.2}
&\rotatebox[origin=c]{90}{Hue -0.1}
&\rotatebox[origin=c]{90}{Hue 0.0}
&\rotatebox[origin=c]{90}{Hue 0.1}
&\rotatebox[origin=c]{90}{Hue 0.2}
&\rotatebox[origin=c]{90}{Hue 0.3}
&\rotatebox[origin=c]{90}{Hue 0.4}
&\rotatebox[origin=c]{90}{Hue 0.5}
&\rotatebox[origin=c]{90}{Grayscale}
&\rotatebox[origin=c]{90}{GaussianBlur 3}
&\rotatebox[origin=c]{90}{GaussianBlur 5}
&\rotatebox[origin=c]{90}{GaussianBlur 9}
&\rotatebox[origin=c]{90}{GaussianBlur 13}
&\rotatebox[origin=c]{90}{GaussianBlur 17}\\
\midrule
0.80 &0.98 &1.00 &1.00 &1.00 &0.96 &0.94 &0.93 &0.91 &0.85 &0.99 &1.00 &1.00 &1.00 &0.96 &0.95 &0.94 &0.92 &0.98 &0.99 &0.99 &0.99 &1.00 &0.99 &0.98 &0.99 &0.99 &0.97 &1.00 &1.00 &1.00 &1.00 &1.00 &1.00 \\
\bottomrule
\end{tabular}}\\
\\
\multicolumn{2}{c}{\begin{tabular}{*{24}{c@{~~}}}
\multicolumn{24}{c}{\textit{Geometric attacks (images)}}\\
\toprule
\rotatebox[origin=c]{90}{HorizontalFlip}
&\rotatebox[origin=c]{90}{Rotate 5}
&\rotatebox[origin=c]{90}{Rotate 10}
&\rotatebox[origin=c]{90}{Rotate 30}
&\rotatebox[origin=c]{90}{Rotate 45}
&\rotatebox[origin=c]{90}{Rotate 90}
&\rotatebox[origin=c]{90}{Crop 0.32}
&\rotatebox[origin=c]{90}{Crop 0.45}
&\rotatebox[origin=c]{90}{Crop 0.55}
&\rotatebox[origin=c]{90}{Crop 0.63}
&\rotatebox[origin=c]{90}{Crop 0.71}
&\rotatebox[origin=c]{90}{Crop 0.77}
&\rotatebox[origin=c]{90}{Crop 0.84}
&\rotatebox[origin=c]{90}{Crop 0.89}
&\rotatebox[origin=c]{90}{Crop 0.95}
&\rotatebox[origin=c]{90}{Crop 1.0}
&\rotatebox[origin=c]{90}{Perspective 0.1}
&\rotatebox[origin=c]{90}{Perspective 0.2}
&\rotatebox[origin=c]{90}{Perspective 0.3}
&\rotatebox[origin=c]{90}{Perspective 0.4}
&\rotatebox[origin=c]{90}{Perspective 0.5}
&\rotatebox[origin=c]{90}{Perspective 0.6}
&\rotatebox[origin=c]{90}{Perspective 0.7}
&\rotatebox[origin=c]{90}{Perspective 0.8}
\\
\midrule
1.00 &0.99 &0.99 &0.99 &0.98 &0.99 &0.58 &0.81 &0.90 &0.95 &0.98 &0.99 &0.99 &0.99 &0.99 &1.00 &1.00 &0.99 &0.99 &0.99 &0.98 &0.96 &0.93 &0.87 \\
\bottomrule
\end{tabular}}\\
\\
\begin{tabular}{*{6}{c@{~~}}}
\multicolumn{6}{c}{\textit{Compression attacks (images)}}\\
\toprule
\rotatebox[origin=c]{90}{JPEG 40}
&\rotatebox[origin=c]{90}{JPEG 50}
&\rotatebox[origin=c]{90}{JPEG 60}
&\rotatebox[origin=c]{90}{JPEG 70}
&\rotatebox[origin=c]{90}{JPEG 80}
&\rotatebox[origin=c]{90}{JPEG 90}\\
\midrule
0.96 &0.97 &0.97 &0.99 &0.99 &1.00\\
\bottomrule
\end{tabular}&
\begin{tabular}{c@{~}ccc@{~}ccc@{~}c}
\multicolumn{8}{c}{\textit{Combined attacks (images)}}\\
\toprule
\rotatebox[origin=c]{90}{(JPEG Crop Brightness)}
&\rotatebox[origin=c]{90}{(40 0.71 0.5)}
&
&\rotatebox[origin=c]{90}{(JPEG Crop Brightness)}
&\rotatebox[origin=c]{90}{(60 0.71 0.5)}
&
&\rotatebox[origin=c]{90}{(JPEG Crop Brightness)}
&\rotatebox[origin=c]{90}{(80 0.71 0.5)}\\
\midrule
\multicolumn{2}{c}{0.86} &&\multicolumn{2}{c}{0.91} &&\multicolumn{2}{c}{0.96} \\
\bottomrule
\end{tabular}\\

\\
\hline

\\
\begin{tabular}{*{9}{c@{~~}}}
\multicolumn{9}{c}{\textit{Valuemetric attacks (videos)}}\\
\toprule
\rotatebox[origin=c]{90}{Brightness 0.5}
&\rotatebox[origin=c]{90}{Brightness 1.5}
&\rotatebox[origin=c]{90}{Contrast 0.5}
&\rotatebox[origin=c]{90}{Contrast 1.5}
&\rotatebox[origin=c]{90}{Saturation 0.5}
&\rotatebox[origin=c]{90}{Saturation 1.5}
&\rotatebox[origin=c]{90}{Hue 0.25}
&\rotatebox[origin=c]{90}{Grayscale}
&\rotatebox[origin=c]{90}{GaussianBlur 9}\\
\midrule
1.00 &1.00 &1.00 &1.00 &1.00 &1.00 &1.00 &1.00 &1.00\\
\bottomrule
\end{tabular}&
\begin{tabular}{*{6}{c@{~~}}}
\multicolumn{6}{c}{\textit{Geometric attacks (videos)}}\\
\toprule
\rotatebox[origin=c]{90}{HorizontalFlip}
&\rotatebox[origin=c]{90}{Rotate 10}
&\rotatebox[origin=c]{90}{Rotate 90}
&\rotatebox[origin=c]{90}{Crop 0.55}
&\rotatebox[origin=c]{90}{Crop 0.71}
&\rotatebox[origin=c]{90}{Perspective 0.5}\\
\midrule
1.00 &1.00 &0.95 &0.95 &1.00 &1.00 \\
\bottomrule
\end{tabular}\\
\\
\hspace{3cm}\begin{tabular}{*{13}{c@{~~}}}
\multicolumn{13}{c}{\textit{Compression attacks (videos)}}\\
\toprule
\rotatebox[origin=c]{90}{JPEG 40}
&\rotatebox[origin=c]{90}{H264 23}
&\rotatebox[origin=c]{90}{H264 30}
&\rotatebox[origin=c]{90}{H264 40}
&\rotatebox[origin=c]{90}{H264 50}
&\rotatebox[origin=c]{90}{H264rgb 23}
&\rotatebox[origin=c]{90}{H264rgb 30}
&\rotatebox[origin=c]{90}{H264rgb 40}
&\rotatebox[origin=c]{90}{H264rgb 50}
&\rotatebox[origin=c]{90}{H265 23}
&\rotatebox[origin=c]{90}{H265 30}
&\rotatebox[origin=c]{90}{H265 40}
&\rotatebox[origin=c]{90}{H265 50}\\
\midrule
1.00 &1.00 &0.98 &0.77 &0.52 &1.00 &0.99 &0.90 &0.70 &1.00 &0.98 &0.74 &0.54 \\
\bottomrule
\end{tabular}&
\begin{tabular}{c@{~}ccc@{~}ccc@{~}ccc@{~}c}
\multicolumn{11}{c}{\textit{Combined attacks (videos)}}\\
\toprule
\rotatebox[origin=c]{90}{(H264 Crop Brightness)}
&\rotatebox[origin=c]{90}{(23 0.71 0.5)}
&
&\rotatebox[origin=c]{90}{(H264 Crop Brightness)}
&\rotatebox[origin=c]{90}{(30 0.71 0.5)}
&
&\rotatebox[origin=c]{90}{(H264 Crop Brightness)}
&\rotatebox[origin=c]{90}{(40 0.71 0.5)}
&
&\rotatebox[origin=c]{90}{(H264 Crop Brightness)}
&\rotatebox[origin=c]{90}{(50 0.71 0.5)}\\
\midrule
\multicolumn{2}{c}{0.95} &&\multicolumn{2}{c}{0.76} &&\multicolumn{2}{c}{0.51} &&\multicolumn{2}{c}{0.50} \\
\bottomrule
\end{tabular}\\
\end{tabular}

    }
    \caption{
        The full list of attacks used for the image evaluation (top) and video evaluation (bottom). For each attack, we also report the bit accuracy of \ours{} on the Meta AI images (top) and SA-V videos (bottom). Selected attacks are visualized in Figure~\ref{fig:app-augs}.
    }

\end{figure*}

\begin{figure*}[t]
  \centering
  \footnotesize
  \begin{tabular}{*{4}{l}}
       Identity & Crop 0.33 & Rotation 10 & Rotation 90 \\
       \begin{minipage}{.22\linewidth}\includegraphics[width=\linewidth]{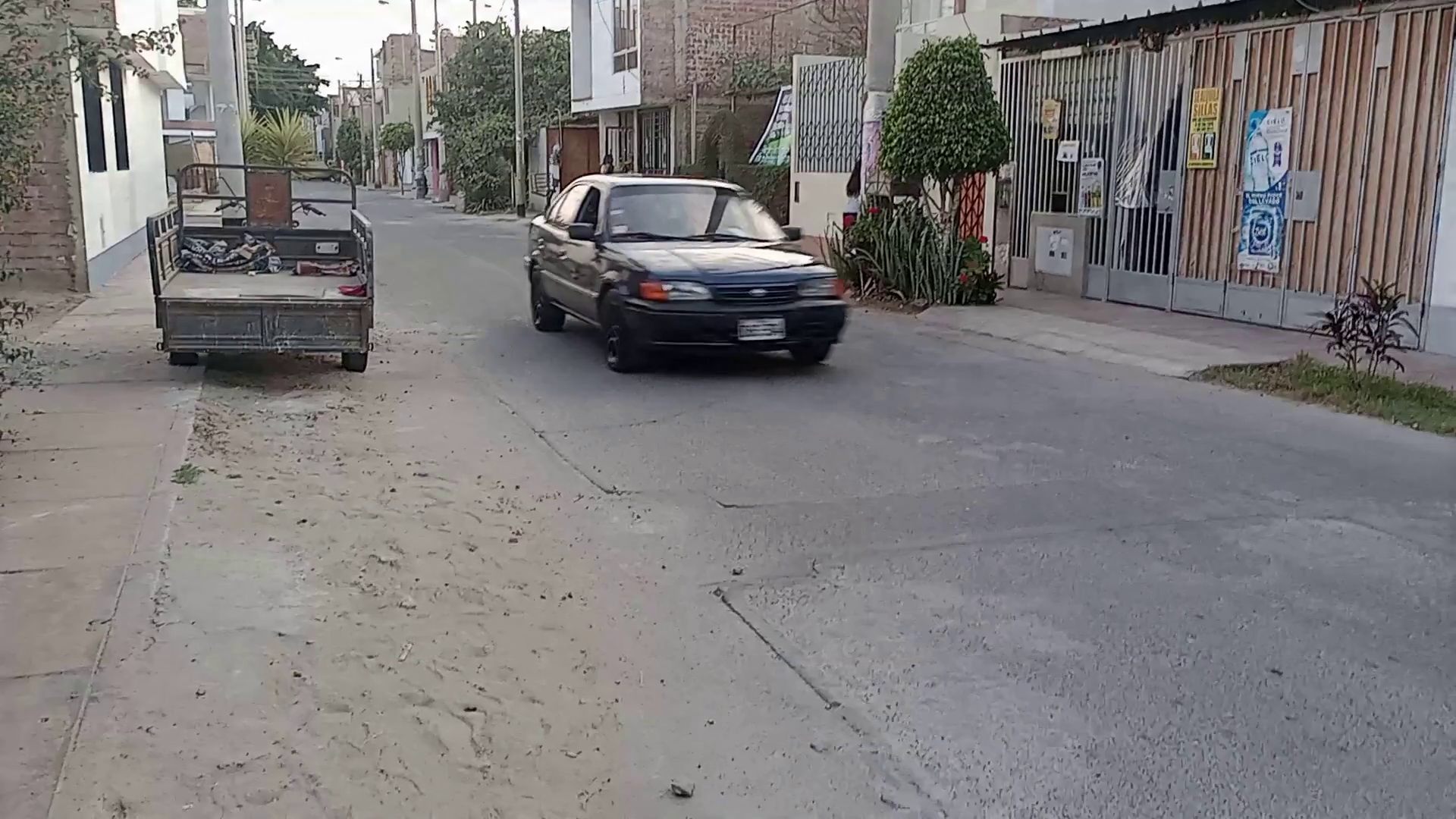}\end{minipage} &  
       \begin{minipage}{.22\linewidth}\includegraphics[width=\linewidth]{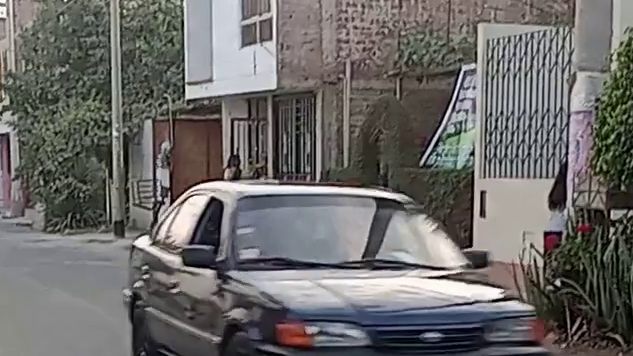}\end{minipage} &  
       \begin{minipage}{.22\linewidth}\includegraphics[width=\linewidth]{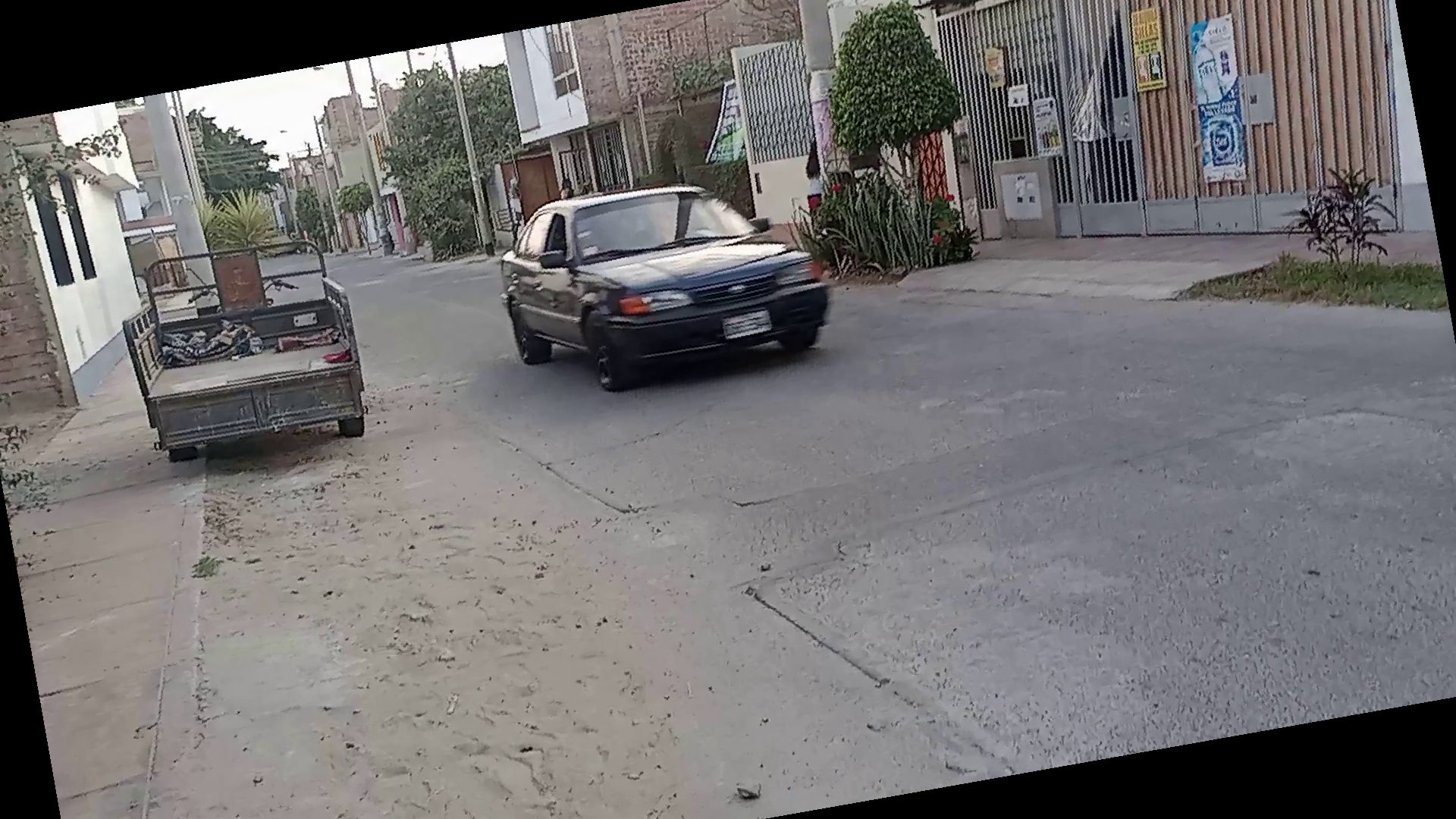}\end{minipage} &  
       \begin{minipage}{.22\linewidth}\includegraphics[width=0.5\linewidth]{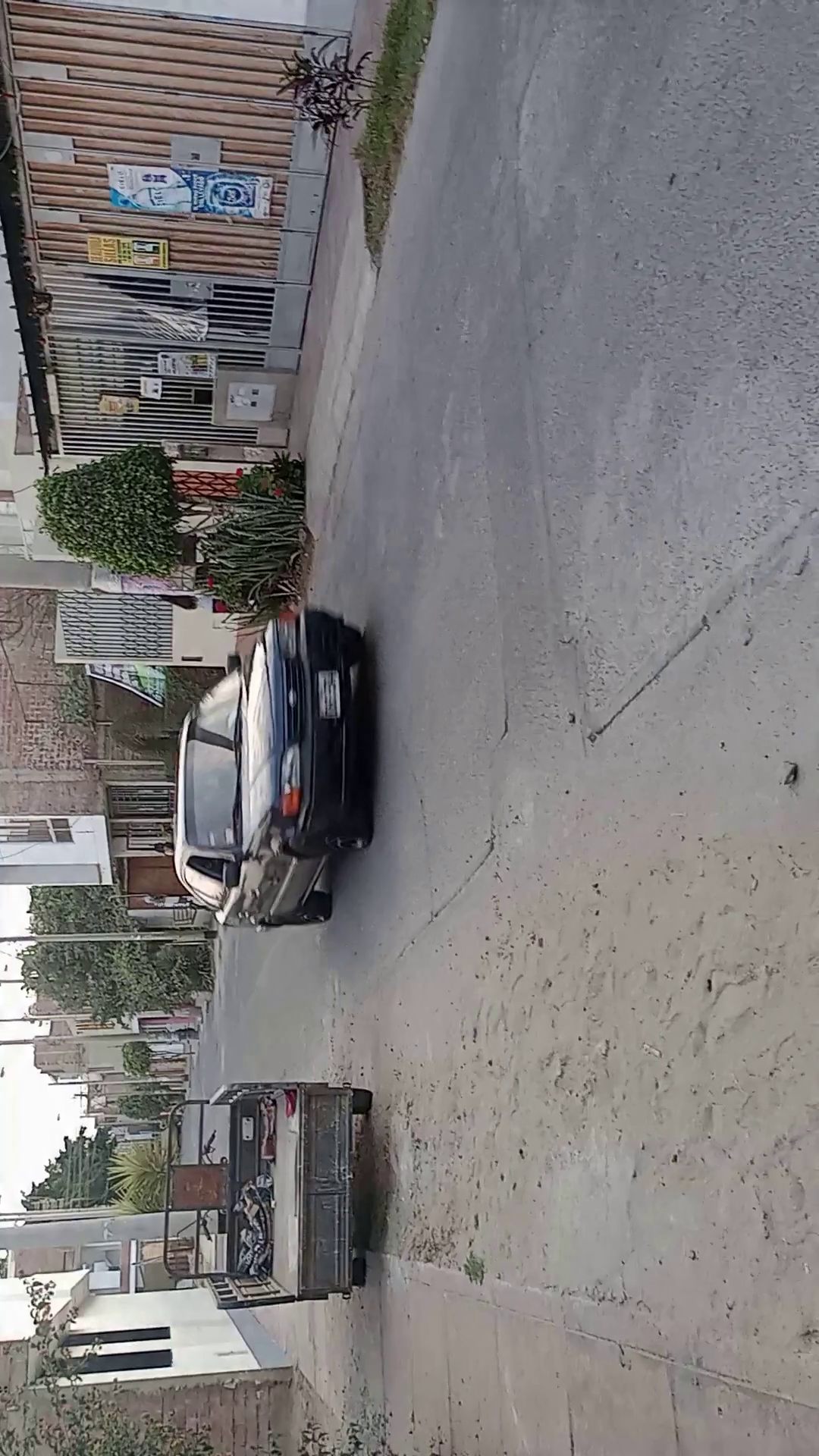}\end{minipage}
       \\ \\
       Contrast 0.5 & Contrast 1.5 & Brightness 0.5 & Brightness 1.5 \\
       \begin{minipage}{.22\linewidth}\includegraphics[width=\linewidth]{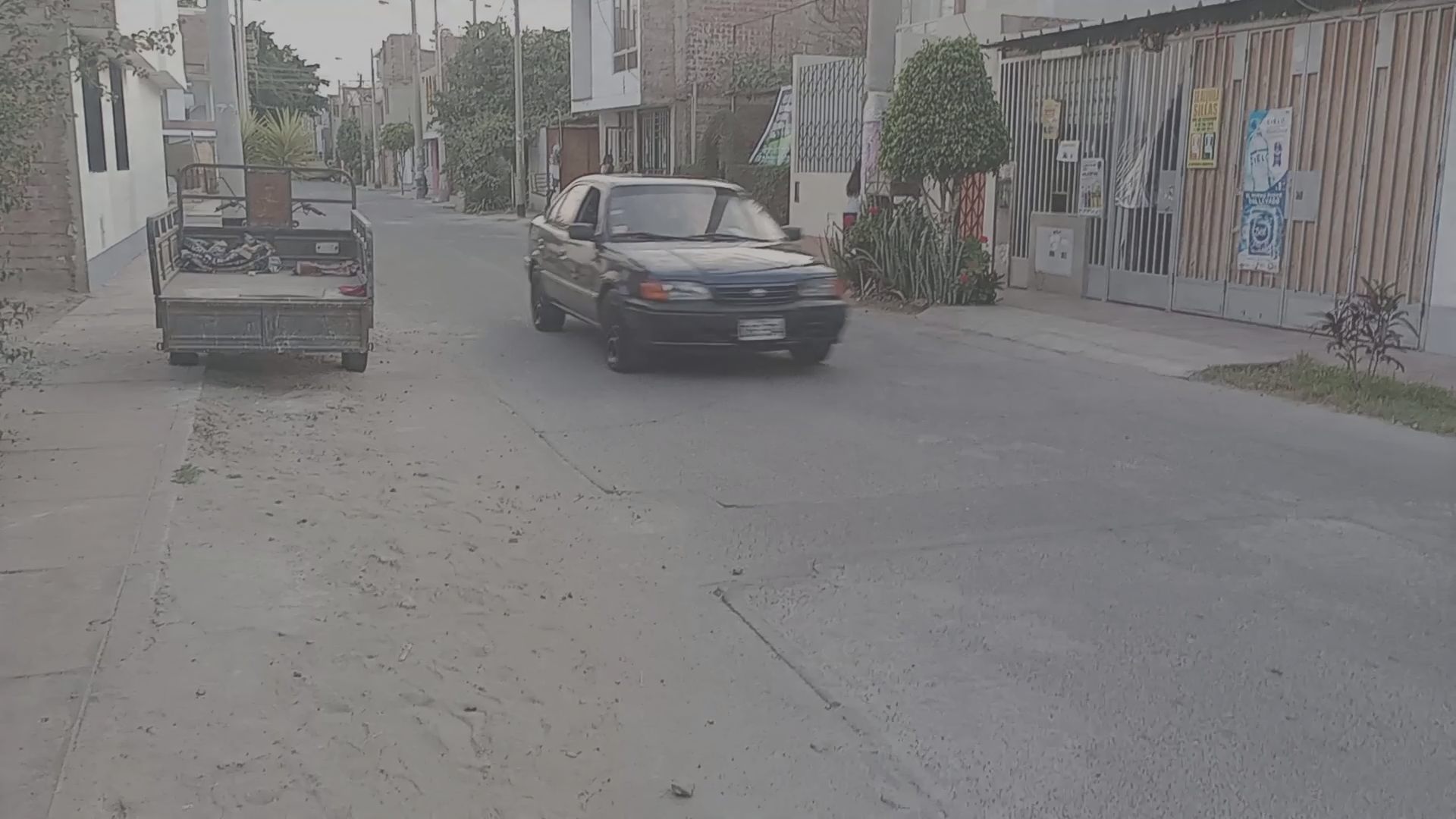}\end{minipage} &  
       \begin{minipage}{.22\linewidth}\includegraphics[width=\linewidth]{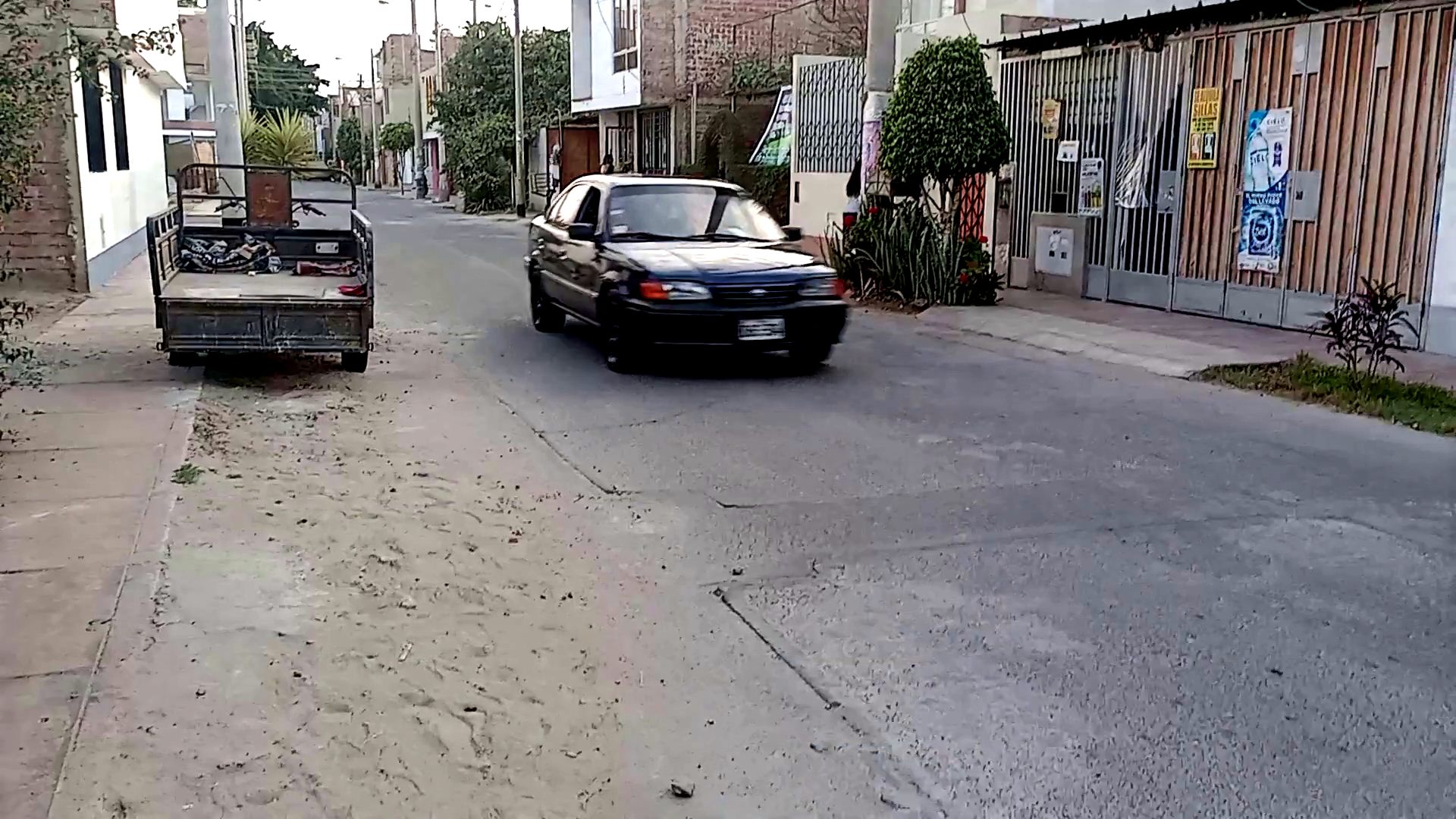}\end{minipage} &  
       \begin{minipage}{.22\linewidth}\includegraphics[width=\linewidth]{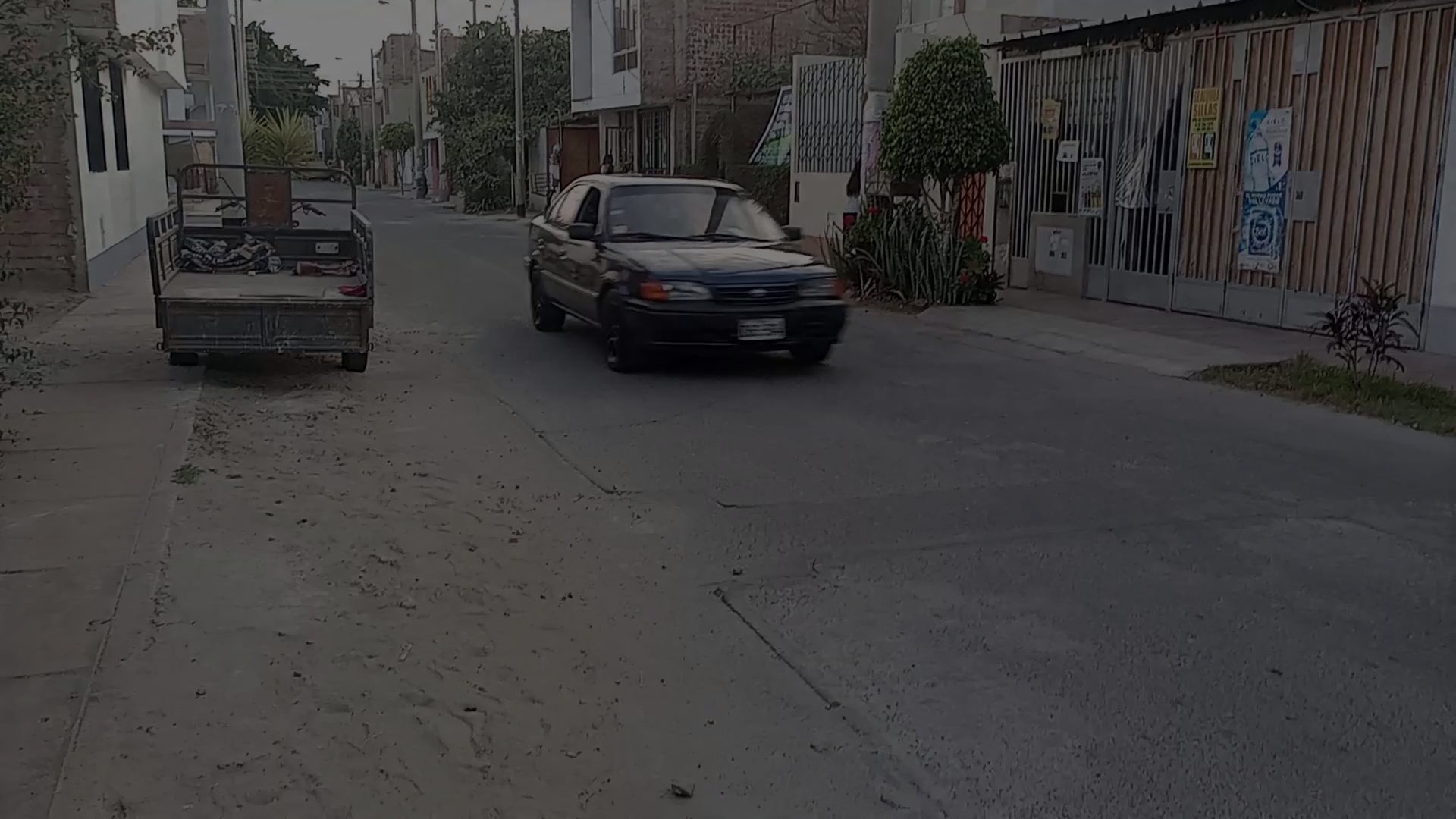}\end{minipage} &  
       \begin{minipage}{.22\linewidth}\includegraphics[width=\linewidth]{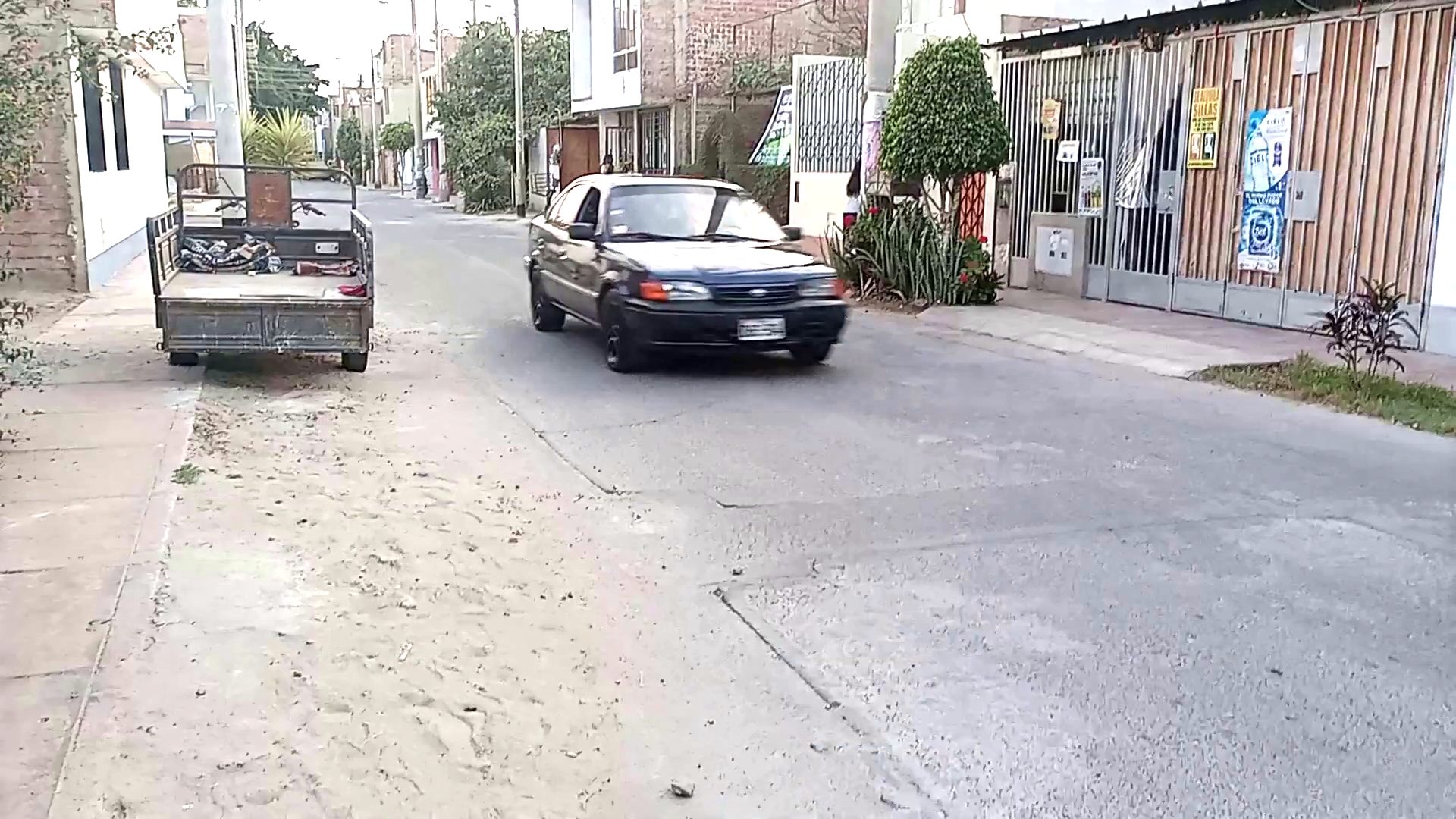}\end{minipage}
       \\ \\
       Hue -0.1 & Saturation 1.5 & Resize 0.5 & JPEG 40 \\
       \begin{minipage}{.22\linewidth}\includegraphics[width=\linewidth]{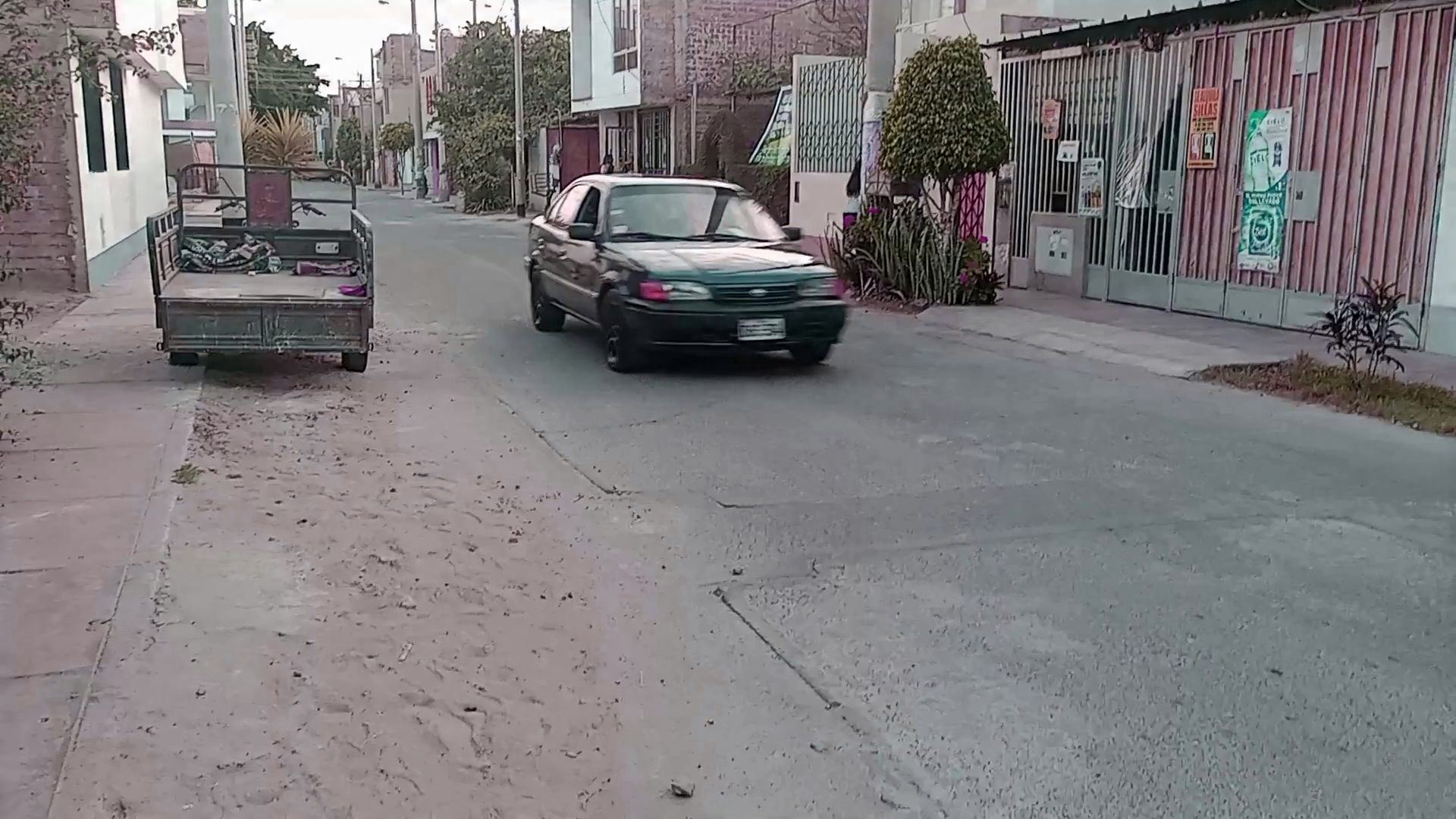}\end{minipage} &
       \begin{minipage}{.22\linewidth}\includegraphics[width=\linewidth]{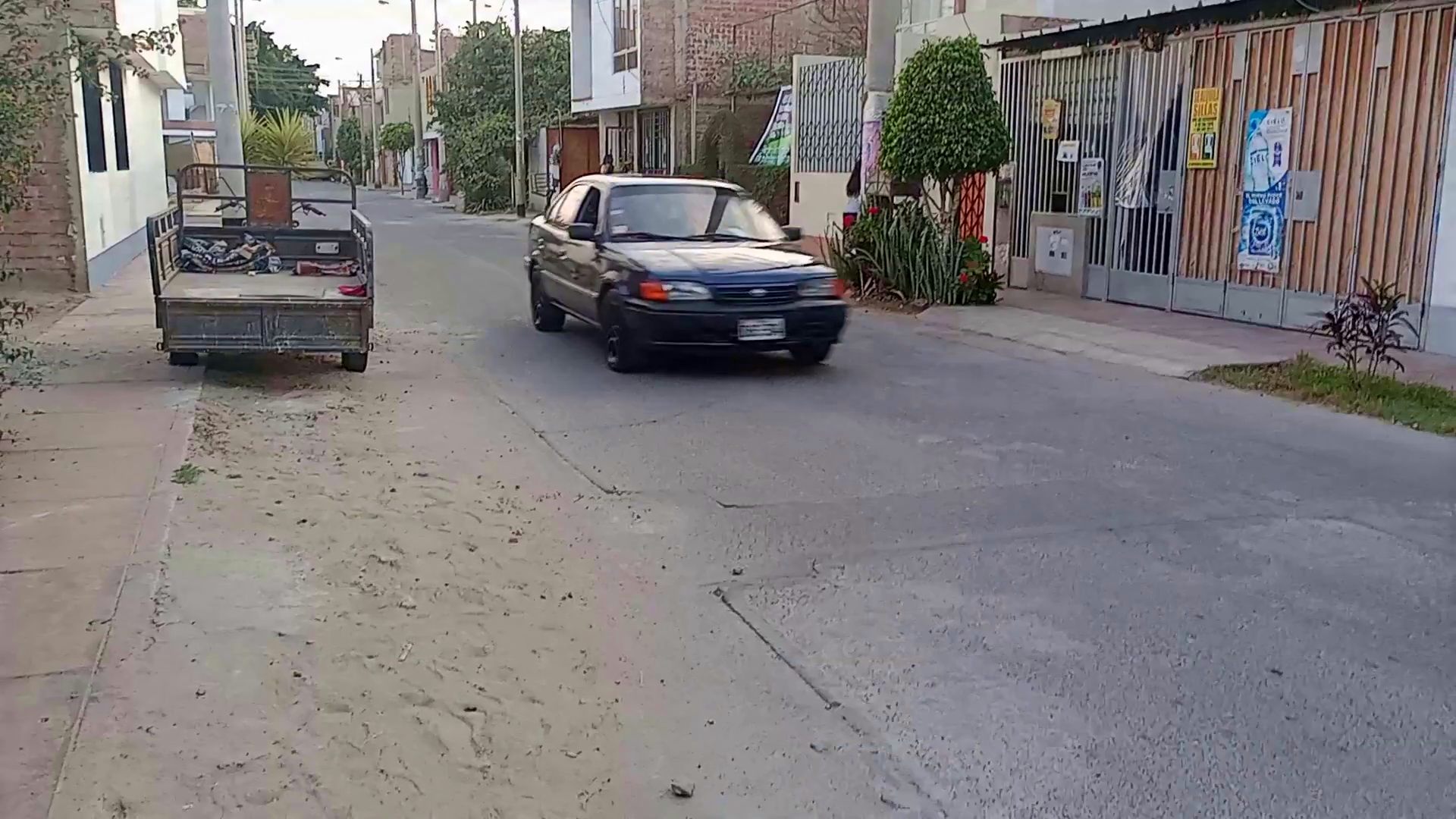}\end{minipage} &  
       \begin{minipage}{.22\linewidth}\includegraphics[width=\linewidth]{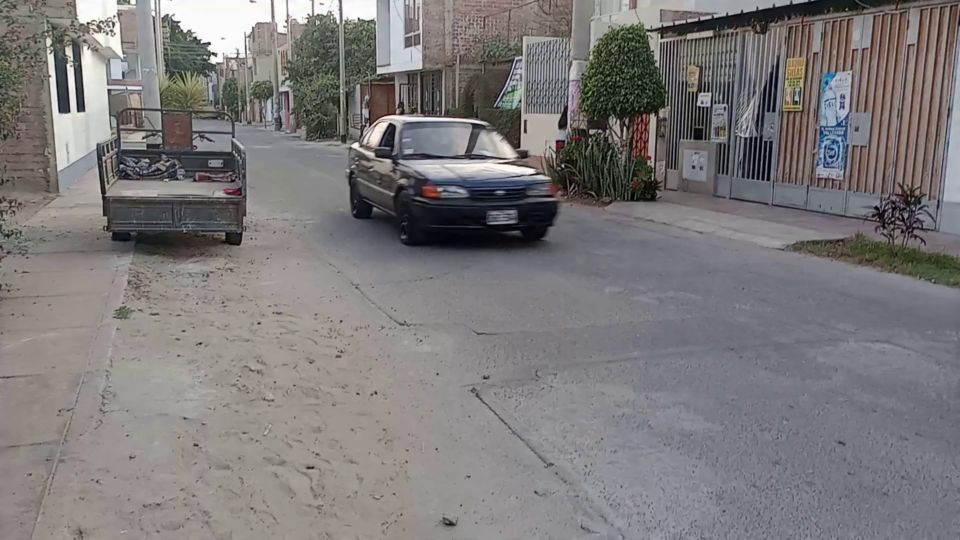}\end{minipage} &  
       \begin{minipage}{.22\linewidth}\includegraphics[width=\linewidth]{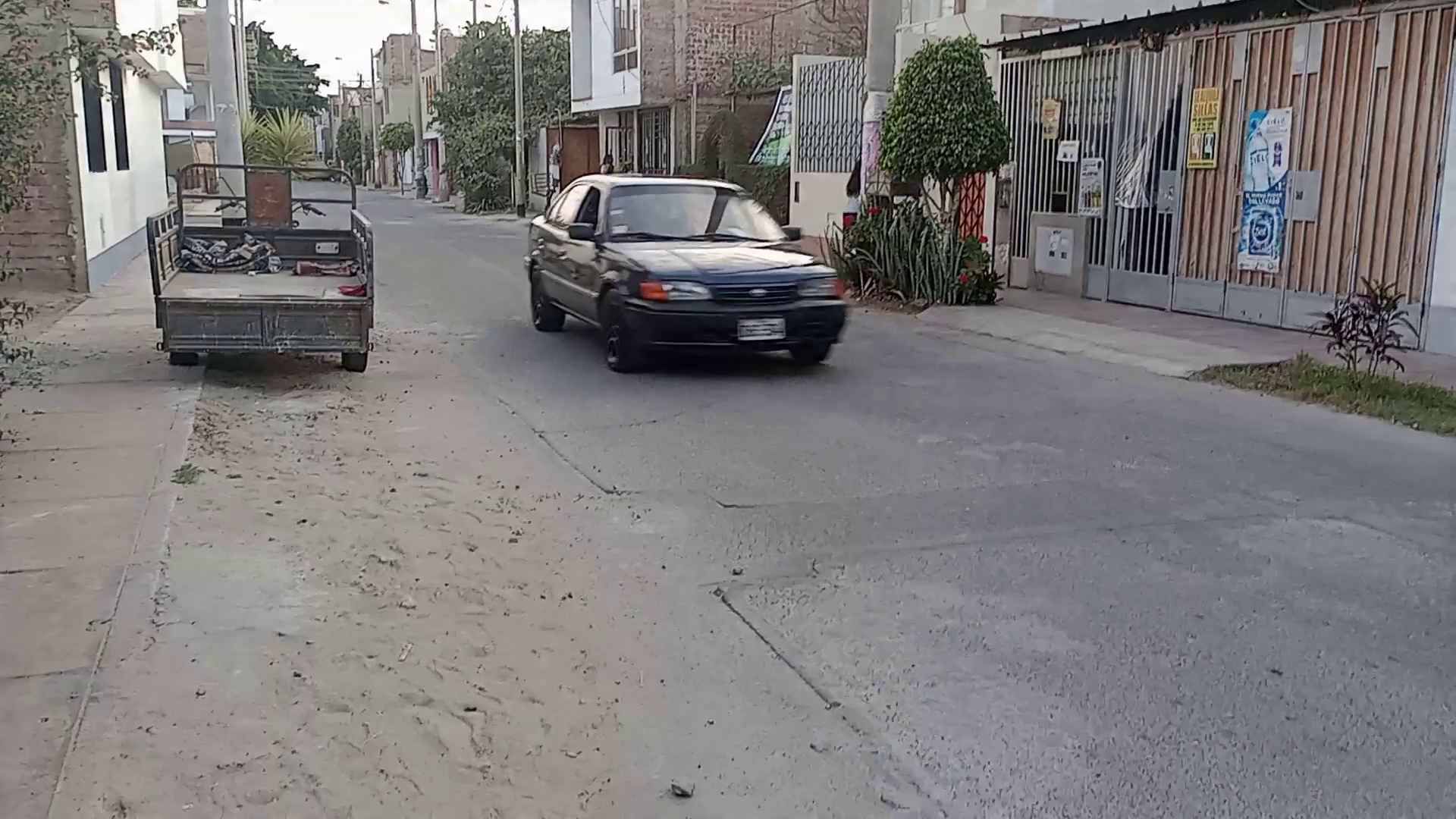}\end{minipage}
       \\ \\
       H264 40 & Horizontal flipping & Gaussian blur 17 & Perspective 0.5 \\
       \begin{minipage}{.22\linewidth}\includegraphics[width=\linewidth]{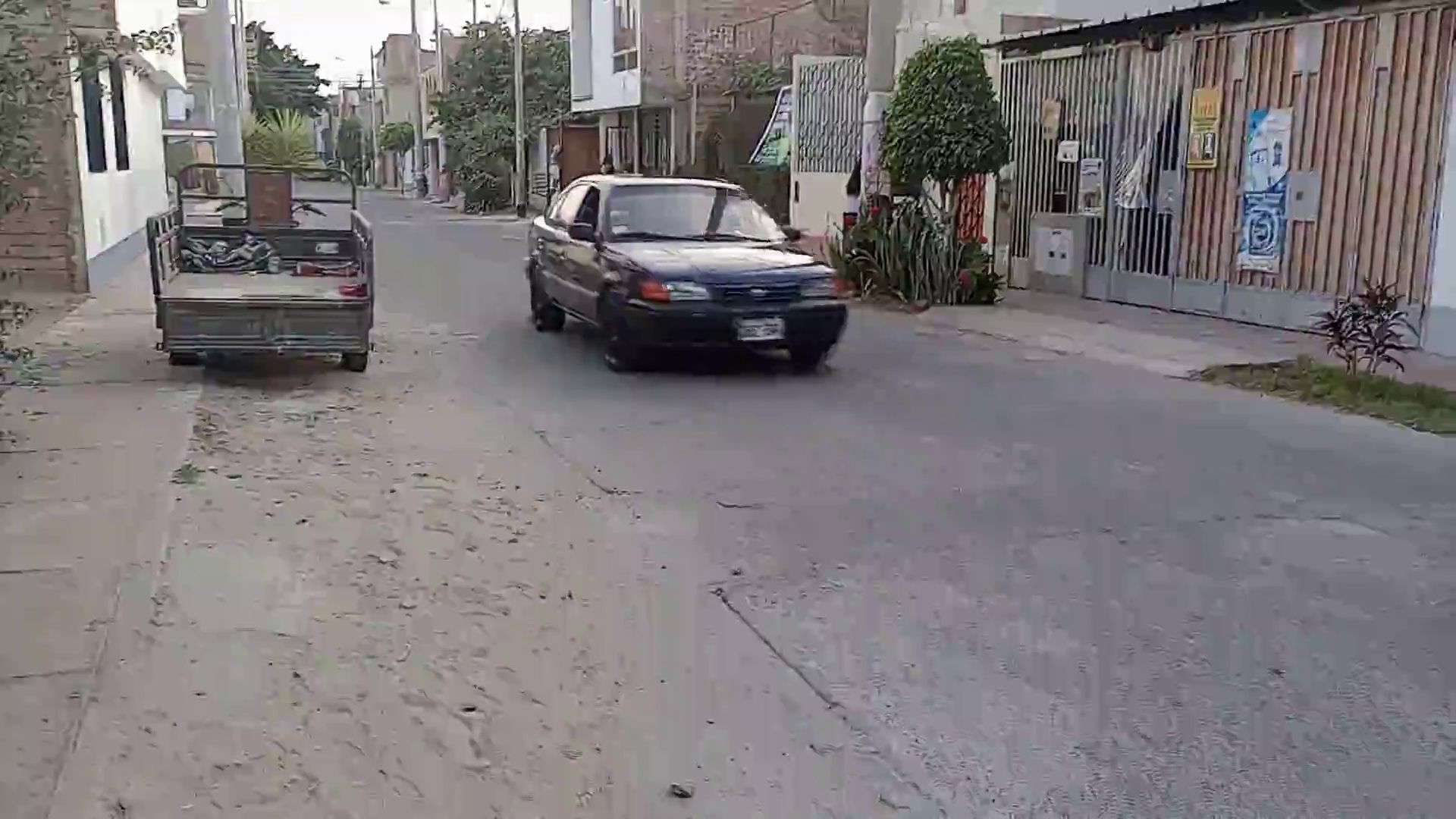}\end{minipage} &  
       \begin{minipage}{.22\linewidth}\includegraphics[width=\linewidth]{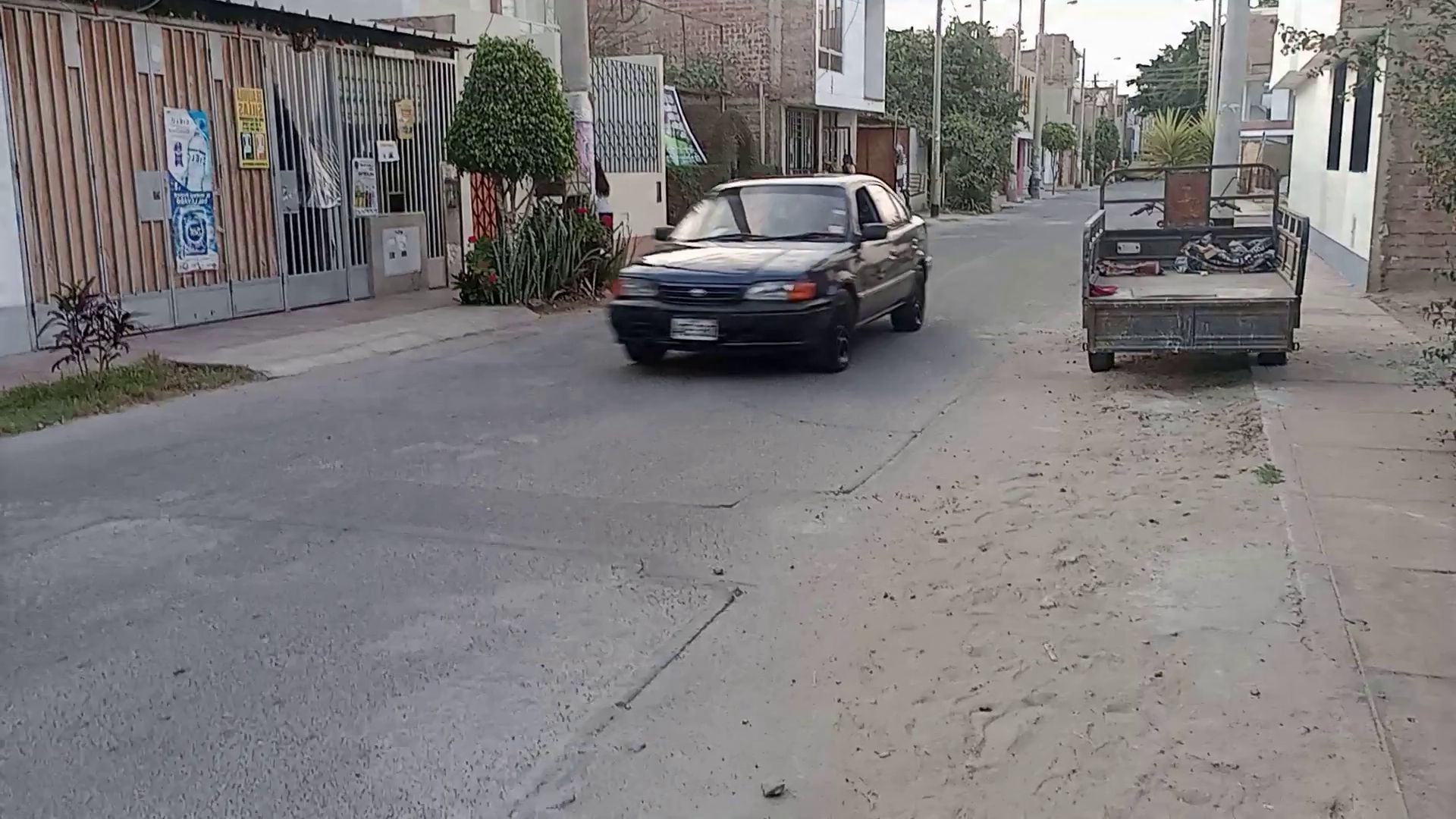}\end{minipage} &  
       \begin{minipage}{.22\linewidth}\includegraphics[width=\linewidth]{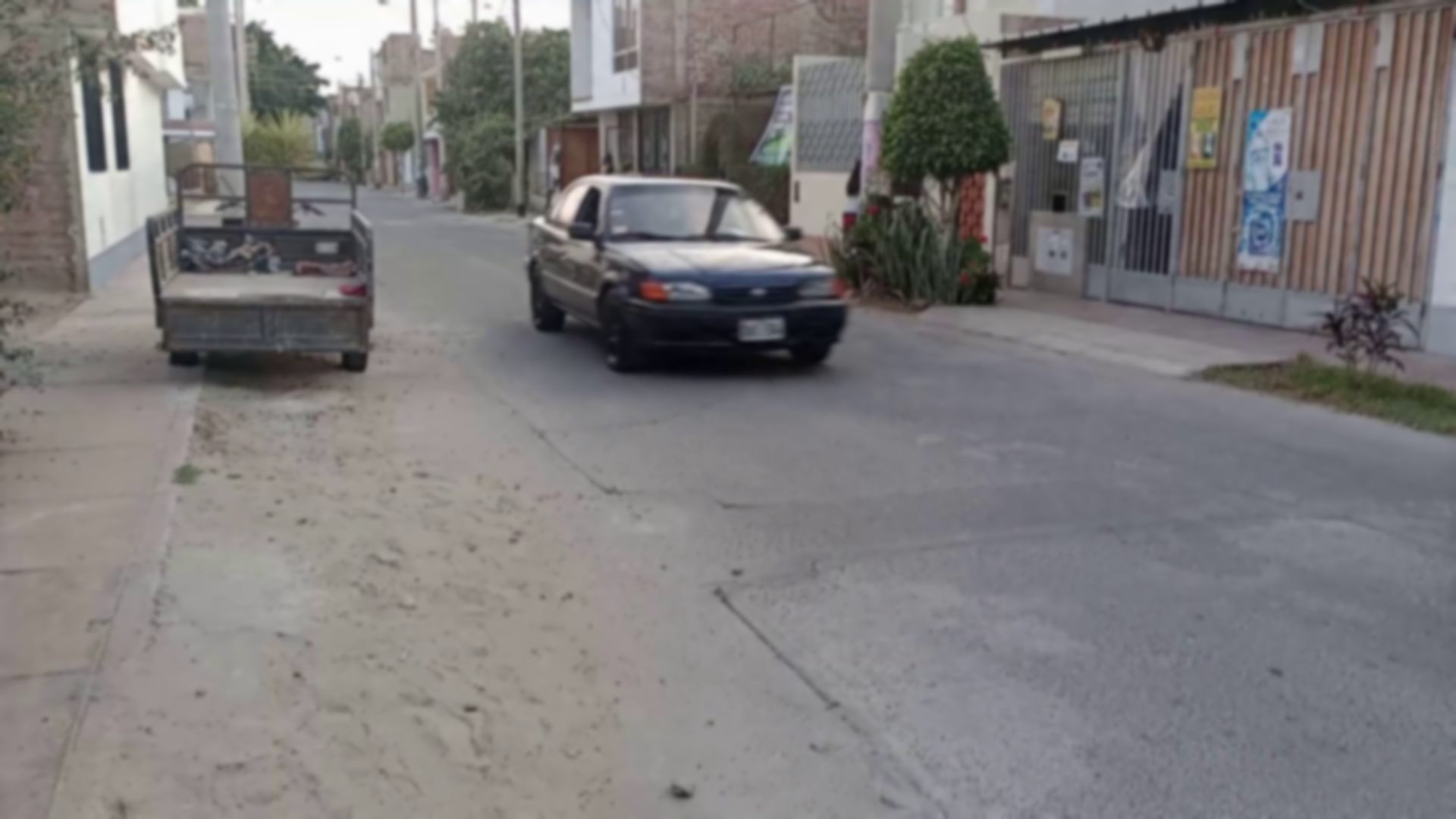}\end{minipage} &  
       \begin{minipage}{.22\linewidth}\includegraphics[width=\linewidth]{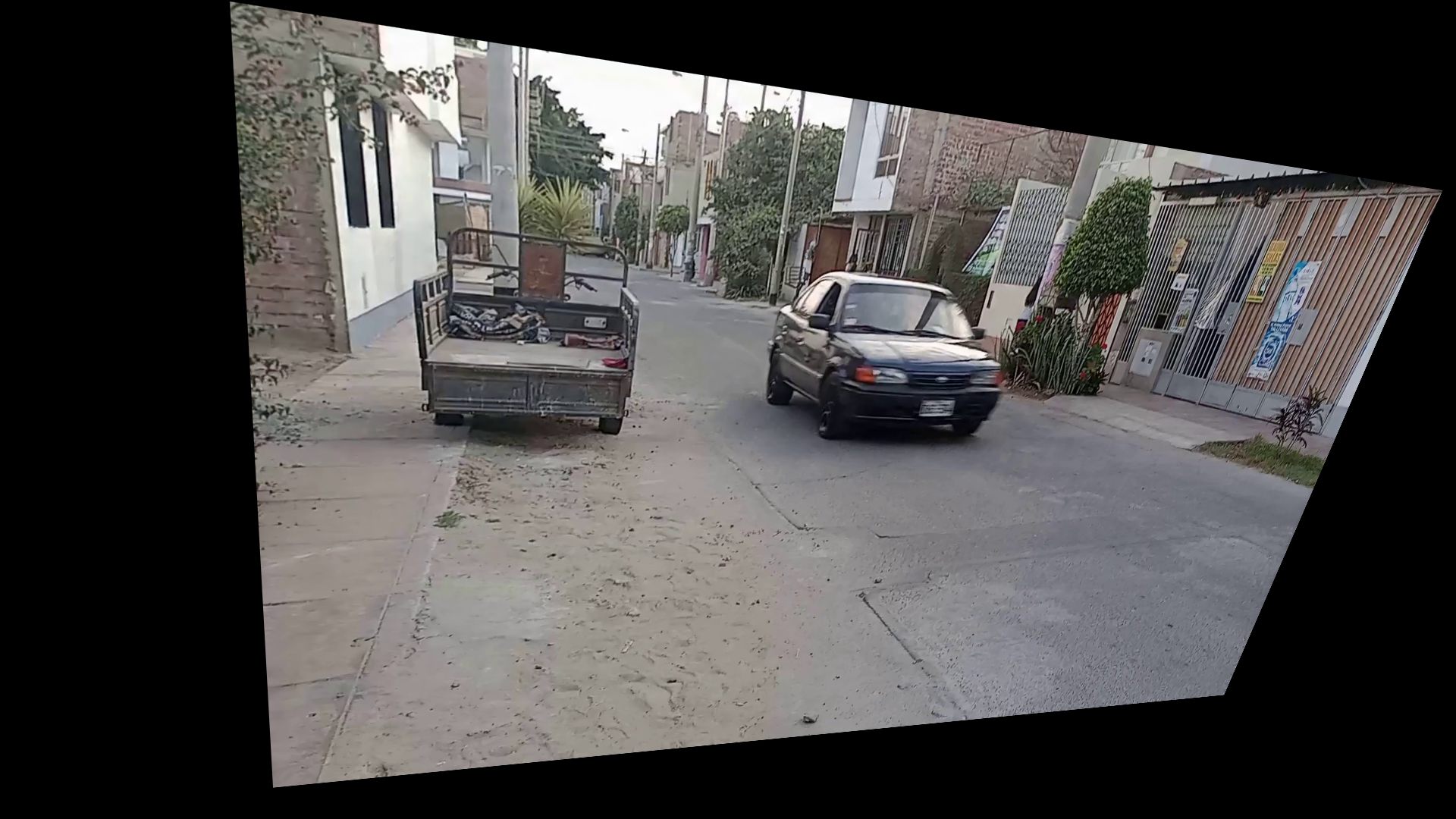}\end{minipage}
       \\
  \end{tabular}
    \caption{Illustration of some of the selected transformations.}
  \label{fig:app-augs}

\end{figure*}

\end{document}